\documentclass[10pt,letterpaper]{article}
\usepackage[top=0.85in,left=2.75in,footskip=0.75in]{geometry}
\usepackage{multicol, multirow}
\usepackage{graphicx}
\usepackage[dvipsnames, table]{xcolor}

\usepackage{comment}
\usepackage{soul}

\usepackage{amsmath,amssymb}

\usepackage{changepage}

\usepackage[utf8x]{inputenc}

\usepackage{textcomp,marvosym}

\usepackage{cite}

\usepackage{nameref,hyperref}

\usepackage{microtype}
\DisableLigatures[f]{encoding = *, family = * }

\usepackage{array}
\usepackage{float}

\newcolumntype{+}{!{\vrule width 2pt}}

\newlength\savedwidth

\usepackage{setspace} 
\raggedright
\usepackage[aboveskip=1pt,labelfont=bf,labelsep=period,justification=raggedright,singlelinecheck=off]{caption}

\makeatletter
\renewcommand{\@biblabel}[1]{\quad#1.}
\makeatother
\usepackage{setspace}
\usepackage{longtable}
\usepackage[export]{adjustbox}
\usepackage{ltablex}
\usepackage{colortbl}
\usepackage{caption}
\usepackage{comment}
\usepackage{pdflscape}
\usepackage{rotating}

\usepackage{lastpage,fancyhdr,graphicx}
\usepackage{epstopdf}
\fancyheadoffset[L]{2.25in}
\fancyfootoffset[L]{2.25in}

\begin{document}
\vspace*{0.2in}

\begin{flushleft}
{\Large
\textbf\newline{How deep is your art: an experimental study on the limits of artistic understanding in a single-task, single-modality neural network} 
}
\newline
\\
Mahan Agha Zahedi\textsuperscript{1\textpilcrow},
Niloofar Gholamrezaei\textsuperscript{2\textpilcrow},
Alex Doboli\textsuperscript{1\textpilcrow},
\\
\bigskip
\textbf{1} Department of Electrical and Computer Engineering, Stony Brook University, Stony Brook, New York, United States of America
\\
\textbf{2} work done when at School of Art, Texas Tech University, Lubbock, Texas, United States of America (now at the Department of Humanities, Regis College, Weston, Massachusetts, United States of America )
\\
\bigskip

%
%
\textpilcrow These authors contributed equally to this work.

* mahan.aghazahedi@stonybrook.edu (MAZ)

\end{flushleft}
\section*{Abstract}
Computational modeling of artwork meaning is  complex and difficult. This is because art interpretation is multidimensional and highly subjective. This paper experimentally investigated the degree to which a state-of-the-art Deep Convolutional Neural Network (DCNN), a popular Machine Learning approach, can correctly distinguish modern conceptual art work into the galleries devised by art curators. Two hypotheses were proposed to state that the DCNN model uses Exhibited Properties for classification, like shape and color, but not Non-Exhibited Properties, such as historical context and artist intention. The two hypotheses were experimentally validated using a methodology designed for this purpose. VGG-11 DCNN pre-trained on ImageNet dataset and discriminatively fine-tuned was trained on handcrafted datasets designed from real-world conceptual photography galleries. Experimental results supported the two hypotheses showing that the DCNN model ignores Non-Exhibited Properties and uses only Exhibited Properties for artwork classification. This work points to current DCNN limitations, which should be addressed by future DNN models. 

\section*{Introduction}

While the study of art has traditionally been the focus of art history, aesthetics, philosophy, psychology and other related areas, advances in Artificial Intelligence (AI) and Machine Learning (ML) have enabled new avenues of inquiry, like devising novel computational models, such as Deep Neural Networks (DNNs), to automatically classify, recognize, and generate artwork \cite{3}. It has been reported that DNNs can identify art genres, artists, and the time range of an art object’s creation \cite{1,2,3,4,8,9}. Applications of these DNN models include helping art curators and historians understand, explore, and navigate through the numerous artworks in museums, galleries, and online sources. Investigating AI/ML models also offers insight on how low-level visual features can lead towards the discovery of high-level semantic knowledge, like image content and object significance, and thus possibly lead to unsupervised knowledge discovery, including tacit knowledge, abstractions, and conceptual reasoning.  

Any attempt to mechanically analyze artwork should reflect the nature of art and how it differs from other types of images. 
This work employed Jerrold Levinson’s philosophy of art as it offers a concrete definition of art, inclusive of both traditional and conceptual works of art. Levinson considers a work of art to incorporate two major properties, Exhibited Properties (EXPs) and Non-Exhibited Properties (NEXPs) \cite{5}. EXPs are the visible elements of art objects, such as color, texture, and form. NEXPs are the ones that are essential artistic aspects of artwork, though they are not simply visible in art objects. NEXPs are accessible by relating art objects and EXPs to human history, culture, and individuals who created the work \cite{6}. Fig 1 summarizes the two kinds of properties.  

\begin{figure}[ht]
    \centering
    \includegraphics[width=0.9\textwidth]{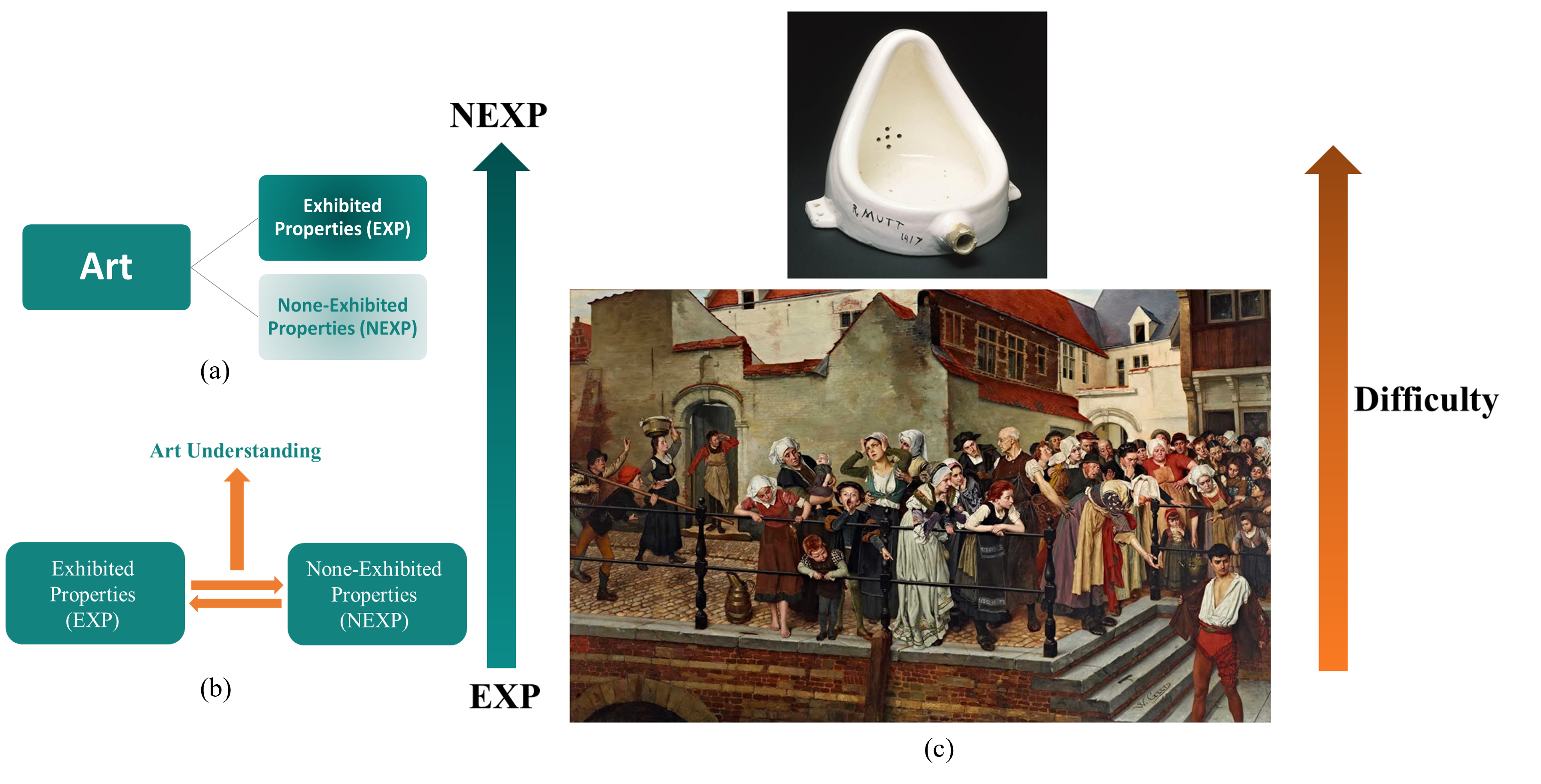}
    \caption{{\bf Art analysis according to Levinson’s definition of the elements of arts.} (a) Art consists of Exhibited Properties (EXPs), illustrated in a darker color, and None-Exhibited Properties (NEXP), depicted in a lighter color. (b) NEXPs are accessed by relating EXPs to the historical discourse of art. (c) The difficulty in art interpretation is shown as a spectrum with an example for each end, top: ``Fountain'' by Marcel Duchamp, and bottom: ``The Accident'' by William Geets.}
\end{figure}

EXPs sometimes directly point to NEXPs but other times they do not. Rather, understanding NEXPs may require complex contextualization and interpretation. Moreover, some artwork contains more EXPs than NEXPs, while some work, particularly artwork identified as conceptual art, is highly loaded with NEXPs. For example, the painting ``The Accident'' by William Geets (1899) (Fig 1(c)-bottom) is a narrative figurative work that 
can be understood to a great extent just by looking at the picture, as it contains more EXPs than NEXPs. In contrast, the famous work ``Fountain'' by Marcel Duchamp (1917) is meaningful mainly based on its NEXPs (Fig 1(c)-top). 
Based on Levinson’s theory, what makes Duchamp’s urinal art, and hence different from other mass-produced urinals, is not its shape, color, or style but the intention of the artist toward the object in relation to the historical discourse of art \cite{7}. 
DNN-based computational methods used for automated art-related activities rely on EXP processing. While EXPs might be sufficient to tackle some art genres, like iconoclasm and medieval European religious art \cite{3}, it is unclear if EXPs are sufficient to identify NEXPs in modern artwork, e.g., intention and historical conditions. A recent model of visual aesthetic experience suggests two parallel, quasi-independent processing modes: bottom-up, perceptual processing universal among all humans (similar to EXP processing) and top-down, cognitive processing that accounts for contextual information, artist intention, and artwork presentation circumstances (similar to NEXP processing)~\cite{0}. As summarized in Section Related Work, previous work suggests that DNN models can gain some insight on artwork meaning (semantics) starting only from EXPs, like color, texture, and shapes~\cite{1, 2, 4, 12, 15}. However, there are no comprehensive studies on the degree to which NEXP recognition can emerge during DNN training using artwork images, and whether such NEXPs are sufficient to distinguish art objects from non-art objects or other artwork, especially in case of conceptual arts. Such studies are important not only to identify and characterize the limitations of DNN models but also to understand if NEXPs of art objects can be sufficiently well distinguished using only their EXPs, thus if an art object is fully specified within its body of similar work, e.g., gallery or art show, or if NEXPs depend to a significant degree on elements not embodied into an art object, like contextual elements, the artist’s intention, and the viewer’s interpretation. 

This paper presents a comprehensive experimental study that tackled understanding the degree to which a state-of-the-art Deep Convolutional Neural Network (DCNN) learns EXPs and NEXPs and then uses the learned knowledge to classify such artworks in the same galleries and exhibitions as artists and curators. 
As discussed in Section Related Work, existing work does not consider the boundary between the AI/ML's perception of a computer image and an image’s interpretation as art. Thus, computational modeling of art is often not grounded in the theory of art. For example, AI/ML models are trained on art images labeled with their styles, genres, or authors but without information about their contexts, intentions, interpretations, or emotions. It is unknown the degree to which AI/ML models, like DCNNs, can automatically pick up these essential details. To address this limitation, this work devised an experimental study integrating semantic and conceptual ideas in aesthetics with AI/ML modeling and experimentation. Given that 
EXPs of art objects are the basis of DCNN model training while NEXPs are likely to be learned, 
the following two hypotheses were defined to experimentally study the importance of EXPs and NEXPs in the automated classification of artwork into different galleries:

{\em Hypothesis I}: Deep Convolutional Neural Network (DCNN) models do not capture NEXPs well for art gallery classification. Therefore, their classification results are not influenced by NEXPs. 

{\em Hypothesis II}: The similarities and differences of EXPs within and between art galleries determine the difficulty level of classification using DCNN models.	

The experimental method devised to study the two hypotheses includes three experiments that used datasets assembled by an art expert. The used DCNN model was the VGG- 11 \cite{10} pre-trained on ImageNet database \cite{11}. The model was then retrained using art images. The datasets designed for the study included images of contemporary photography of diverse style and conceptual orientation. Our art expert chose exhibitions of artists from different countries and photographs that reflect different approaches toward fine art photography, like realism, abstraction, commercial, and conceptual photography. As already discussed, conceptual art, like Duchamp’s ``Fountain'', often represents ordinary ready-made or mass-produced objects (or their photographs) as a work of art, which exclusively relies on the ideas intended towards the artwork rather than artistic style. Hence, the two above hypotheses suggested that including conceptual photography increases the difficulty of classifying a dataset into galleries. A high EXP diversity within a gallery also increases the difficulty level. To further explore
the degree to which the model learns beyond EXPs to classify artwork based on their NEXPs, e.g., concept, ideas, and historical context, the study added a gallery of none-art images of ordinary objects that resembled in their appearance the conceptual fine art photography exhibitions included in the experiment. 
The two hypotheses indicate that the model should show poor performance for the dataset with conceptual photography and non-art images because of the high EXP similarity of their galleries. 
 Experimental results were analyzed using statistical and classification metrics. 
The results of the three experiments confirmed the validity of the two hypotheses. 


The paper has the following structure. Related work was summarized next, followed by the presentation of the experimental methodology. Results and their discussion were described next. The paper ends with conclusions and further research directions.

\section*{Related Work}
Recent work has proposed using modern AI/ML methods for automated analysis of artworks, such as style recognition, classification, and generation \cite{1,2,3,4}. A comprehensive overview paper discusses recent computational and experimental approaches to visual aesthetics, including modern AI/ML methods~\cite{Brachmann2017}. The AI/ML methods often use Deep Convolutional Neural Networks (DCNNs), a DNN type devised for computer image processing \cite{1,2,4,12}. To address the need of large datasets for DCNN training [13], which is often difficult to meet for art objects, the traditional solution is to pre-train a DCNN using large databases of images, e.g., ImageNet, and then retrain only the output and intermediate layers using art images \cite{1,2,4,12,14,15}.  
Fig~2 summarizes the reported using of artwork EXPs and NEXPs for automated art understanding activities.

\begin{figure}[ht]
    \centering
    \includegraphics[scale = 0.8]{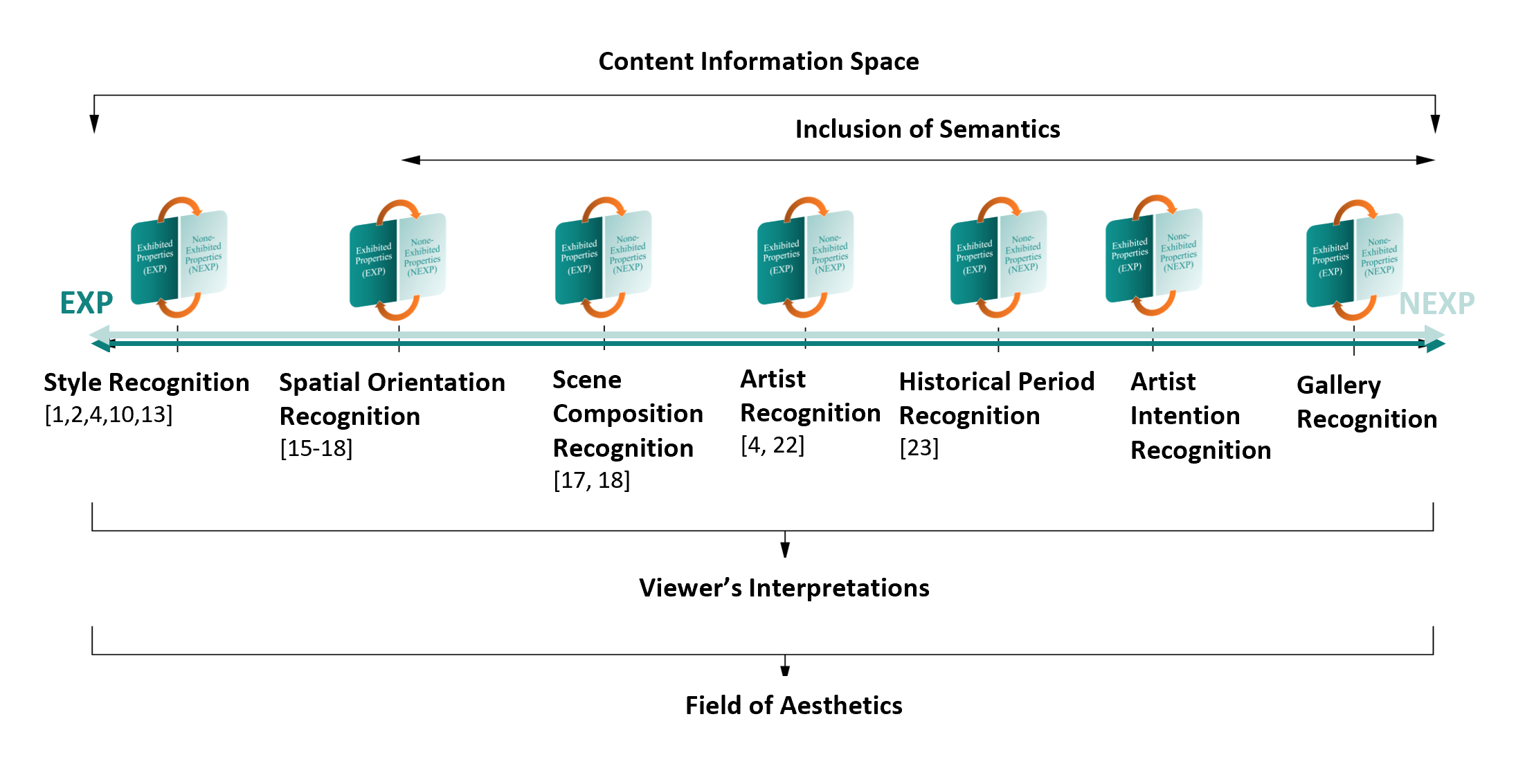}
    \caption{{\bf Automated art understanding.} 
    NEXPs are more critical for an automated activity as more artistic aspects must be considered (i.e. style recognition to gallery recognition).}
\end{figure}

Automated style recognition attempts to identify the artistic style of art objects, like paintings and porcelain objects \cite{1,2,4,12,15}. This work uses EXPs, like color and texture. For example, \cite{23} examines the classification of artistic styles into their respective historical periods based on the ideas of Heinrich Wölfflin, a prominent art historian (1846-1945). Wölfflin explains that different artistic styles reveal their respective historical contexts. Therefore, a machine could classify artwork into historical periods by relying on the stylistic characteristics of artwork \cite{23}.
Different edge orientations are characteristic to traditional artwork from different cultures~\cite{Redies2017}. Statistical differences in image composition are presented between traditional art, bad art, and twentieth century abstract art~\cite{Redies2017b}. Seven DCNN models were tested for three art datasets to classify genres (i.e. landscapes, portraits, abstract paintings, etc.), styles (e.g., Renaissance, Baroque, Impressionism, Cubism, Symbolism, etc.), and artists \cite{1}. Classification uses mostly color information to achieve for some styles a recognition accuracy similar to human experts. However, the authors state that some EXPs, like gestural brushstrokes, can be misleading.        
Also, certain styles are hard to be automatically differentiated from each other, like Post Impressionism and Impressionism, Abstract Expressionism and Art Informel, or Mannerism and Baroque \cite{2}. A dual-path DCNN model recognizes both artistic style and painting content \cite{16}. DCNN are proposed 
to recognize non-traditional art styles too, like Outsider Art style \cite{15}. Adding more features to DCNN training does not improve classification accuracy, which is due, in the author's analysis, to the curse of dimensionality. 

Work on uncovering semantic information about art stems from the goal to understand the content of art objects, including the orientation of an object, the objects in a scene, and the central figures of a scene \cite{10,17,18,19}. Only EXPs are used in this work. Object orientation, e.g., if a painting is correctly displayed, uses low-level features, like simple, local cues, image statistics, and explicit rules \cite{9,17,20,21}. For example, using low-level features to train DNNs has been reported to be as effective as human interpretation across different granularities and styles \cite{17}. The method performs better for portrait paintings than for abstract art, as portraits arguably include more reliable and repetitive cues, which improves DCNN learning. 
Distinguishing image classes seems to focus on localized parts of a few, large objects. Low intra-class variability of the parts is important in a part being selected. Different semantic parts might be selected for objects of related classes, like wheels for cars and windows for buses. Generative Adversarial Networks (GANs) are proposed for hierarchical scene understanding \cite{22}. Analysis shows that the early layers learn physical positions, like the spatial layout and configuration, the intermediate layers detect categorical objects, and the latter layers focus on scene attributes and color schemes. 

DCNN have been also used to recognize an artist that authored an artwork from a group of possible artists by learning the artist-specific, visual features (hence EXPs) of his/her work \cite{4}. During DCNN training, various regions of an art image are occluded, so that the sensitivity of that region for correct classification can be established \cite{4}. 
Experiments suggest that artist recognition uses low-level features, like material textures, color, edges, and the empty areas used to create visual patterns \cite{4}. 
Other work advocates for using intermediate level features, like localized regions, and semantic features, e.g., scene content and composition \cite{8}. 
Performance decreases if the pool of artists from which selection is made increases. 

In summary, current work on using DCNN models for the automated study of art utilizes EXPs, including features (e.g., color, space, texture, form, shape), principles of art (like movement, unity, harmony, variety, balance, contrast, proportion and patterns), and subject topics (i.e. composition, pose, brushstrokes and historical context). DCNNs are suggested to have the capability of unsupervised learning to relate low-level EXPs to higher-level semantics, like objects and object parts, importance of visual cues, hierarchical compositions, and extraction of hidden structures \cite{3,8,17,18,22}. Therefore, DCNNs seem to learn at least some facets of the NEXPs - EXPs relationships, but it is unclear the degree to which this learning happens for situations in which NEXPs are the principal features in deciding the results, like modern artwork inclusion in a gallery. Arguably, the broader theoretical problem is if reductionist approaches can explain NEXPs only based on EXPs of art objects in similar and dissimilar time periods, styles, genres, and galleries, or whether NEXPs cannot be fully understood without considering subjective factors, like artist intention, viewer interpretation, and social context.        

\section*{Methodology}

\begin{figure}[ht]
    \centering
    \includegraphics[scale = 0.6]{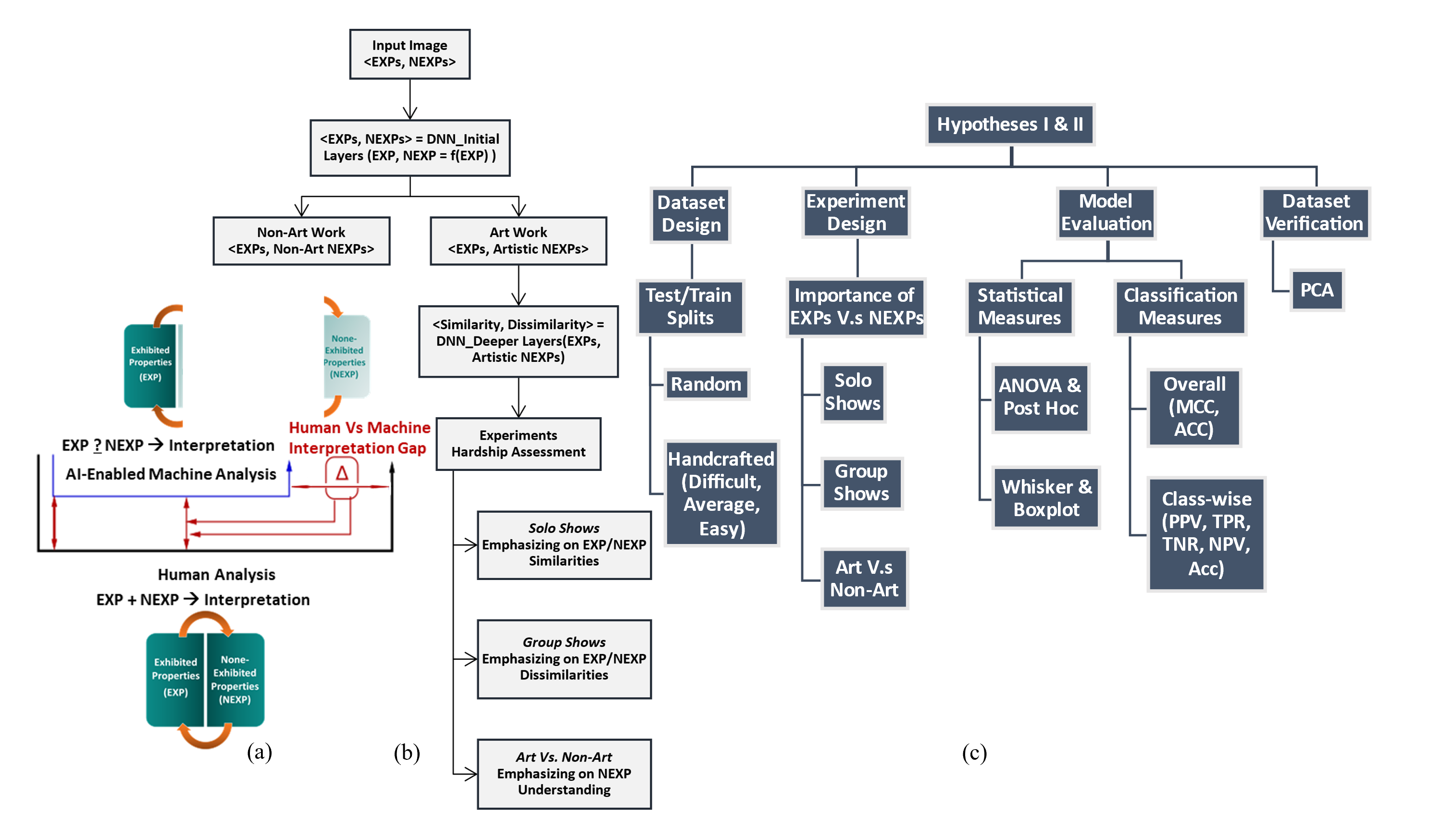}
    \caption{{\bf Methodology summary.} (a) The gap ($\Delta$) between human analysis and  DCNN classification is due to limitations of DCNN model. 
    (b) Dataset design method. (c) The methodology for validating Hypothesis I and Hypothesis II.}
\end{figure}

Three experiments were conducted to assess the validity of the two hypotheses presented in the Introduction section. Fig 3(c) summarizes the experiments. 
The experimental work included devising the required datasets to check the hypotheses for a comprehensive set of cases. Then, the designed datasets were verified with respect to their intended purpose for the study. A trained and fine-tuned DCNN was used to classify the images of the datasets into galleries. The effectiveness of DCNN model was evaluated using statistical measures and classification metrics.   
The gaps (labeled as~$\Delta$ in Fig 3(a)) between artwork classification using DCNN model and artwork understanding by human experts were also studied. 
The components of the experimental study are discussed next.

\subsection*{Dataset Design}

The validity of the two hypotheses stated in the Introduction section was verified by studying how different mixtures of EXPs and NEXPs of artwork determined DCNN classification performance. Two EXP- and NEXP-related factors were expected to set the level of classification difficulty of an art dataset: (1)~the similarity of EXPs between different galleries (of the same dataset) while their NEXPs are different, and (2)~the diversity of EXPs within a single gallery. Fig~3(a) reflects how the similarities and dissimilarities of EXPs and NEXPs decided the expected difficulty level in art classification. Fig.~3(b) summarizes the related parameters. A dataset was considered to be more difficult to classify, if its images from multiple galleries (with different themes, and hence likely with distinct NEXPs) were more similar with respect to their EXPs, and/or if its images in a gallery had a high diversity of EXPs. For instance, two galleries that contain black and white photography look more similar with respect to their EXPs (e.g., their colors). If the two hypotheses are correct, then this similarity makes it harder for the DCNN model to classify the images into their correct gallery, even if these galleries have different NEXPs, i.e. their different historical context and meaning. Moreover, a high diversity of EXPs within a single gallery increases the classification difficulty because it is harder for the DCNN model to find common EXPs to identify the images that belong to the same gallery, even if the gallery’s artworks have similar NEXPs. This observation is especially evident for group exhibitions, since they were curated around similar themes and meanings but they contain artworks that look different from each other as distinct artists created them. Hence, separate datasets studied the using of EXPs and NEXPs in solo and group galleries for classification. 

\begin{figure}[ht]
    \centering
    \includegraphics[scale = 0.5]{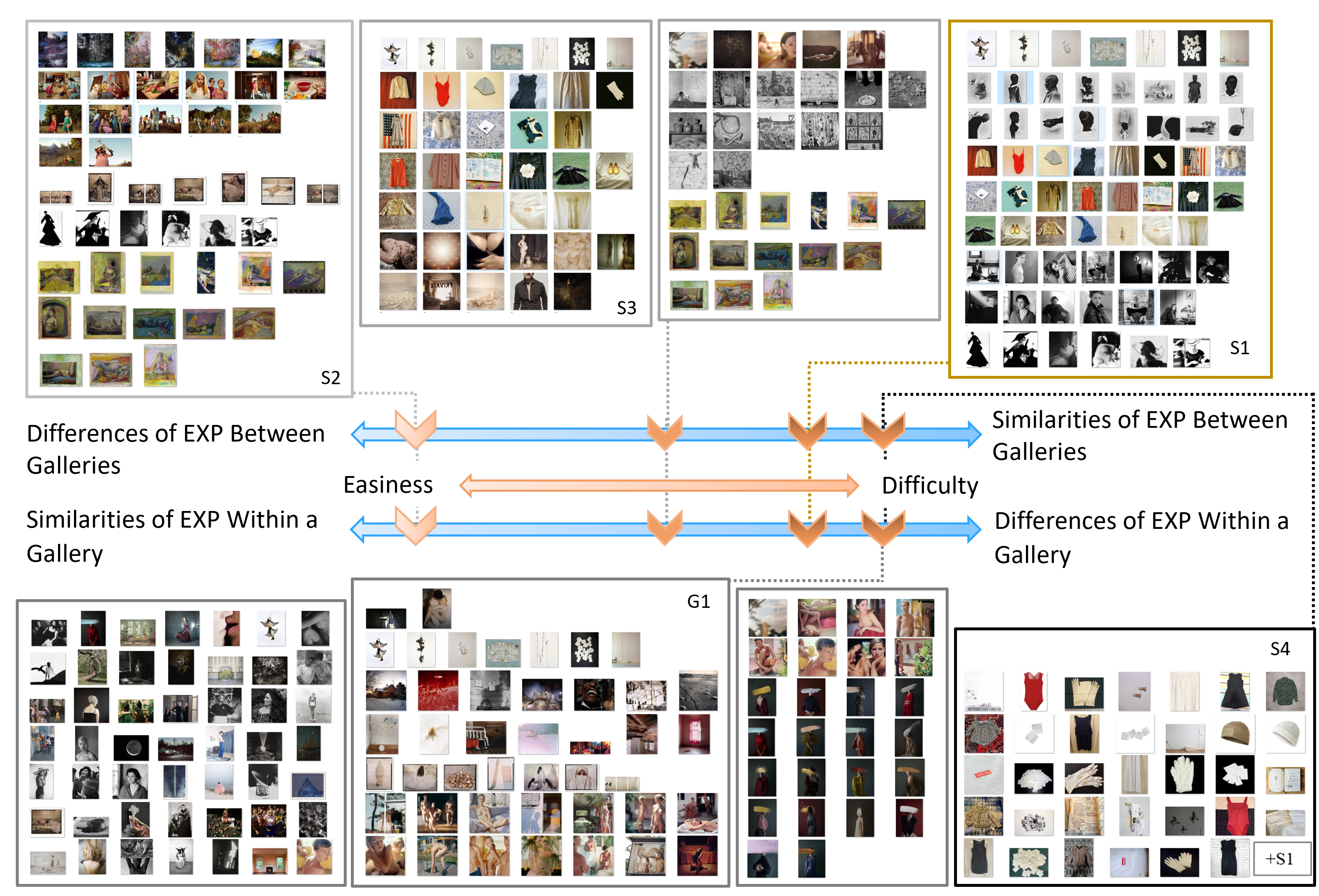}
    \caption{{\bf Dataset summary.} The designed datasets S1, S2, S3, SF1-4, S4 and G1 pertain to a broad difficulty spectrum according to their EXPs and NEXPs dissimilarities and similarities within or between galleries.}
    \label{fig4}
\end{figure}

Fig~3(b) and Fig~4 summarize the datasets utilized in training and testing the DCNN classification model. 
The datasets comprehensively covered artwork EXPs and NEXPs to examine the degree to which the model learns and uses NEXPs to classify images into their correct galleries. Second, insights into the limits of DCNN classifiaction as compared to human art experts were also explored. The assembled datasets were input to DCNN model with random and handpicked test/train splits to create difficult, average, and easy classification situations. These splits targeted different mixtures of EXPs and NEXPs sets. Four subsets, called {\em difficult}, {\em average}, {\em easy}, and {\em random} were created for each datasets. Third, to further describe the model’s ability to use NEXPs, a dataset was assembled to test  the capacity of the model to separate non-art images (with no NEXPs) from artwork (with NEXPs). Fourth, to observe the degree to which the dataset size influences classification results, the devised datasets included galleries of different sizes, e.g., number of gallery images.



The considered art galleries were chosen from existing online exhibitions curated by established art curators, and not by the art expert that participated to our study. 
To limit the scope of the datasets, we picked artwork that uses photography as an underlying medium, so that the galleries reflect diverse approaches towards fine arts photography in contemporary art.
The artwork pursues different artistic attitudes, like highly conceptual, representational, and abstract. Certain art objects mix photography with paint, some used digital, and others are analog photography. 
We also included group exhibitions, as group exhibitions are curated around a common theme or concept but are vastly different in their styles and formal features. 
Table~\ref{t0} summarizes the characteristics of the galleries in Dataset~S1 (see the Appendix for the other datasets). Table~\ref{t1} details the EXPs and Table~\ref{t2} the NEXPs of the galleries in this dataset. Table~\ref{t3} shows the outliers of the galleries, such as the images that are different from the rest. 

\begin{table}[ht]
\setlength{\tabcolsep}{8pt}
\renewcommand{\arraystretch}{1.8}
\captionsetup{margin=0in}
\caption{\bf Characteristics of Dataset S1:} the related galleries (e.g., names of the exhibitions or shows), the number of images per gallery, and the total number of images per dataset.
\label{t0}
\small
\begin{tabular}{p{1.2cm}|p{3.8cm}|p{2cm}|p{1.2cm}}
\rowcolor{lightgray!50}
\bf\normalsize Dataset & \bf\normalsize Related galleries & \bf\normalsize  Number of Images & \bf\normalsize  Total \\
S1 & {\emph {Heat + High Fashion \newline  Mukono \newline  My Mother’s Clothes \newline Scene \newline Trigger}} & 6 \newline 16 \newline 22 \newline 13 \newline 7 & 64 \\ 
\end{tabular}
\end{table}

\setlength{\LTleft}{-2.25in}
\setlength{\LTright}{0in}
\setlength{\tabcolsep}{7pt}
\renewcommand{\arraystretch}{1.7}
\small
\begin{singlespace}
\begin{longtable}{p{0.1\linewidth}|llll}
\captionsetup{margin=-2.25in}
\caption{\bf EXPs of the galleries in Dataset S1}
\label{t1} \\
\rowcolor{lightgray}
\multirow{1}{*}{\bf\normalsize Gallery} & \multicolumn{4}{c}{\bf\normalsize EXPs}  \\ 
\rowcolor{lightgray!50} 
\cellcolor{lightgray} & \multicolumn{1}{l|}{\bf\normalsize Medium, Color} & \multicolumn{1}{p{2.1cm}|}{\bf\normalsize Shape, Form,\newline Texture } & \multicolumn{1}{l|}{\bf\normalsize Composition$^1$} & \multicolumn{1}{l}{\bf\normalsize Subject Matter$^2$} \\ 

{\emph {Heat +\newline  High \newline Fashion}} & \multicolumn{1}{p{4.4cm}|}{- Black and white photography\newline - Monochromatic \newline - Dark value\newline - Light value \newline - Midtones  \newline - High contrast} & \multicolumn{1}{p{2.1cm}|}{- Figurative \newline - Organic \newline - Plain \newline - Open forms \newline - Painterly}  & \multicolumn{1}{p{5.2cm}|}{- Closed compositions \newline - Open composition \newline - Tendency toward symmetry \newline - A-symmetrical \newline - Centered \newline - Alignment of the  subject matter\newline - Horizontal frames \newline - Vertical frames \newline - Emphasis on a single subject matter \newline - Empty and quiet composition \newline - Busy and crowded compositions}  &  \multicolumn{1}{p{3.1cm}}{- Female body \newline - Female torso \newline - Portraits \newline - Hidden human faces \newline - Fashion \newline - Interior space \newline - Shadows, reflections} \\ \hline

{\emph {Mukono}} & \multicolumn{1}{p{4.4cm}|}{- Black and white photography\newline - Monochromatic\newline - Dark value\newline - Light value \newline - Midtones \newline - High contrast \newline - Medium contrast \newline - Low contrast \newline - Low saturation \newline - Neutral colors}   & \multicolumn{1}{p{2.1cm}|}{- Figurative \newline - Organic \newline - Plain \newline - Open forms \newline - Closed forms} & \multicolumn{1}{p{5.2cm}|}{- Closed composition \newline - Tendency toward symmetry \newline - Asymmetrical \newline - Centered alignment of the subject matter \newline - Horizontal frame \newline - Vertical frame \newline - Emphasis on a single subject matter \newline - Empty and quiet composition} & \multicolumn{1}{p{3.1cm}}{- Human torso/male torso \newline - Female torso \newline - Portraits \newline - Hidden human faces \newline - Nature/landscape \newline - Human body \newline - Everyday objects \newline - Still life \newline - Animals} \\ \hline

{\emph {My Mother’s Clothes}} & \multicolumn{1}{p{4.4cm}|}{- Color photography \newline - High saturation \newline - Medium saturation\newline - Low saturation\newline - Cool colors\newline - Warm colors \newline - Neutral colors\newline - Dark value\newline - Light value \newline - Midtones  \newline - High contrast \newline - Low contrast\newline - Medium contrast\newline - Chromatic} & \multicolumn{1}{p{2.1cm}|}{- Organic \newline - Geometric \newline - Textured \newline - Plain \newline - Open form \newline - Linear \newline - Closed form \newline - Decorative \newline - Pattern \newline - Floral \newline - Text } & \multicolumn{1}{p{5.2cm}|}{- Closed composition \newline - Tendency toward symmetry \newline - Asymmetrical \newline - Centered alignment of the subject matter \newline - Square frames \newline - Emphasis on a single subject matter \newline - Busy and crowded compositions \newline - Empty and quiet composition} & \multicolumn{1}{p{3.1cm}}{- Female clothes \newline - Still life \newline - Everyday objects \newline - Domestic space} \\ \hline

{\emph {Scene}} & \multicolumn{1}{p{4.4cm}|}{- Black and white photography\newline - Monochromatic \newline - Dark value \newline - Light value \newline - Midtones \newline - High contrast \newline - Medium contrast} & \multicolumn{1}{p{2.1cm}|}{- Figurative \newline - Textured \newline - Plain \newline - Open form \newline - Closed-form\newline - Pattern} & \multicolumn{1}{p{5.2cm}|}{- Closed composition \newline - Tendency toward symmetry \newline - Asymmetrical \newline - Centered alignment of the subject matter \newline - Square frame \newline - Emphasis on a single subject matter} & \multicolumn{1}{p{3.1cm}}{- Human body  \newline - Female body \newline - Male body \newline - Human torso \newline - Male torso \newline - Female torso \newline - Interior space \newline - Shadows, reflections \newline - Artists} \\ \hline

{\emph {Trigger}} & \multicolumn{1}{p{4.4cm}|}{- Color photography\newline - Medium saturation \newline - Low saturation \newline - Neutral colors \newline - Cool colors \newline - Dark value \newline - Light value \newline - Mid-tones  \newline - High contrast \newline - Low contrast \newline - Medium contrast \newline - Chromatic} & \multicolumn{1}{p{2.1cm}|}{- Plain \newline - Organic\newline - Geometric \newline - Closed form}  & \multicolumn{1}{p{5.2cm}|}{- Closed composition \newline - Tendency toward symmetry \newline - Asymmetrical \newline - Centered alignment of  the subject matter \newline - Horizontal frame \newline - Vertical frame \newline - Emphasis on a single subject matter \newline - Empty and quiet composition} &  \multicolumn{1}{p{3.1cm}}{- Interior space \newline - Domestic space \newline - Still life  \newline - Animals}  \\ \hline
\end{longtable}
\begin{adjustwidth}{-2.25in}{0in}
\noindent\footnotesize{\(^1\)}Arrangements of visual elements within a frame \newline
\footnotesize{\(^2\)}What we are looking at\newline
\end{adjustwidth}
\newpage
\small
\setlength{\tabcolsep}{5pt}
\renewcommand{\arraystretch}{1.7}
\begin{longtable}{p{0.1\linewidth}|llll}
\captionsetup{margin=-2.25in}
\caption{\bf NEXPs of the galleries in Dataset~S1}
\label{t2} \\
\rowcolor{lightgray}
\multirow{1}{*}{\bf\normalsize Gallery} & \multicolumn{3}{c}{\bf\normalsize NEXPs}  \\ 
\rowcolor{lightgray!50} 
\cellcolor{lightgray} & \multicolumn{1}{l|}{\bf\normalsize Context$^3$} & \multicolumn{1}{l|}{\bf\normalsize Intention$^4$} & \multicolumn{1}{l}{\bf\normalsize Meaning} \\

{\emph {Heat +\newline  High \newline Fashion}} & \multicolumn{1}{p{3.4cm}|}{- Modern and contemporary \newline - New York “bohemian.”} & \multicolumn{1}{p{6.5cm}|}{- Engaging female fashion through photography \newline - Realism yet a sense of ambiguity through blurred images and painterly qualities of the medium of photography (in that sense, it can place itself against modernist photography and its notion of medium specificity)} & \multicolumn{1}{p{6.5cm}}{- Photographs of  dancer Isadora Duncan \newline - Fashion photography that reflects the time and historical context \newline - Female identity through fashion and clothing \newline - Realism yet a sense of ambiguity through blurred images and painterly qualities of the medium of photography}  \\ \hline

{\emph {Mukono}}  & \multicolumn{1}{p{3.4cm}|}{- Contemporary photography\newline - Cultural studies}  & \multicolumn{1}{p{6cm}|}{- Realism and documentary photography} & \multicolumn{1}{p{6.5cm}}{- Documenting people around the world \newline - Race, ethnicity, culture-racial and cultural identity}  \\ \hline

{\emph {My Mother’s Clothes}}  & \multicolumn{1}{p{3.4cm}|}{- Contemporary photography conceptual art (using ready-made objects as works of art)} & \multicolumn{1}{p{6.5cm}|}{- Conceptual photography inspired by conceptual art and the use of readymade/ordinary objects and blurring the boundary between art and life \newline -Photographing her mothers’ clothes and personal items as a form of portraits or chronology of her mother’s life \cite{51} \newline - Using art and photography to cope with the loss of her mother and her mother’s suffering from Alzheimer\cite{51}} & \multicolumn{1}{p{6.5cm}}{- Her mother’s clothes and personal items as her mother’s portrait/body= clothes as a metonymy of the person \newline - Remembering the past, memories of her mother, perhaps a sense of nostalgia \newline - Coping with Truma of loss of her mother and her suffering from Alzheimer \newline - Gender expression/identity \newline - Social class in America} \\  \hline

{\emph {Scene}} & \multicolumn{1}{p{3.4cm}|}{- 1960s underground\newline/avant-garde artists’\newline scenes in the United States} & \multicolumn{1}{p{6.5cm}|}{- Realistic photographs of Avant-garde artists in New York during the 1960s \newline - Documentary photography} & \multicolumn{1}{p{6.5cm}}{- Photography and realism \newline - Indexicality \newline - Representation of avant-garde artists in NYC during the c1960s \newline - Human emotion and psychological expression}  \\ \hline

{\emph {Trigger}} & \multicolumn{1}{p{3.4cm}|}{- Contemporary photography \newline - Conceptual photography} & \multicolumn{1}{p{6.5cm}|}{- Conceptual photography \newline - Photographing everyday objects and domestic space (her hometown) \newline - Capturing time passing through photography} & \multicolumn{1}{p{6.5cm}}{- Artist’ hometown and the lives of people who has lived there \cite{50}. \newline - Passage of time \cite{50} \newline - Her personal experiences \cite{50}\newline - Collision of past and present \cite{50}\newline - Domestic space and meaningful—perhaps personal everyday objects \cite{50}\newline - Time, temporality}  \\ \hline
\end{longtable}
\begin{adjustwidth}{-2.25in}{0in}
\noindent\footnotesize{\(^3\)}Historical, social, political, cultural conditions in which the work is created \newline
\footnotesize{\(^4\)} Intention refers to artists' intention to make a work of art with a meaningful connection to previous works of art and the history of art.  \newline
\end{adjustwidth}
\normalsize
\small
\setlength{\tabcolsep}{5pt}
\renewcommand{\arraystretch}{1.7}
\begin{longtable}{p{0.1\linewidth}|p{7cm} p{0.5cm}p{0.1\linewidth}|p{7cm}}
\captionsetup{margin=-2.25in}
\caption{\bf Gallery outliers for Dataset~S1. Galleries \emph{Heat + High Fashion}, \emph{My Mother’s Clothes}, and \emph{Scene} \newline do not have any outlier.}
\label{t3} \\
\rowcolor{lightgray}
\multirow{1}{*}{\bf\normalsize Gallery} & {\bf\normalsize Outlier \(^5\)} & \cellcolor{white} & {\bf\normalsize Gallery} & {\bf\normalsize Outlier} \\ \cline{1-2}\cline{4-5} 

{\emph {Mukono}}& Most images in this gallery are human portraits with a high contrast between the figure and the background. However, these images have different features from those.   
    \begin{center}
       \includegraphics[ width=0.8\linewidth,valign=t]{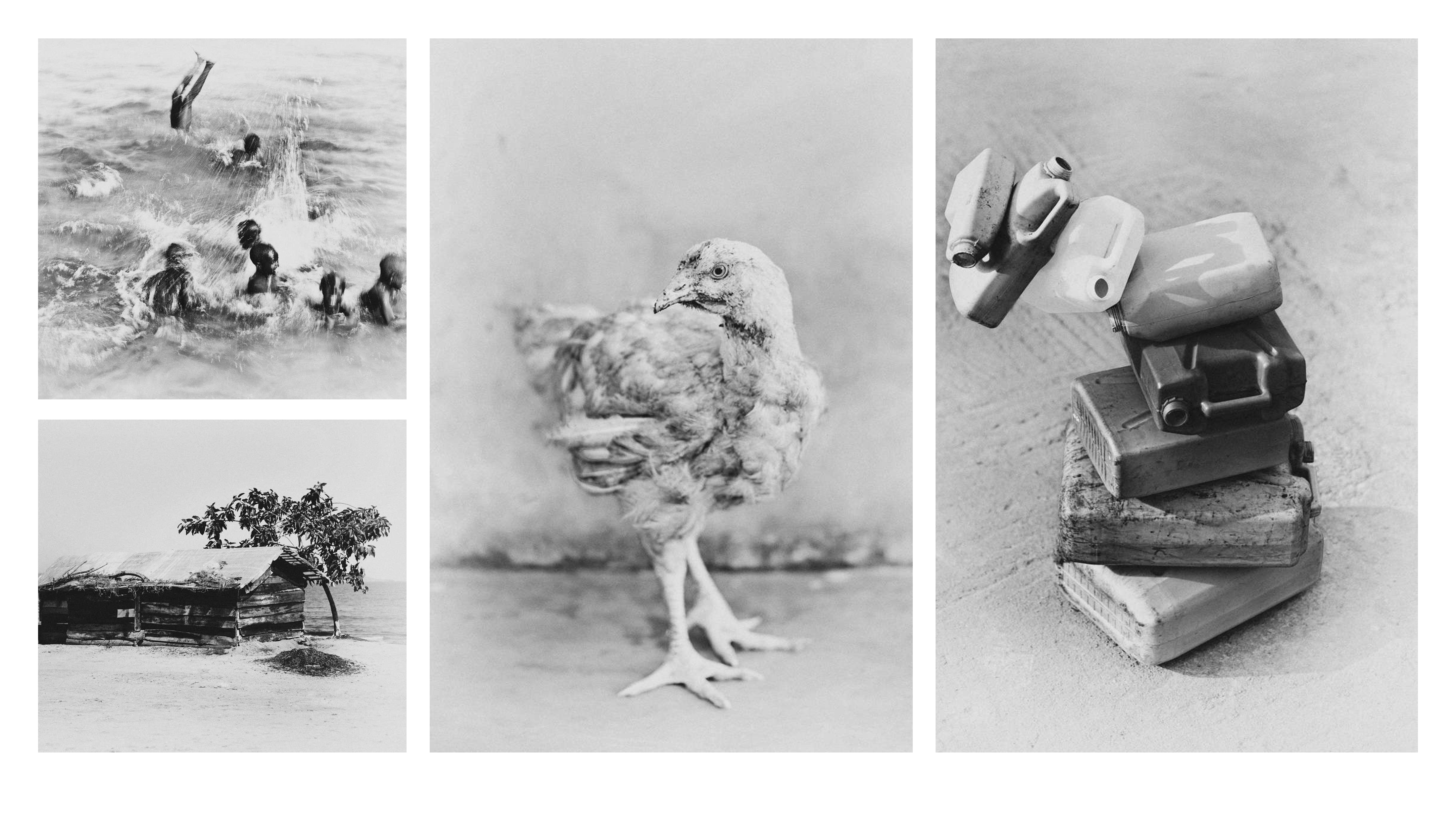}
    \end{center} & &

{\emph {Trigger}}& Each image in this gallery has distinct characteristics. It is hard to find an outlier due to the high diversity within the gallery. However, this image is slightly more different from the rest as it is different in terms of its subject and composition. While the other images in the gallery are close to a symmetrical arrangement, the composition in this image is asymmetrical. 
    \begin{center}
      \includegraphics[width=0.4\linewidth,valign=t]{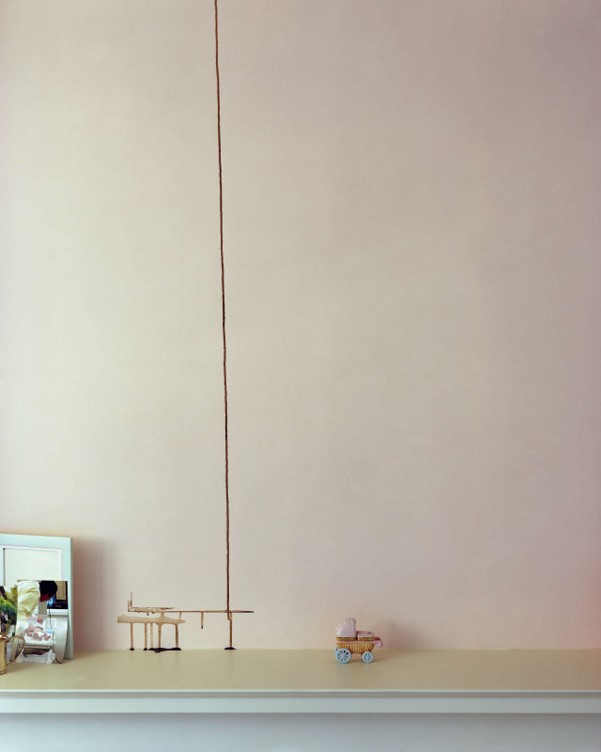}  
    \end{center}
\\ \cline{1-2}\cline{4-5}

\end{longtable}
\end{singlespace}
\begin{adjustwidth}{-2.25in}{0in}
\noindent\footnotesize{\(^5\)}An image that is different from the rest due to multiple differences or one major distinction.
\end{adjustwidth}
\normalsize

\subsection*{Design Verification}

Principal Component Analysis (PCA) was used to visualize the designed datasets to be  verified with respect to their purpose for the experimental study. 
Each image in a dataset underwent the same pre-processing as the images used with the DCNN model (except for data augmentation transformations), such as to resize and center-crop to a 224 $\times$ 224-pixel image in RGB (color images) and L (grayscale images) color space. 
The first three principal components of each image were then plotted in a 3D scatter plot. Each gallery was shown using a separate color. The formation of distinct clusters with points of the same color indicates the success of image classification based on the chosen features. Random placement of points of the same color indicates the opposite.  


\subsection*{DCNN Model}

The three experiments used a VGG-11 DCNN model~\cite{10} pre-trained on ImageNet dataset~\cite{11} and discriminatively fine-tuned \cite{24}. The model came from the PyTorch framework. 
To avoid overfitting, batch normalization \cite{26} was used as a regularization technique along with data augmentation using methods {\em Random Rotation}, {\em Random Horizontal Flip}, and {\em Random Crop with Padding} from Torchvision library~\cite{torchvision2016}. The learning rate obtained by method {\em Cyclical Learning Rates}~\cite{25} for the transferred features was an order of magnitude lower than that of the output classifier layer. DCNN model training utilized Adam Optimizer \cite{27} with a cross-entropy loss function.

\subsection*{Model Evaluation}

For comprehensive evaluation without uncertainty and with the least possible overstatement of the results~\cite{28}, the trained DCNN model was tested using statistical measures and classification metrics. Statistical measures assessed the model behavior in response to the intended experimental setup. Unimodal distributions indicate if the model has a single behavior, and multimodal distributions show if the model responds in multiple ways or if there are undetected, underlying variables present. Comparison of means through ANOVA tests can identify interpretable characteristics of the model’s responses to the designed inputs. Classification metrics evaluate a model’s response to a specific task,
and can facilitate an unbiased and more balanced assessment of the DCNN model.

{\bf Statistical measures}. A total of 1400 trials were collected for each experiment to sample the space of possibilities, and unique seeds to torch and all other random processes were used to maximize the likelihood of finding outliers. 
A one-way ANOVA test with a confidence interval of 0.99 ($\alpha$ = 0.01) was performed as the primary measure for statistical comparison of the model’s overall performance. Although ANOVA test is robust to the non-normality of the distribution and to some degrees of the heterogeneity of variances with equal sample sizes \cite{29,30}, we still performed Leven’s test of homogeneity of variances \cite{31} and Shapiro-Wilk normality test~\cite{31}, and visually verified these assumptions by assessing the histograms and normal Q-Q plots. For large sample sizes, like ours, minuscule derivations from normality can be flagged as statistically significant by parametric tests \cite{33,34,35}, hence the need to visually inspect the distributions. To pinpoint the different pairs and to consider the deviations from normal distributions of homogeneous variance when using ANOVA test for comparison of means, Games-Howell (nonparametric) \cite{36} and Dunnett’s T3 (parametric) \cite{37} post-hoc tests were carried out to account for the violation of homoscedasticity or equality of variances, and Tukey’s test \cite{38} for controlling Type~I error, or the likelihood of an incorrect rejection of the hypothesis. All statistical tests were performed by SPSS and plots were created by seaborn API.

{\bf Classification metrics}.
Eight class-wise measures were computed to observe the DCNN model’s performance for each art gallery: Positive Predictive Value (PPV, Precision), True Predictive Rate (TPR, Recall, Sensitivity), True Negative Rate (TNR, Specificity), Negative Predictive Value (NPV), False Positive Rate (FPR), False Negative Rate (FNR), False Discovery Rate (FDR), and Class-wise Accuracy (Acc). In addition to the overall accuracy (ACC), Matthew’s Correlation Coefficient (MCC) was utilized to avoid overemphasized (inflated) results~\cite{39}. For simplicity and without information loss, only measures primarily evaluating True values (True Positive, True Negative), PPV, TPR, TNR and NPV were used to analyze the DCNN model’s performance. The full report of the discussed statistical measures and plotted results were presented in the supporting materials section.

 \section*{Experiments}

\subsection*{Experiment I. The importance of EXPs Vs. NEXPs in classification of solo shows}

{\bf Dataset S1 description}. 
As summarized in Table~\ref{t1} and Table~\ref{t2}, this dataset was designed by our art expert to have the most dissimilar NEXPs and the most similar EXPs among its galleries as compared to datasets S2 and S3. If Hypotheses~I and II were true, then the high diversity of NEXPs (Hypotheses~I) and high similarity of EXPs among galleries (Hypotheses~II) would result in a poor performance of the DCNN model in classifying the artwork images into their correct galleries. Strong classification performance rejects at least one of the hypotheses.

{\bf Results}. The PCA 3D plots in Fig 5(a) depict the similarity of the data points (e.g., art images) of the three subsets, {\em difficult}, {\em average}, and {\em easy}. Different colors indicate different galleries. Moving from subset~{\em difficult} to subset~{\em easy}, the plots showed the gradual formation of single-color clusters of points, with fewer occlusions and mixtures of images from different galleries. However, there were no fully homogeneous clusters. The PCA plots, with three principal components variance of 79.5\%, 67.82\%, and 54.44\% for the three subsets, validate their purpose with respect to studying the impact of the training/test sets on the DCNN performance.

\begin{figure}[ht]
    \centering
    \includegraphics[scale=0.6]{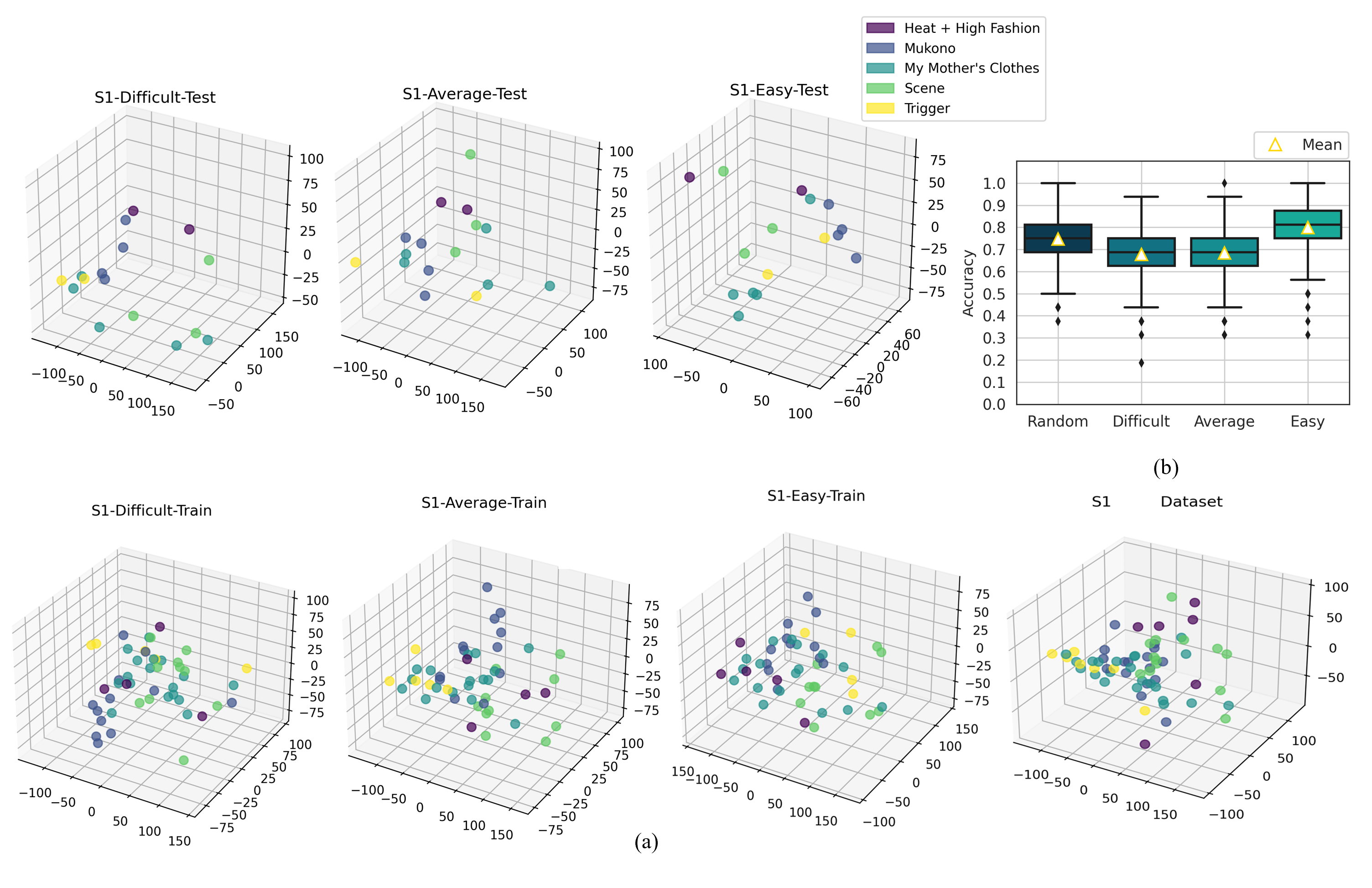}
    \caption{{\bf Results for Dataset S1.} (a) PCA plots for the training and test data of subsets {\em difficult}, {\em average}, and {\em easy}. (b) Box and whisker plots of the overall accuracies (ACC) of the DCNN classification of the four subsets. Subsets {\em difficult} and {\em average} have outliers of 20\% and 100\%. 
    }
\end{figure}

The performance results obtained for classifying Dataset S1 into galleries using the DCNN model were as follows. ANOVA post-hoc tests indicated that there was no statistically significant difference between the ACC of subsets~{\em difficult} and {\em average} (p\textsubscript{ Tukey HSD} = 0.236, 99\% C.I = [-0.0049, 0.0199], p\textsubscript{ Dunnett T3} = 0.374, 99\% C.I = [-0.0057, 0.0207], p\textsubscript{ Games-Howell} = 0.283, 99\% C.I = [-0.0056, 0.0206]), even though the box plots in Fig 5(b) and the bar plots of the class-wise metrics in Fig 6 suggested otherwise. ANOVA post-hoc tests for the rest of the subset pairs showed a statistically significant difference (p $<$ 0.01 for all tests). Hence, the DCNN model had a distinct behaviour for each of the four subsets.  

\begin{figure}[ht]
    \centering
    \includegraphics[scale=0.9]{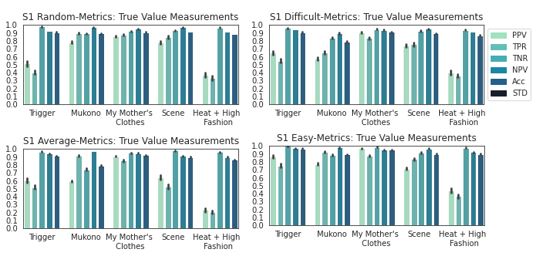}
    \caption{\bf Class-wise metrics results for Dataset S1}
\end{figure}

Next, the capacity of the DCNN model to correctly identify a specific gallery was studied for the four subsets of Dataset~S1. The class-wise metrics in Fig~6 displayed distinct distributions for the four subsets. The closeness of MCC averages (i.e. 0.69, 0.61, 0.62, and 0.76) and of ACC averages (e.g., 0.75, 0.68, 0.68, and 0.80) for subsets {\em random}, {\em difficult}, {\em average}, and {\em easy} suggest a very low possibility of random assignment of gallery labels by the model. Also, the closeness of MCC averages and ACC averages in addition to the ascending trends also support the intention about the desired difficulty levels of the four subsets.
However, metric Acc was not reliable in understanding the model's ability to find the correct gallery labels, as it barely changed for any gallery. More insight was obtained by analyzing the other metrics. As NPV and TNR measure True Negatives (TNs), their relatively continuous high values suggest that DCNN model was quite successful in differentiating galleries, such as indicating that a certain artwork image is not part of a gallery. PPV and TPR measure True Positives (TP), hence the DCNN's ability to identify the correct gallery of an artwork image. As Fig~6 indicates, PPV and TPR were consistently low for galleries ``Trigger'' and ``Heat + High Fashion'', and low for subsets {\em difficult} and {\em average} for galleries ``Scene'' and ``Mukono''. PPV and TPR were consistently high only for Gallery~``My Mother's Clothes'' and for subsets {\em random} of galleries ``Mukono'' and ``Scene''. Hence, with the exception of one gallery, ``My Mother's Clothes'', DCNN model struggled with but finding the correct gallery of the artwork images. An additional experiment was performed to clarify whether the high performance obtained for Gallery ``My Mother's Clothes'' was due to EXPs or NEXPs being learned by the model. The experiment, called Experiment~4 is discussed at the end of this section. In this experiment, additional images were added as part of the new gallery ``Non-Art'', so that these images were very similar in their EXPs with galleries ``Trigger'' and ``My Mother’s Clothes'' but had no NEXPs, as they did not represent artwork. As shown in Figure~\ref{f14}, the new gallery worsened the classification performance, which suggests that the DCNN model did not learn NEXPs but was negatively affected by the increased EXP similarity between distinct galleries. In conclusion, the experiments using Dataset~S1 confirmed the two hypotheses.

A detailed analysis was then performed by the art expert on the classification results to understand how NEXPs and EXPs influenced the performance. Even though Gallery ``Trigger'' was expected to be the hardest to classify among all galleries, DCNN results showed that it was the second hardest to classify. Instead, Gallery ``Heat+ High Fashion'' had the lowest performance. This is likely because it is one of the three grayscale galleries. It shares similar EXPs with Gallery ``Scene'', and its size is slightly smaller than the other two galleries (see the supporting materials section). Assuming that the model learned NEXPs and used them for classification, distinguishing the three grayscale galleries with dissimilar NEXPs (due to their historical differences) would be easier. However, this situation was not observed.
As summarized in Table~\ref{t1}, Gallery ``My Mother's Clothes'' had the most the diverse EXPs as compared to the other galleries (and the most similar EXPs within the gallery), which explains the high performance in correctly finding the gallery for its images. 
For the situations with high TP performance, the values were similar for subsets {\em random} and {\em easy}. Thus, randomly selecting training images for these situations is likely to include sufficient features to support a relatively correct classification, such as having enough diverse EXPs, as estimated by our art expert. Moreover, DCNN performance is less linked to NEXP diversity. 

Cross-referencing PCA and class-wise metrics showed that inter-class similarities of principal components represented by Euclidian distance in a 3D~space pose more challenges than within-class dissimilarities. This was observed in three instances for Dataset~S1: (i)~The performance for Gallery ``Heat+ High Fashion'' was the lowest, even though two test images in subset {\em average} were close to each another. This is likely because other classes’ datapoints were concentrated nearby. (ii)~The low performance for Gallery ``Trigger'' is due to the occluded points in subset {\em difficult} and the distant points in subset {\em average}. 
The best performance was obtained for subset {\em easy}, as there were no points from other classes present between the two test image datapoints. (iii)~Gallery ``Scene'' ‘s performance was the lowest for subset {\em average} and about similarly high for the other subsets. This was likely due to the placement in close proximity of its three test images as well as the closeness of subset {\em average}'s points to other galleries. The gallery size was not critical on its own in setting the difficulty level of a gallery,  however, it biased the classifier in some cases towards the larger galleries.

{\bf Dataset S2 description}. Dataset S2 was designed to have the most dissimilar EXPs and most similar NEXPs between galleries as compared to datasets S1 and S3. If Hypotheses~I and II were true, then the high similarity of NEXPs (Hypotheses~I) and high diversity of EXPs between galleries (Hypotheses~II) would result in strong performance of the DCNN model in classifying the artwork images into their correct galleries. Poor classification performance rejects at least one of the hypotheses.

{\bf Results}. PCA 3D plots in Fig 7(a) illustrate the similarity of the datapoints (e.g., art images) of the three subsets. As observed for Dataset~S1 too, clusters of points of the same color (hence, pertaining to the same gallery) were observed as the DCNN model classified subsets~{\em difficult}, {\em average}, and then~{\em easy}. There were multiple instances of fully homogeneous clusters. 
PCA plots, with three principal components variance of 61.18\%, 60.63\%, and 64.06\% for the three subsets, validate their purpose with respect to studying the impact of the training/test sets on the DCNN performance.

\begin{figure}[ht]
    \centering
    \includegraphics[scale=0.6]{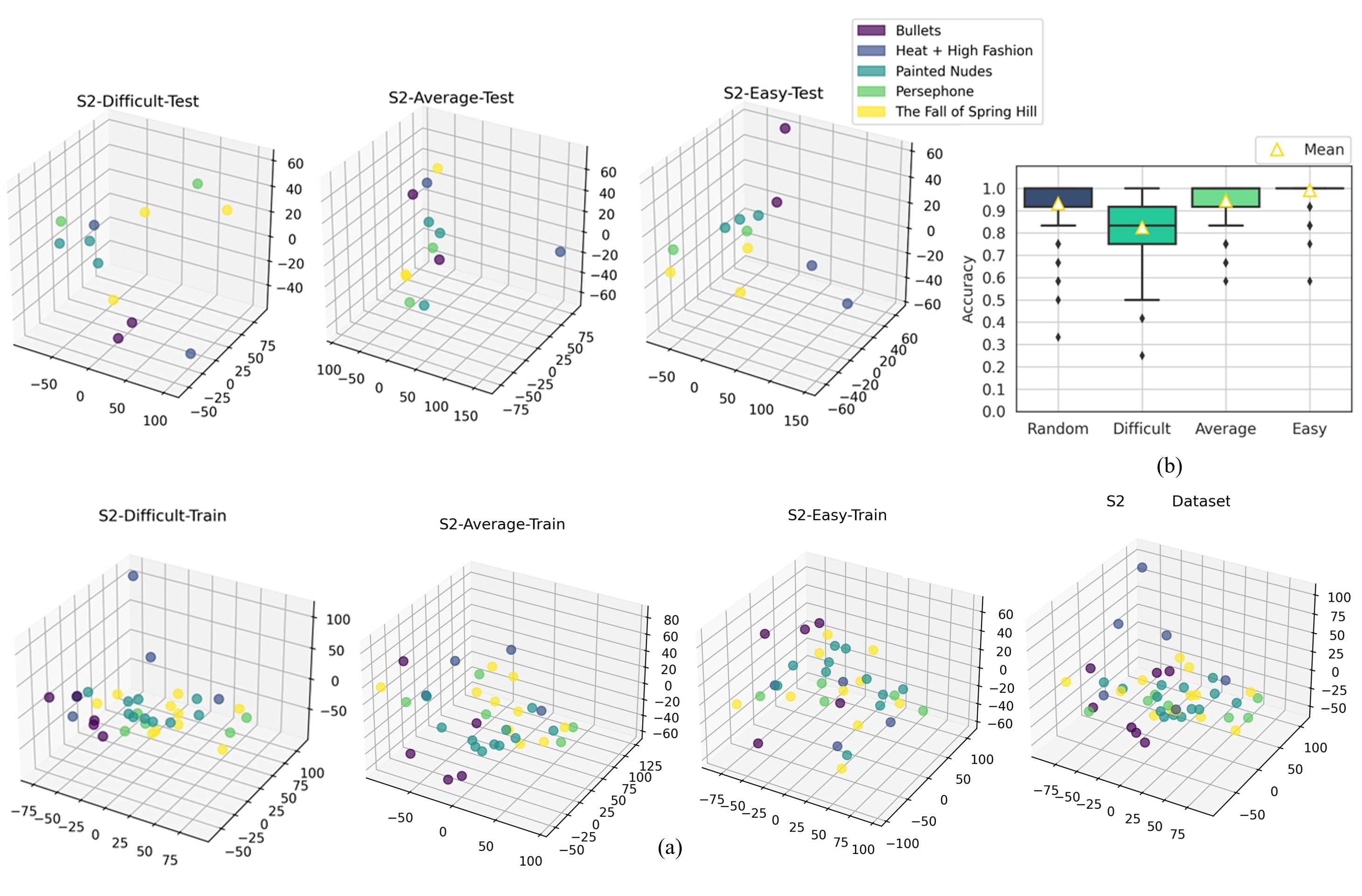}
    \caption{{\bf Results for Dataset S2.} (a) PCA plots for the training and test data of subsets {\em difficult}, {\em average}, and {\em easy}. (b) Box and whisker plots of the overall accuracies (ACC) of the DCNN classification of the four subsets. 
    }
\end{figure}

The resulting performance for classifying Dataset S2 into galleries using the DCNN model were discussed next. ANOVA post-hoc tests showed a statistically significant difference between the ACC values of all subsets (p $<$ 0.01 for all tests). Although box plots of subsets {\em random} and {\em average} in Fig 7(b) are almost identical, the two subsets are still distinguishable through their outliers. This is supported by their class-wise metrics in Fig 8 too. Hence, the DCNN model had a distinct behavior for each of the four subsets.

\begin{figure}[ht]
    \centering
       \includegraphics[scale=0.8]{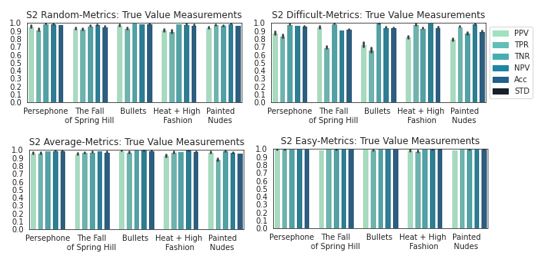}
    \caption{\bf Class-wise metrics results for Dataset S2}
\end{figure}

The capacity of the DCNN model to correctly identify a specific gallery was studied using the four subsets of Dataset S2. The class-wise metrics in Fig 8 displayed distinct distributions for the subsets. The minimal changes of metric Acc show that it is unreliable in finding the correct gallery labels. The closeness of MCC averages (i.e. 0.92, 0.80, 0.94, and 0.99) and of ACC averages (e.g., 0.93, 0.82, 0.94 and 0.99) for subsets {\em random}, {\em difficult}, {\em average}, and {\em easy} indicate a very low possibility of random assignment of gallery labels by the model. The ascending trends support the desired difficulty levels of the four subsets. Classification performance was strong for all galleries. This supports the validity of Hypothesis~I and II.

The detailed analysis of the other metrics produced the following observations. Subset {\em easy} had almost perfect values for all metrics and for all galleries. 
Slightly worse, Subset {\em average} had close to perfect metric values for all galleries with the exception of galleries ''Painted Nudes'' and ``Heat + High Fashion''. 
The relatively low value of TPR for Gallery ''Painted Nudes'' indicates that False Negatives (FNs) are the cause of the poorer performance, e.g., the DCNN model did not recognize well the images of this gallery. 
False Positives (FPs) reduce the classification performance for galleries ``Heat + High Fashion'' and ``Fall of Spring Hill'', i.e. the DCNN model mistook images in Gallery ''Painted Nudes'' as being in one of the two galleries.
For subset {\em random}, the DCNN model  struggled with recognizing images for galleries ``Persephone'' and ``Bullets''. It misclassified their images as being in Gallery ''Painted Nudes''. Galleries ``Heat + High Fashion'' and ``The Fall of Spring Hill'' had the lowest and second lowest PPV and TRP values. For subset {\em difficult}, the images in galleries ``Fall of Spring Hill'' and ``Bullets'' were hard to classify. 
A likely reason for the model confusing images in the two galleries 
as being in galleries ``Heat + High Fashion'' or ''Painted Nudes'' is the high concentration of human figures in these galleries. 
 



{\bf Dataset S3 description}. 
Dataset S3 was designed to have its amounts of inter and within similarities and dissimilarities of EXPs and NEXPs between those of datasets S1 and S2. 
If Hypotheses~I and II were true, then its in between amount of dissimilarities and similarities of NEXPs (Hypotheses~I) and of EXPs (Hypotheses~II) would result a classification performance that is between those for datasets S1 and S2. A strong or a poor performance rejects at least one of the hypotheses.

{\bf Results}. PCA 3D plots in Fig 9(a) depict the similarity of the data points (e.g., art images) of the three subsets. Similar to datasets S1 and S2, more homogeneous point clusters were formed while processing subsets~{\em difficult}, {\em average}, and~{\em easy}. Like for dataset S1, there were no fully homogeneous clusters. 
The PCA plots, with three principal components variance of 81.12\%, 79.28\%, and 73.05\% for the three subsets, validate their purpose with respect to studying the impact of the training/test sets on the DCNN performance.

\begin{figure}[ht]
    \centering
        \includegraphics[scale=0.6]{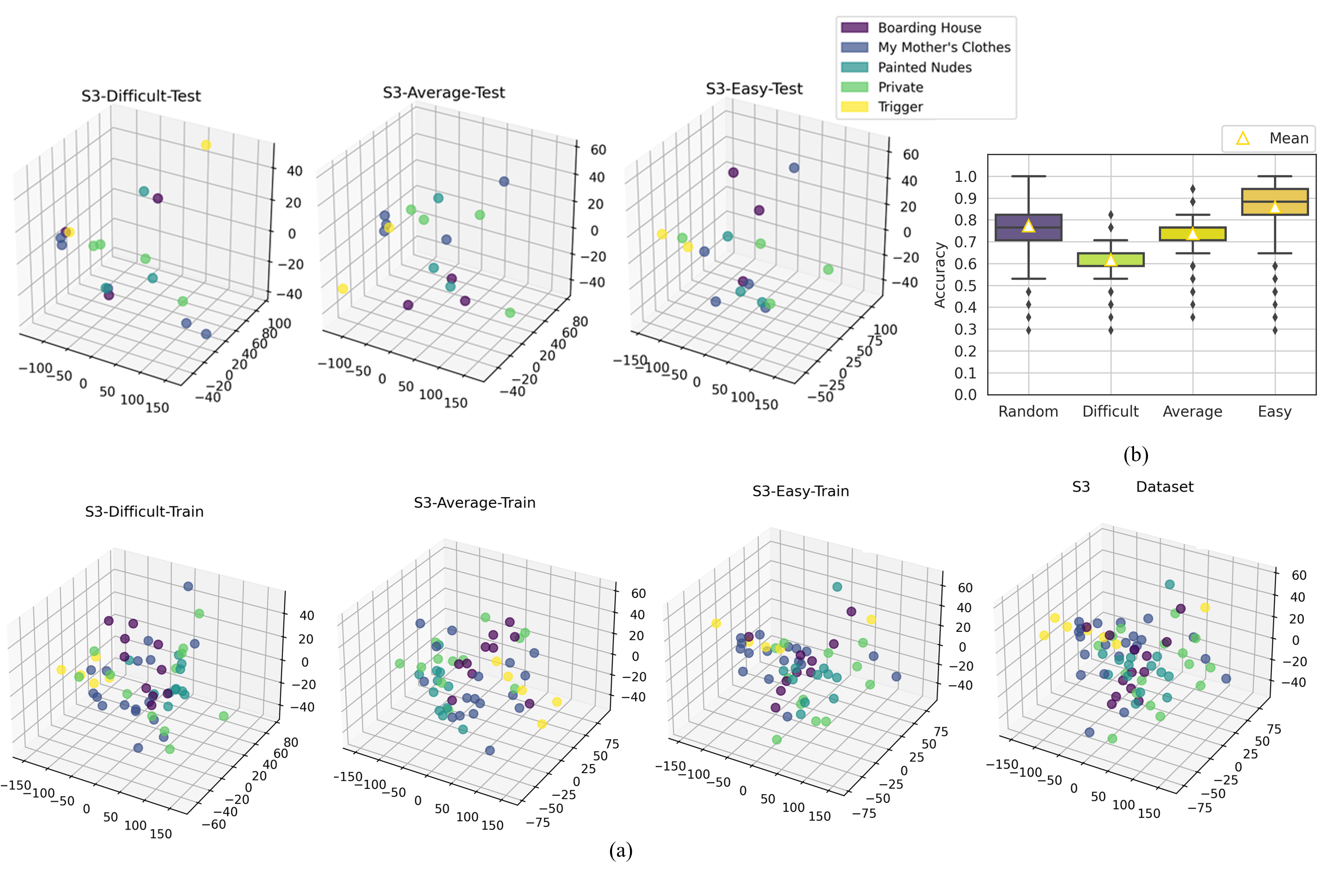}
    \caption{{\bf Results for Dataset S3.} (a) PCA plots for the test sets. (b) Box and whisker plots of the overall accuracies.}
\end{figure}


The resulting performance for classifying Dataset S3 into galleries using the DCNN model was presented next. ANOVA post-hoc tests showed a statistically significant difference between the ACC of all subsets (p $<$ 0.01 for all tests). The box plots of all subsets in Fig 9(b) indicated three distinct distributions. Hence, the DCNN model had a distinct behavior for each of the four subsets.

\begin{figure}[ht]
    \centering
        \includegraphics{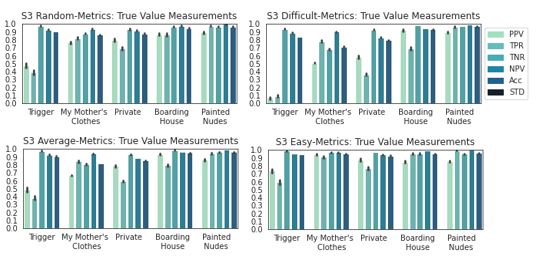}
   \caption{\bf Class-wise metrics results for Dataset S3}
\end{figure}

The capacity of the DCNN model to correctly identify a specific gallery was studied using the four subsets of Dataset S3. The class-wise metrics in Fig 10 displayed distinct distributions for the four subsets. The minimal changes of metric Acc suggest that it is unreliable in finding the correct gallery labels. The closeness of MCC averages (i.e.  0.73, 0.53, 0.68 and 0.84) and of ACC averages (e.g., 0.77, 0.62, 0.74 and 0.869) for subsets {\em random}, {\em difficult}, {\em average}, and {\em easy} indicate a very low possibility of random assignment of gallery labels by the model. The ascending trends support the desired difficulty levels of the four subsets. TNR and NPV values are high for all situations, hence the DCNN model can often correctly indicate if an artwork does not pertain to a gallery. PPV and TPR values are superior than for Dataset S1 but lower than for Dataset S2. Results supprt the two hypotheses in the introductory section.

The detailed analysis of the metrics showed that the DCNN model consistently succeeded in classifying galleries ''Boarding House'' and ''Painted Nudes'' for all subsets. This was due to the high EXP dissimilarity of the two galleries and the rest of the dataset. Gallery ''Boarding House'' was the only grayscale gallery, and ''Painted Nudes'' was the only mixed medium gallery 
with textures of paint and brush on top of photography. 
The low performance for all subsets of Gallery ``Trigger'' was because of its heavy using of NEXPs, as the gallery presents conceptual art.
For the cases with a low performance, like Gallery ``Trigger'' and subset {\em difficult} of Gallery ``Private'', TPR values are often less than PPV values, hence, the artwork in these galleries were incorrectly assigned to other galleries.  Images in galleries ``Private'' and ``Boarding House'' were assigned to Gallery ``My Mother’s clothes''. 



\subsection*{Experiment II. The importance of EXPs Vs. NEXPs in classification of group shows}

{\bf Dataset G1 description}. This experiment investigated the impact of the EXPs diversity within a gallery on DCNN classification performance while NEXPs similarity remained high for a gallery. To that end, the devised dataset included two group exhibitions by multiple artists with distinctive styles, hence diverse EXPs, while their artwork shared NEXPs that allowed the curator to assemble them in a single group exhibition. If hypotheses I and II were true then the classification performance for Dataset~G1 should be worse than the performance in Experiment~I due to the increased EXP diversity within a gallery. Otherwise, at least one of the tw hypotheses should be rejected.  

{\bf Results}. The PCA 3D plots in Fig 11(a) present similar trends as the trends observed for Experiment I. However, the obtained overall clusters of images are less homogeneous as the experiment shifted from subsets {\em difficult} to subset {\em easy}. The data variance was 67.79\%, 64.69\%, and 62.02\% for subsets {\em difficult}, {\em average}, and {\em easy}. This confirmed the validity of the subsets with respect to their purpose for the experiments.

The performance results about DCNN model's capacity to correctly classify Dataset~G1 into galleries were summarized next. ANOVA post-hoc analysis exhibited four distinct behaviors (p $<$ 0.01 or p = 0 in all tests). The box plot in Fig 11(b) of the overall metrics along with their numerical values, e.g., MCC was 0.54, 0.30, 0.56, and 0.64 and ACC was 0.65, 0.46, 0.66, and 0.71, also confirmed the desired levels of difficulties of the three subsets. MCC values were ~0.7 to 0.16 larger than ACC values. 

\begin{figure}[ht]
    \centering
        \includegraphics[scale=0.6]{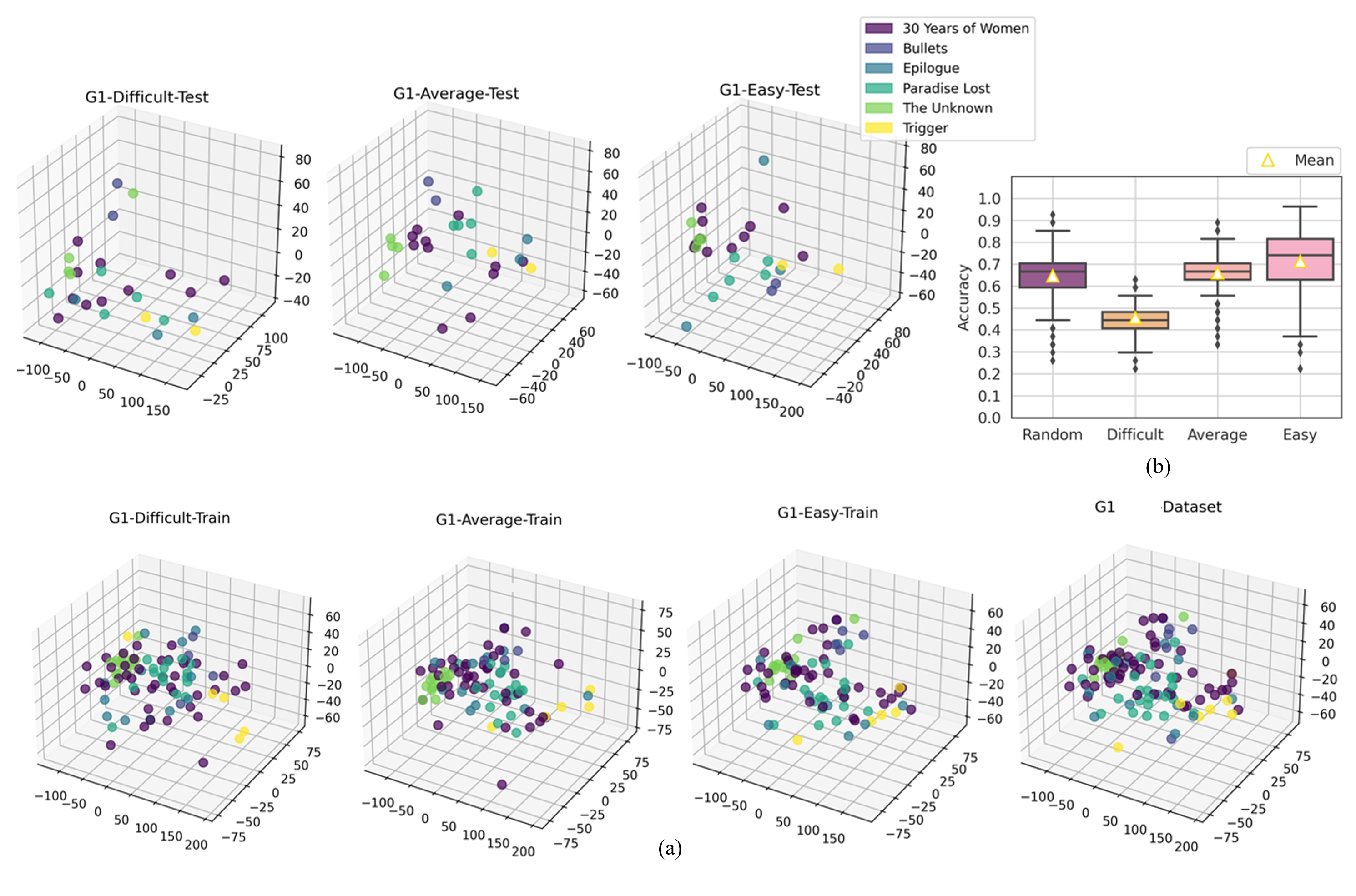}
    \caption{{\bf The results for Dataset G1.} (a) PCA plots for test data for subsets~{\em difficult}, {\em average}, and {\em easy}. (b) Box and whisker plot of overall accuracies (ACC) of the DCNN classification of the four subsets.}
\end{figure}

\begin{figure}[ht]
    \centering
        \includegraphics{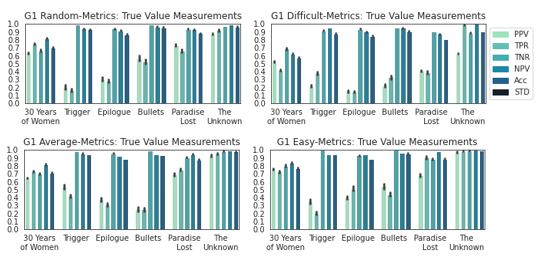}
    \caption{\bf Class-wise metrics results for Dataset G1}
\end{figure}

Class-wise metrics in Fig 12 show  a decreased DCNN performance as compared to the classification performance obtained for Dataset~S1. PPV and TPR values were consistently low for all galleries, except Gallery ``The Unknown'', which offered the best performance for Dataset~G1. Galleries ``Trigger'', ``Epilogue'', and ``Bullets'' were the hardest, second hardest, and third hardest to classify, as Experiment~II went from considering subset~{\em random} to subset~{\em easy}. Hence, the increased diversity of the within-gallery EXPs had an important influence on lowering the DCNN model’s capacity to correctly identify a gallery. Moreover, the high within-gallery EXPs diversity of subset {\em difficult} made this subset to be the hardest to classify among all the subsets used in the three experiments of this work. Note, however, that the increased EXP diversity did not affect the expected difficulty level of subsets~{\em difficult}, {\em average}, and {\em easy}. The lower capacity of DCNN model to correctly classify Dataset~G1 support hypotheses I and II. 

A more detailed analysis showed that for subset
{\em difficult} of Gallery ``The Unknown'', PPV dropped while its TPR stayed high, suggesting that the number of FP increased. Images from other galleries were misclassified to this gallery.
Also, our expectation for Gallery ``30 Years of Women'' was incorrect, as DCNN model had an average performance for this gallery. One possible reason could be its very large number of data points as compared to the other galleries. Our expectation about Gallery ``The Unknown'' were correct despite its sample size being about 2.8 times smaller than for Gallery ``30 Years of Women''. Future work will address the two unexpected situations.  

\subsection*{Experiment III. The Importance of EXPs Vs. NEXPs in Distinguishing Art Images from Non-Art Images}

{\bf Dataset S4 description}. This experiment investigated the validity of the two hypotheses depending on the size of the datasets, including images that were not art. In addition to the galleries in Dataset~S1, Dataset~S4 included a new gallery of non-art images of ready-made, ordinary objects, like human clothes. These images were similar in their EXPs with galleries ``Trigger'' and ``My Mother’s Clothes'', but had no NEXPs. 
If Hypotheses~I and~II were true then Dataset~S4 would be more difficult to classify than~Dataset~S1, including having a lower performance for galleries ``Trigger'' and ``My Mother’s Clothes'' than their classification performance obtained for Dataset~S1. The performance obtained for Gallery ``Non-Art'' should be also low. These outcomes were expected due to Gallery ``Non-Art'' having no NEXPs but presenting similar EXPs as the two galleries above. Otherwise, at least one of the two hypotheses is rejected. 

Experiments were run for two versions of Gallery ``Non-Art'', a 34-image version and an 18-image version. This experiment also addressed the question obtained after the experiment for Set~SF1, which is that the gallery size does not influence classification using NEXPs, but if two galleries are similar in their EXPs, the larger gallery offers better performance.

{\bf Results}. The PCA 3D plots in Fig 13(a) show similar trends as the trends for Experiment~I (Fig~5(a)) and Experiment~II (Fig~7(a)), and confirm the designed levels of difficulty of the four subsets. The data variance was 73.27\%, 68.97\%, and 60.62\% for subsets {\em difficult}, {\em average}, and {\em easy}. 

The performance results for classifying Dataset~S4 into galleries using the DCNN models were as follows. One-way ANOVA results for the two versions of Dataset~S4 (e.g., with 34 and 18 extra non-art images) showed a statistically significant difference (p $<$ 0.01). However, based on the box plot in Fig~14(b), the statistical sensitivity was negligible. The class-wise analysis in Fig~14(a) of the two versions showed identical trends aside from the expected bias because of Gallery ``Non-Art''. Hence, the rest of the experiments focused only on the 34-image version of this gallery.

\begin{figure}[ht]
    \centering
        \includegraphics[scale=0.6]{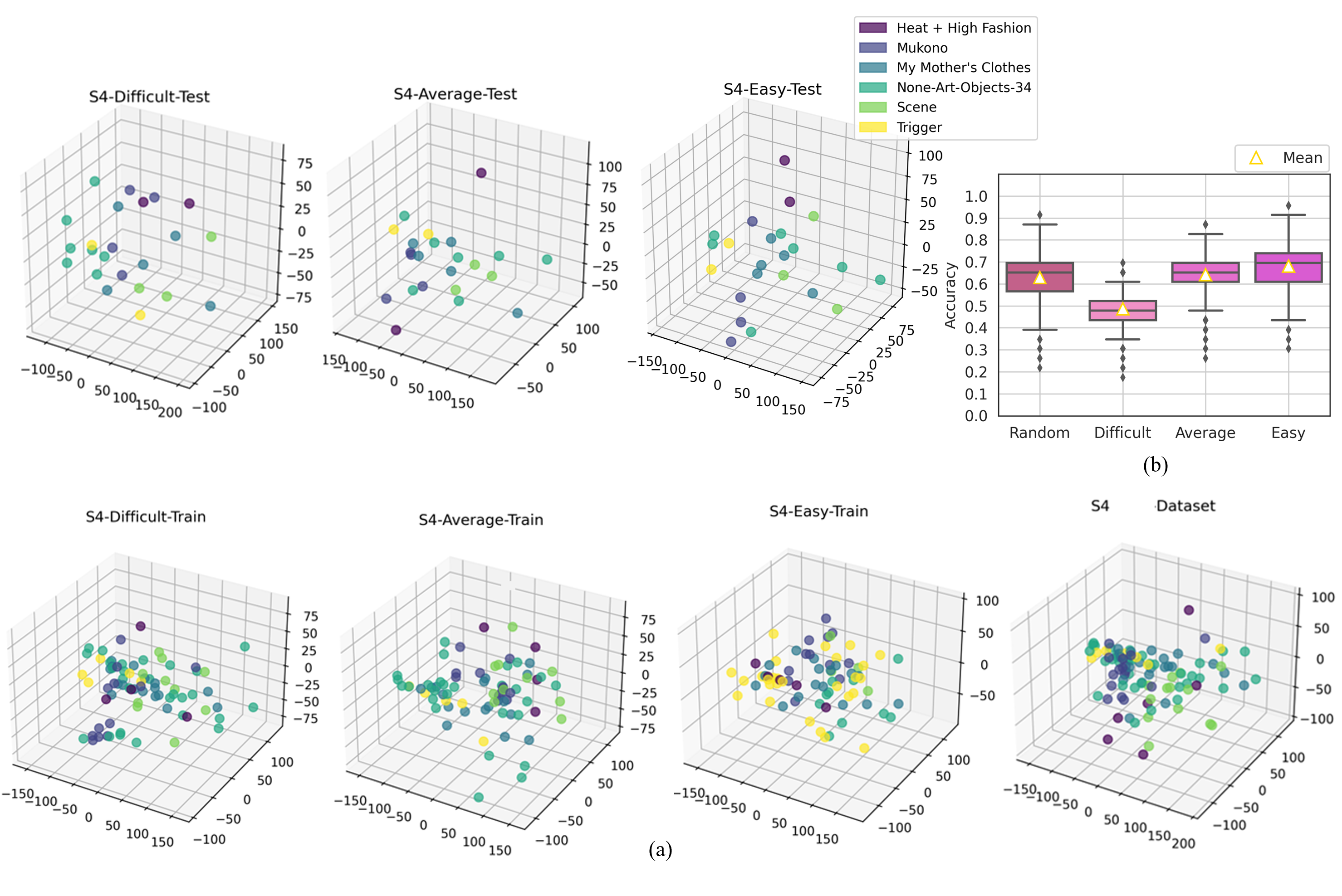}
    \caption{{\bf Results for Dataset S4.} (a) The PCA plots for the training and test sets for subsets {\em difficult}, {\em average}, and {\em easy}. (b) The box and whisker plot of the overall accuracies (ACC) of the DCNN classification of the four subsets.}
\end{figure}

\begin{figure}[ht]
    \centering
        \includegraphics[scale = 0.9]{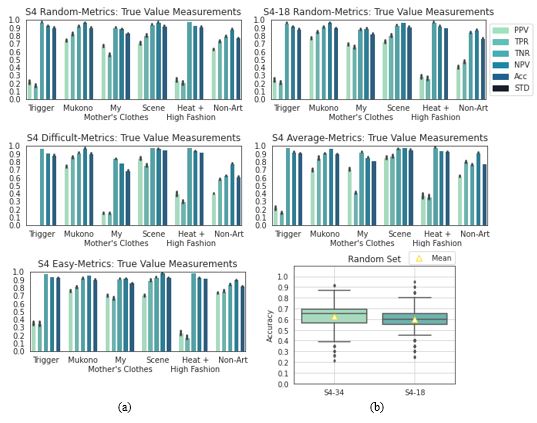}
    \caption{{\bf Results for Dataset S4 class-wise and ``Non-Art'''s 34-image/18-image versions.} (a) Results for the class-wise metrics for Dataset S4 (b) The box and whisker plot of the overall accuracies (ACC) of the DCNN classification for S4-34 and S4-18.}
    \label{f14}
\end{figure}

ANOVA post-hoc analysis showed no statistically significant difference between subsets {\em average} and {\em random} (p$_{Tukey~HSD}$ = 0.001, 99\% C.I = [0.0021, 0.0249]), p$_{Dunnett~T3}$ = 0.002, 99\% C.I = [0.0015, 0.0255]), and p$_{Games-Howell}$ = 0.002, 99\% C.I = [0.0017, 0.0254]), and a strongly significant difference for the other subsets (p $<$ 0.01 or p = 0 in all tests). However, the box plots for subsets~{\em average} and {\em random} present minimal differences, such as outliers and IQR, similar to the previous two experiments. Thus, the DCNN model had distinct behavior for the four subsets. 

The capacity of DCNN model to identify a certain gallery for the four subsets of Dataset~S4 was discussed next. The class-wise metrics in Fig~16(a) indicated different distributions for the four subsets. Similar to set~G1, MCC were 0.55, 0.35, 0.57, and 0.62, and ACC values were 0.63, 0.49, 0.64, and 0.68. The overall and class-wise performance confirmed the expectation that the non-art gallery confused the DCNN model. PPV and TPR values decreased for galleries ``Trigger'' and ``My Mother's Clothes'' as compared to their performance for Dataset~S1. PPV and TPR values for Gallery ``Non-Art'' were low except subsets {\em random} and {\em easy} for which is was higher.  


More specifically, DCNN classification performance was the highest for galleries ``Mukono'' and ``Scene'' (for all their subsets), as they had distinct EXPs while not having similar EXPs with Gallery ``Non-Art''. The large size of ``Mukono'' and ``Scene'' also explains why the model performed better for these galleries as compared to Gallery ``Heat + High Fashion''. The lowest performance among all experiments in this work was obtained for subset {\em difficult} of Gallery ``Trigger''. 
Galleries ``My Mother’s Clothes'', ``Heat+ High Fashion'', and ``Non-Art'' also produced a low classification performance. The performance for subsets {\em easy} was better for all galleries, aside galleries ``Trigger'' and ``Heat+ High Fashion''. 
Art galleries contain detectable EXPs that should differentiate art objects from non-art images of the same object. However, Experiment~III showed that DCNN model’s understanding of EXPs is not sufficient yet.


\section*{Discussion}

Human experts assembled galleries and exhibitions based on their interpretations grounded in mental processing of the visual art images through their explicit and tacit knowledge, obtained through formal training and experience, as well as the ideas specific to their context \cite{43}. Some of their analysis and decisions can be explained through rules, like those summarized in art history \cite{43}, but other are subjective interpretations. It can be argued that there is currently no formalized, quantitatively-defined art ontology and procedural analysis method that could serve as the theoretical backbone for automatically understanding art, including artwork grouping into galleries based on its meaning, artist intention, and viewer interpretation. Instead, art galleries reflect a qualitative, narrative interpretation of art objects based on assembling EXPs into NEXPs that define the meaning, intention, and interpretation of artwork. A conclusion of this work is that using general-purpose vision databases have likely only a limited role in curating art, such as to use them to train DNNs to recognize low-level features, because their meaning is absent (i.e. NEXPs). 
Experiments showed that differentiating between difficulty levels (e.g., subsets {\em difficult}, {\em average}, and {\em easy}) is not cumulative, so that it can be easily quantified statistically, as there are no significant statistical differences between the subsets distinguished by the art expert. 

While other research suggests that DCNNs can reliably learn object fragments and then use these fragments in some scene understanding \cite{22}, this work argues that learning does not include all object features needed to group related artwork into galleries. Features that define an object’s uniqueness within an artwork are likely not learned, if they are not critical in recognizing the object from other objects. For example, a unique but repetitive combination of color on a grayscale image can be specific to an artist and help distinguish his work from other artwork. Due to its repetitive nature, a DCNN might learn the specific feature. However, rare features (e.g., EXPs) are not learned if they pertain to repetitive, high-level concepts (i.e NEXPs). Experiments showed that an artist’s signature was not picked up by DCNN model unless it was based on repetitive EXPs that could be learned, like having a yellow stripe over a greyscale image. A consequence of this observation is that aggregated, statistical metrics can observe global, systematic differences but not individual features. Histograms and outlier analysis, e.g., the number, position, and type of outliers, could address this limitation. 
New metrics are required to capture the assessment by experts, like novelty, craftmanship, and viewer perception of artwork. These metrics must be conditioned by the cultural context of the expert’s assessment.     

\begin{figure}[ht]
    \centering
        \includegraphics[scale=0.9]{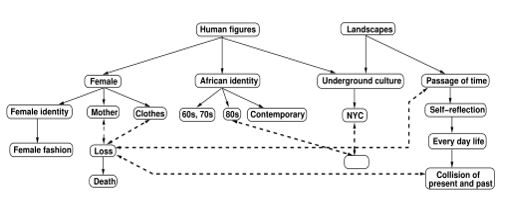}
    \caption{{\bf Gallery themes summary}}
\end{figure}

Fig 15 summarizes the themes of some of the galleries used in the experments. They include genres, like human figures and landscapes. Art objects having, but not necessarily, female figures as one of their central pieces addressed themes, like female identity, female fashion, African identity in the sixties and seventies, eighties, and contemporary. Fig~15 shows an ontology fragment of these concepts, in which arrows indicate the general – to specific relation and dashed lines the combination of concepts that co-occur in an art object. These relations are one kind of possible associated meanings, but other interpretations exist too. Extracting possible meanings for an art object includes identifying the symbolism of the concepts as well as conceptual interpretations, like analogies and metaphors, for the relations among concepts. Moreover, object EXPs, like color, shapes, texture, position, hues, illumination, and so on, can have a certain symbolism, interpretation, or induce a certain feeling to the viewer \cite{43}. For example, common objects in an art composition could point to everyday life, and possibly to the collision of present and past \cite{43}. Or, the relative positioning of objects or their unusual postures, e.g., a chair’s position, can serve a certain purpose in an artwork’s theme narrative \cite{43}.     
Experiments suggest that DCNN classification difficulty relates to the ambiguity of how EXPs, i.e. the visuals of physical objects, relate to NEXPs and their higher-level semantics, like intention and interpretation, which is an artwork’s projection into the idea space. 
The difficulty increases with the abstraction levels of the ontology (Fig 15) where ambiguities occur. Having multiple narratives for an object is also part of the possible ambiguities.    

The analysis of the differences between human art curation and DCNN classification shows several limitations of DCNN models in learning and understanding higher-level semantics. All learned differences are based on visual EXPs, like texture, tones, shapes, and objects. However, models are not capable to dynamically reprioritize the importance of EXPs depending on the process that would lead to understanding NEXPs and the meaning of an art object. For example, Baxandall explains that understanding art is a problem-solving process that constructs a narrative that expresses an object’s meaning \cite{43}. The object must be reinterpreted and reprioritized in the context of the narrative, while possibly dropping significant amounts of general-purpose learning using generic image databases, like ImageNet. 

Another limitation of DCNNs related to NEXP learning refers to creating plausible narratives expressing the theme of an artwork. Narratives are based on the connections to the artist’s or observer’s context (including previous art), and the causal relationships between objects or their symbolic meaning, as well the mapping relations in the case of meanings based on analogies, metaphors, and abstractions \cite{44,45}. Some insight related to the historical context can be inferred using details, like clothing, hair style, or furniture. However, some of these details might not be captured during DCNN learning as they are less frequent than other features. Also, while recent methods can identify and learn some analogical mappings \cite{18}, these methods are symbolic and use numeric metrics to establish the mappings. Current DCNNs cannot learn well mappings, including some with qualitative, subjective, social, and emotional knowledge. A possibility would be to collect such data through surveys and then incorporate it to the DCNN learning process \cite{15}. However, surveys are likely to be ineffective in helping to find which EXPs and NEXPs are the cause for the survey inputs, even though a human expert can indicate quite accurately how visual cues, like color and pattern, produce a certain interpretation or emotion.   


Finally, there is similarity between art creation defined as  open-ended problem solving and other creative processes, like engineering design. Problem framing in design relates to theme selection in art, while creating the structure (architecture) of an engineering solution corresponds to creating the structure of a painting scene. The two solution spaces are constrained by various design rules and aesthetic rules, respectively, e.g., proportions, projections, coloring, and so on \cite{43}. However, there are major differences too. While engineering is mostly guided by numerical performance values that express the objective quality of a design and to a much lesser degree by subjective factors, like preference for some functions, art creation is guided by arguably no quantitative analysis, being subjected only to qualitative, subjective evaluations. Besides, an engineering solution has a well-defined meaning and purpose, which is perceived in the same way by all. In contrast, the meaning of art depends on the artists and viewers, gets shaped by different cultures, and evolves over time.

\section*{Conclusions}

Modern theories of art suggest that Exhibited Properties (EXPs) and Non-Exhibited Properties (NEXPs) characterize any work of art. EXPs are visible features, like color, texture and form, and NEXPs are artistic aspects that result by relating an art object to human history, culture, the artist's intention, and the viewer's perception. Current work on using Deep Neural Network (DNN) models to computationally characterize artwork suggests that DNNs can learn EXPs and can gain some insight on meaning aspects tightly related to EXPs, but there are no extensive studies about the degree to which NEXPs are learned during DNN training, and then used for automated activities, like classifying artwork into galleries. To address this limitation, this work conducted a comprehensive set of experiments about the degree to which Deep Convolutional Neural Network (DCNN) models learn NEXPs of artwork. Two hypotheses were formulated to answer this question: The first hypothesis states that DCNN models do not capture NEXPs well for art gallery classification. The second hypotheses states that EXP similarities and differences within and between art galleries determine the difficulty level of DCNN classification.	

Three experiments were devised and performed to verify the two hypotheses using datasets about art galleries assembled by an art expert. Experiments used the VGG-11 DCNN pretrained on ImageNet database, and then retrained using art images. The three experiments considered the following situations: (1) using EXPs and NEXPs for classification of art objects in solo (single artist) galleries, (2) utilizing EXPs and NEXPs for classification of art objects in group galleries, and (3) distinguishing art objects from non-art objects, and the impact of dataset size on classification results. Datasets were put together for each situation and for different difficulty levels of DCNN classification. Results were analyzed using statistical and classification measures.

The experimental study validated the two hypotheses. VGG-11 DCNN did not learn NEXPs sufficiently well to support accurate classification of modern artwork into galleries similar to those curated by human experts, and EXPs were insufficient for understanding, interpreting, and classifying artwork. Higher EXP similarity among galleries or higher EXP diversity within a gallery increased the difficulty level of classification in spite of their NEXP values, which suggests that EXPs were the determining factor in classification. 
Dataset size was not a main factor in improving DCNN classification, but increasing dataset size can help galleries with similar EXPs. 
This work suggests that any attempt to automate art understanding should be equipped with mechanisms to capture well EXP and NEXP of artwork.

The three experimental studies are useful not only to characterize the general limitations of DCNN models, but also to understand if NEXPs of art objects can be distinguished only using their EXPs, thus if an art object is fully specified within its body of similar work, e.g., gallery, or if NEXPs depend to a significant degree on elements not embodied into an art object, like contextual elements, the artist’s intention, and the viewer’s interpretation. Experimental results support the second perspective. 

\section*{Further Research Directions}

The DCNN model studied in this work can arguably be a rough, qualitative predictor of artwork understanding by a person without artistic training. The model’s art ``knowledge'' comes by superimposing features learned from a few art galleries on the features learned using images from the general vision domain. Experiments with DCNN models that would aggressively transfer knowledge from the art domain (and not only a few galleries) would add to the understanding of how well DCNNs can learn NEXPs. Another avenue of future work would consider other DNN models, such as VisionTransformer \cite{45} and ConvNeXt \cite{46}, alongside with Transfer Learning techniques with a higher learning capacity, e.g., cascaded network architectures. 
Finally, the design and analysis of the datasets and experiments
could 
explore the DCNN's preferences and biases, i.e. whether shape or color are more important in classification, or which features are tend to be misclassified by measuring the frequency of the misclassification instances at the image level. 




%
%
%
\bibliography{Bib.bib}

\section*{Supporting information}
\paragraph*{S1 Appendix: Metrics Definition}
\paragraph*{S2 Appendix: Dataset Descriptions}
\paragraph*{S3 Appendix: Follow-Up Experiments: The Effect of Dataset Size on Classification}
\paragraph*{S4 Appendix: Histogram Plots}
\paragraph*{S5 Appendix: Normal Q-Q Plots}
\paragraph*{S6 Appendix: Summary Box Plots}
\paragraph*{S7 Appendix: Class-wise Metrics (FPR, FNR, and FDR)}
\paragraph*{S8 Appendix: Statistical Details}
\paragraph*{S9 Appendix: Dataset Metadata}

\end{document}


\vspace*{0.2in}
\section*{Supporting Information}

\subsection*{S1 Appendix: Metrics Definition}
\subsubsection*{Abbreviations}

\begin{tabular}{ c c c c}
True Positive &	 TP	& False Positive & 	FP\\
True Negative &	 TN	& False Negative & FN  
\end{tabular}

\subsubsection*{Definitions}

\begin{tabular}{l c c c}
    \( Positive Predictive Value (PPV,Precision) = \) & \(\frac{TP}{TP+FP}=\) & \(\frac{\# True Positives}{\# Total Positive Predictions}\) \\
    \(True Predictive Rate (TPR,Recall,Sensitivity) =\) & \(\frac{TP}{TP+FN}=\) & \(\frac{\# True Positives}{\# Total Positives}\) \\
    \(True Negative Rate (TNR,Specificity) = \) & \(\frac{TN}{TN+FP}=\) & \(\frac{\# True Negatives}{\# Total Negatives}\) \\
    \(Negative Predictive Value (NPV)= \) & \(\frac{TN}{TN+FN}=\) & \(\frac{\# True Negatives}{\# Total Negaive Predictions}\) \\
    \(False Positive Rate (FPR) =\) & \(\frac{FP}{FP+TN} =\) & \( \frac{\# False Negatives}{\# Total Negaitves}\) \\
    \(False Negaitve Rate (FNR) = \) & \(\frac{FN}{TP+FN} =\) & \(\frac{\# False Negatives}{\# Total Positives}\) \\
    \(False Discovery Rate (FDR) =\) & \(\frac{FP}{TP+FN} = \) & \(\frac{\# False Positives}{\# Total Positives}\) \\
    \(Class-wise/Overall  Accuracy (Acc) =\) & \(\frac{T}{T+F} = \) & \(\frac{\# True Positives \& Negatives}{\# Total Positives \&  Negatives}\) \\
\end{tabular}

\subsection*{S2 Appendix: Dataset Descriptions}
\subsubsection*{Art History/Visual Arts Terms}

\begin{table}[H]
\setlength{\tabcolsep}{5pt}
\renewcommand{\arraystretch}{1.7}
\captionsetup{margin=-2.25in}
\caption{\bf List of common terms in art history and visual arts to describe the EXPs}
\begin{adjustwidth}{-2.25in}{0in}
\begin{tabular}{p{3cm}|p{3.3cm}p{3.1cm}|p{2.5cm} p{2.5cm} p{2.7cm}}
\rowcolor{lightgray!50}
\multicolumn{1}{c|}{\bf\normalsize Medium} & \multicolumn{2}{c|}{\bf\normalsize Color} & \multicolumn{3}{c}{\bf\normalsize Shape, Form, Texture} \\ 
\small
- Black and white\newline photography \newline - Color photography \newline - Mix media, using\newline photography & - High Saturation \newline - Medium saturation \newline - Low saturation \newline - Cool colors \newline - Warm colors  \newline - Neutral colors \newline - Dark Value  \newline - Light value & \newline - Mid-tones \newline - High Contrast \newline - Low contrast \newline - Medium contrast \newline - Monochromatic \newline - Chromatic \newline - Colorful & - Figurative \newline - Organic \newline - Geometric \newline - Textured \newline - Plain \newline - Dynamic \newline - Open form \newline - Painterly & \newline - Linear \newline - Closed form \newline - Formless \newline - Chaotic \newline - Brush strokes \newline - Architectural \newline - Industrial \newline - Abstract & \newline - Decorative \newline - Pattern \newline - Floral \newline - Text/calligraphy 
\end{tabular}
\end{adjustwidth}
\end{table}

\begin{table}[H]
\setlength{\tabcolsep}{5pt}
\renewcommand{\arraystretch}{1.7}
\begin{adjustwidth}{-2.25in}{0in}
\begin{tabular}{p{3cm}p{3cm}p{2.9cm}|p{2.4cm}p{2.9cm}p{2.9cm}}
\rowcolor{lightgray!50}
\multicolumn{3}{c|}{\bf\normalsize Composition$^1$} & \multicolumn{3}{c}{\bf\normalsize Subject Matter$^2$}\\ 
\small
- Open composition \newline - Closed composition \newline - Symmetrical \newline - Tendency toward\newline symmetry \newline - Asymmetrical \newline - Centered alignment\newline of the subject matter \newline  & - Horizontal frame \newline - Vertical frame \newline - Square frame \newline - Multiple frames \newline - Emphasis on a single subject matter & \newline - Busy and crowded\newline compositions \newline - Empty and quiet\newline composition & - Human body \newline - Female body \newline - Male body \newline - Human torso \newline - Male torso \newline - Female torso \newline - Portraits \newline - Hidden human faces & \newline - Still life \newline - Animals \newline - Nature/landscape  \newline - Architecture \newline - Interior space \newline - Industrial space \newline - Urban spaces & \newline - Domestic space \newline - Shadows, reflections \newline - Everyday objects \newline - Graffiti \newline - Dolls/toys \newline - Cars 
\end{tabular}
\end{adjustwidth}
\end{table}

\begin{adjustwidth}{-2.25in}{0in}
\noindent\footnotesize{\(^1\)}Arrangements of visual elements within a frame \newline
\footnotesize{\(^2\)}What we are looking at\newline
\end{adjustwidth}

\subsubsection*{EXPs, NEXP, and Outliers}

\setlength{\LTleft}{-2.25in}
\setlength{\LTright}{0in}
\begin{singlespace}
\setlength{\tabcolsep}{7pt}
\renewcommand{\arraystretch}{1.7}
\small
\begin{longtable}{p{0.11\linewidth}|llll}
\captionsetup{margin=-2.25in}
\caption{\bf Galleries' EXPs}
\\ 
\rowcolor{lightgray}\multirow{\bf\normalsize Gallery} & \multicolumn{4}{c}{\bf\normalsize EXPs}  \\ 
\rowcolor{lightgray!50} 
\cellcolor{lightgray} & \multicolumn{1}{l|}{\bf\normalsize Medium, Color} & \multicolumn{1}{p{3cm}|}{\bf\normalsize Shape, Form, \newline Texture } & \multicolumn{1}{l|}{\bf\normalsize Composition} & \multicolumn{1}{l}{\bf\normalsize Subject Matter} \\ 
{\emph {30 Years of Women}} & \multicolumn{1}{p{4.4cm}|}{- Black and white photography \newline - Color photography \newline - Mix media, using photography \newline - High saturation  \newline - Medium saturation  \newline - Low saturation \newline - Cool colors \newline - Warm colors \newline -  Neutral colors\newline - Dark value \newline - Light value \newline - Mid-tones \newline - High contrast \newline - Low contrast \newline - Medium contrast \newline - Monochromatic  \newline - Chromatic \newline - Colorful} & \multicolumn{1}{p{3cm}|}{- Figurative \newline - Organic \newline - Geometric textured \newline - Plain \newline - Dynamic\newline - Open form\newline - Painterly linear \newline - Closed form \newline - Formless\newline chaotic brush strokes \newline - Architectural \newline - Industrial \newline - Abstract decorative pattern \newline - Floral \newline - Text/calligraphy}  & \multicolumn{1}{p{4.5cm}|}{- Pen composition \newline - Closed composition \newline - Symmetrical \newline - Tendency toward symmetry  \newline - Asymmetrical \newline - Centered alignment of the subject matter \newline - Horizontal frame \newline - Vertical frame \newline - Square frame \newline - Emphasis on a single subject matter \newline - Busy and crowded compositions \newline - Empty and quiet composition} &  \multicolumn{1}{p{3.1cm}}{- Human body \newline - Female body \newline - Male body \newline - Human torso  \newline - Female torso \newline - Male torso \newline - Portraits \newline - Hidden human faces \newline - Still life \newline - Animals \newline - Nature/landscape \newline - Architecture \newline - Interior space \newline - Industrial space \newline - Domestic space \newline - Shadows, reflections \newline - Everyday objects \newline - Graffiti}  \\ \hline
 
{\emph {Boarding House}} & \multicolumn{1}{p{4.4cm}|}{-Black and white photography \newline - Low saturation \newline - Neutral colors \newline - Dark value \newline - Light value \newline - Mid-tones  \newline - Medium contrast \newline - Monochromatic} & \multicolumn{1}{p{3cm}|}{- Organic \newline - Figurative \newline - Textured \newline - Open form \newline - Closed form \newline - Linear \newline - Repetitions of a shape or an element \newline - Pattern}  & \multicolumn{1}{p{4.5cm}|}{- Square frame \newline - Busy/crowded compositions \newline - Open composition \newline - Closed composition \newline - Tendency toward symmetry \newline - Asymmetrical \newline - Centered alignment of the subject matter \newline - Empty and quiet composition} &  \multicolumn{1}{p{3.1cm}}{- Human body  \newline - Still life \newline - Animals \newline - Nature/landscape  \newline - Abandoned urban space  \newline - Graffiti  \newline - Everyday objects  \newline - Dolls/toys}  \\ \hline

{\emph {Bonsai}} & \multicolumn{1}{p{4.4cm}|}{- Black and white photography \newline - 
 Monochromatic \newline - Warm tone \newline - Neutral colors \newline - Dark value \newline - Light value \newline - Mid-tones  \newline - High contrast \newline - Low contrast \newline - Medium contrast} & \multicolumn{1}{p{3cm}|}{- Organic shapes \newline - Natural \newline - Textured \newline - Plain \newline - Open forms \newline - Painterly \newline - Formless \newline - Abstract}  & \multicolumn{1}{p{4.5cm}|}{- Close composition \newline - Asymmetrical composition \newline - Tendency toward symmetrical composition \newline - Vertical composition \newline - Horizontal composition \newline - Simple and quiet \newline - Centered alignment of the subject matter \newline - Emphasis on a single subject matter} &  \multicolumn{1}{p{3.1cm}}{- Landscape/nature \newline - Animals/ owls \newline - Shadows/reflections}  \\ \hline

{\emph {Bullets}} & \multicolumn{1}{p{4.4cm}|}{- Colored photography \newline - Limited colors \newline - Medium saturation \newline - Low saturation \newline - Warm colors \newline - Neutral colors \newline - Dark value \newline - Light value \newline - Mid-tone \newline - Medium contrast \newline - Chromatic} & \multicolumn{1}{p{3cm}|}{- Figurative  \newline - Organic  \newline - Textured  \newline - Closed forms  \newline - Decorative pattern  \newline -Text/calligraphy}  & \multicolumn{1}{p{4.5cm}|}{- Horizontal \newline compositions \newline - Vertical compositions \newline - Multi-frames \newline - Closed composition \newline - Asymmetrical \newline - Tendency toward symmetry \newline - Centered alignment \newline - Busy and crowded composition} &  \multicolumn{1}{p{3.1cm}}{- Female body \newline - Arabic calligraphy\newline/text \newline - Interior space}  \\ \hline

{\emph {Close}} & \multicolumn{1}{p{4.4cm}|}{- Color photography \newline - High saturation \newline - Medium saturation \newline - Cool colors \newline - Warm colors \newline - Dark value \newline - Light value \newline - Mid-tones  \newline - High contrast \newline - Medium contrast \newline - Chromatic \newline - Colorful} & \multicolumn{1}{p{3cm}|}{- Figurative \newline - Organic \newline - Geometric \newline - Open form \newline - Closed form \newline - Architectural \newline - Decorative \newline - Pattern \newline - Floral}  & \multicolumn{1}{p{4.5cm}|}{- Busy and crowded \newline - Asymmetrical \newline - Square frames \newline - Open composition  \newline - Closed composition \newline - Tendency toward symmetry \newline - Centered alignment of the subject matter \newline - Emphasis on a single subject} &  \multicolumn{1}{p{3.1cm}}{- Human body \newline - Female body \newline - Male body \newline - Human torso \newline - Male torso \newline - Female torso \newline - Portraits \newline - Hidden human faces \newline - Nature/landscape \newline - Architecture \newline - Interior space \newline - Domestic space \newline - Shadows, reflections}  \\ \hline

{\emph {Converging \newline Territories}} & \multicolumn{1}{p{4.4cm}|}{- Color photography (coupler printed) \newline - Medium saturation \newline - Low saturation \newline - Cool colors \newline - Warm colors \newline - Neutral colors \newline - Dark value \newline - Light value \newline - Mid-tones  \newline - Low contrast \newline - Medium contrast \newline - Chromatic} & \multicolumn{1}{p{3cm}|}{- Figurative \newline - Textured surfaces \newline - Repetition of Arabic calligraphic elements \newline - Open forms \newline - Closed forms \newline - Linear \newline - Decorative pattern}  & \multicolumn{1}{p{4.5cm}|}{- Vertical frames \newline - Horizontal frames \newline - Multiple frames \newline - Symmetrical\newline - Asymmetrical \newline - Centered alignment of the subject matter \newline - Emphasis on a single subject matter} &  \multicolumn{1}{p{3.1cm}}{- Female body \newline - Portraits \newline - Hidden human faces \newline - Still life \newline - Arabic calligraphy \newline - Female veils}  \\ \hline

{\emph {Eat \newline Flowers}} & \multicolumn{1}{p{4.4cm}|}{- Color photography \newline - High saturation \newline - Medium saturation \newline - Low saturation \newline - Cool colors \newline - Warm colors \newline - Neutral colors \newline - Dark value \newline - Light value \newline - Mid-tones  \newline - High contrast \newline - Low contrast \newline - Medium contrast \newline - Monochromatic \newline - Chromatic \newline - Colorful} & \multicolumn{1}{p{3cm}|}{- Organic and natural \newline - Floral \newline - Figurative \newline - Text/ letters \newline - Plain \newline - Textured \newline - Abstract \newline - Decorative \newline - Formless \newline - Open form \newline - Painterly \newline - Linear}  & \multicolumn{1}{p{4.5cm}|}{- Open composition \newline - Closed composition \newline - Tendency toward symmetry \newline - Asymmetrical \newline - Centered alignment of the subject matter \newline - Horizontal frame \newline - Vertical frame \newline - Emphasis on a single subject matter \newline - Busy and crowded compositions \newline - Empty and quiet composition} &  \multicolumn{1}{p{3.1cm}}{- Human body  \newline - Female body  \newline - Male body  \newline - Human torso  \newline - Portraits  \newline - Interior spaces  \newline - Hidden human faces \newline - Still life  \newline - Animals \newline - Nature/landscape  \newline - Shadows, reflections \newline - Everyday objects \newline - Dolls/toys \newline - Cars \newline - Text/calligraphy  \newline - Fruits}\\ \hline

{\emph {Epilogue}} & \multicolumn{1}{p{4.4cm}|}{- Color photography \newline - Black and White photography \newline - High saturation \newline - Medium saturation \newline - Low saturation \newline - Cool colors \newline - Warm colors  \newline - Neutral colors  \newline - Dark Value \newline - Light value  \newline - Mid-tones \newline - High contrast \newline - Low contrast \newline - Medium contrast \newline - Monochromatic \newline - Chromatic \newline - Colorful} & \multicolumn{1}{p{3cm}|}{- Figurative \newline - Organic \newline - Geometric \newline - Textured \newline - Plain \newline - Dynamic \newline - Open form \newline - Linear \newline - Closed form \newline - Formless \newline - Chaotic \newline - Architectural \newline - Industrial  \newline - Abstract \newline - Pattern \newline - Floral}  & \multicolumn{1}{p{4.5cm}|}{- Vertical  \newline - Horizontal \newline - Squares \newline - Asymmetrical \newline - Near symmetrical \newline - Crowded and busy \newline - Empty and simple \newline - Open composition  \newline - Closed composition \newline - Centered alignment of the subject matter \newline - Multiple frames \newline - Emphasis on a single subject matter} & \multicolumn{1}{p{3.1cm}}{- Human body \newline - Portraits \newline - Still life \newline - Nature/landscape \newline - Architecture \newline - Interior space \newline - Industrial space \newline - Urban spaces \newline - Domestic space \newline - Shadows, reflections \newline - Everyday/objects \newline - Dolls/toys}  \\ \hline

{\emph {Evidence}} & \multicolumn{1}{p{4.4cm}|}{- Color photography \newline - High Saturation \newline - Medium saturation \newline - Cool colors \newline - Warm colors \newline - Dark Value \newline - Light value  \newline - Mid-tones \newline - High Contrast \newline - Medium contrast \newline - Chromatic  \newline - Colorful \newline - Dramatic \newline - Monochromatic} & \multicolumn{1}{p{3cm}|}{- Figurative \newline - Organic \newline - Geometric \newline - Open form \newline - Closed form \newline - Architectural \newline - Pattern \newline - Floral \newline - Dramatic lights}  & \multicolumn{1}{p{4.5cm}|}{- Busy and crowded \newline - Empty and quiet \newline - Asymmetrical \newline - Square frames \newline - Open composition \newline - Closed composition \newline - Tendency toward \newline symmetry \newline - Centered alignment of the subject matter \newline - Emphasis on a single subject } &  \multicolumn{1}{p{3.1cm}}{- Human body \newline - Female body \newline - Male body \newline - Human torso \newline - Male torso \newline - Female torso \newline - Portraits \newline - Hidden human faces \newline - Nature/landscape \newline - Architecture \newline - Interior space \newline - Domestic space \newline - Shadows, reflections \newline - Dramatic light}  \\ \hline

{\emph {Familiar Landscapes}} & \multicolumn{1}{p{4.4cm}|}{- Color photography  \newline - Medium saturation \newline - High saturation \newline - Cool colors \newline - Warm colors \newline - Dark Value \newline - Light value  \newline - Mid-tones \newline - High Contrast \newline - Medium contrast \newline - Chromatic \newline - Colorful} & \multicolumn{1}{p{3cm}|}{- Organic \newline - Textured \newline - Open form \newline - Painterly \newline - Formless \newline - Floral}  & \multicolumn{1}{p{4.5cm}|}{- Horizontal \newline - Vertical \newline - Asymmetrical \newline - Fairly busy and \newline crowded} &  \multicolumn{1}{p{3.1cm}}{- Nature/landscape \newline - Light  \newline - Spring}  \\ \hline

{\emph {Heat +\newline  High \newline Fashion}} & \multicolumn{1}{p{4.4cm}|}{- Black and white photography\newline - Monochromatic \newline - Dark Value\newline - Light value \newline - Midtones  \newline - High Contrast} & \multicolumn{1}{p{3cm}|}{- Figurative \newline - Organic \newline - Plain \newline - Open forms \newline - Painterly}  & \multicolumn{1}{p{4.5cm}|}{- Closed compositions \newline - Open composition \newline - Tendency toward symmetry \newline - A-symmetrical \newline - Centered \newline - Alignment of the  subject matter\newline - Horizontal frames \newline - Vertical frames \newline - Emphasis on a single subject matter \newline - Empty and quiet composition \newline - Busy and crowded \newline compositions}  &  \multicolumn{1}{p{3.1cm}}{- Female body \newline - Female torso \newline - Portraits \newline - Hidden human faces \newline - Fashion \newline - Interior space \newline - Shadows, reflections} \\ \hline

{\emph {Hivernacle}} & \multicolumn{1}{p{4.4cm}|}{- Mixed media (photography\newline and paint) \newline -  Black, white, grays + gold \newline - Medium saturation \newline - Low saturation \newline - Warm colors \newline - Neutral colors  \newline - Dark Value \newline - Light value  \newline - Mid-tones \newline - High Contrast \newline - Medium contrast \newline - Monochromatic} & \multicolumn{1}{p{3cm}|}{- Figurative \newline - Organic \newline - Plain \newline - Open forms \newline - Closed forms \newline - Brush strokes}  & \multicolumn{1}{p{4.5cm}|}{- Closed composition \newline - Tendency toward \newline symmetry \newline - Asymmetrical \newline - Centered alignment of the subject matter \newline - Horizontal frame \newline - Vertical frame \newline - Emphasis on a single subject matter \newline - Empty and quiet composition} &  \multicolumn{1}{p{3.1cm}}{- Female body \newline - Female torso \newline - Hidden human faces}  \\ \hline

{\emph {Kawa = \newline  Flow}} & \multicolumn{1}{p{4.4cm}|}{- Black and white photography \newline - Color Photography \newline - Monochromatic \newline - Photography \newline - Cool colors \newline - Warm colors \newline - Dark Value \newline - Light value  \newline - Mid-tones \newline - Low saturated colors \newline - Neutral colors \newline - High Contrast \newline - Low contrast \newline - Medium contrast} & \multicolumn{1}{p{3cm}|}{- Figurative  shapes \newline - Organic shapes \newline - Textured  \newline - Plain \newline - Open forms \newline - Painterly \newline - Formless \newline - Abstract \newline - Floral}  & \multicolumn{1}{p{4.5cm}|}{- Open composition \newline - Close composition \newline - Asymmetrical composition \newline - Symmetrical composition \newline - Vertical composition \newline - Horizontal composition \newline - Square composition \newline - Busy and crowded composition \newline - Simple and quiet \newline - Centered alignment of the subject matter \newline - Emphasis on a single subject matter} &  \multicolumn{1}{p{3.1cm}}{- Landscape/nature  \newline - Human body \newline - Female body \newline - hidden human faces  \newline - Still life \newline - Animals  \newline - Shadows/reflections \newline - Human and nature relationship}  \\ \hline

{\emph {Little Deaths}} & \multicolumn{1}{p{4.4cm}|}{- Color photography \newline - High saturated \newline - Colors \newline - Low saturated colors \newline - Medium to high contrast \newline - Medium saturation \newline - Cool colors \newline - Warm colors \newline - Neutral colors \newline - Chromatic  \newline - Colorful} & \multicolumn{1}{p{3cm}|}{- Natural \newline - Organic \newline - Textured \newline - Open form \newline - Painterly \newline - Formless \newline - Chaotic \newline - Floral}  & \multicolumn{1}{p{4.5cm}|}{- Asymmetrical horizontal \newline - Vertical  \newline - Closed compositions \newline - Centered alignment of the subject matter \newline - Emphasis on a single subject matter \newline - Busy and crowded compositions \newline - Domestic space} &  \multicolumn{1}{p{3.1cm}}{- Landscape/
nature \newline - Animal  \newline - Human body  \newline - Female body \newline - Male body  \newline - Human torso}  \\ \hline

{\emph {Mukono}} & \multicolumn{1}{p{4.4cm}|}{- Black and white photography\newline - Monochromatic\newline - Dark Value\newline - Light value \newline - Midtones \newline - High Contrast \newline - Medium contrast \newline - Low contrast \newline - Low saturation \newline - Neutral colors}   & \multicolumn{1}{p{3cm}|}{- Figurative \newline - Organic \newline - Plain \newline - Open forms \newline - Closed forms} & \multicolumn{1}{p{4.5cm}|}{- Closed composition \newline - Tendency toward symmetry \newline - Asymmetrical \newline - Centered alignment of the subject matter \newline - Horizontal frame \newline - Vertical frame \newline - Emphasis on a single subject matter \newline - Empty and quiet composition} & \multicolumn{1}{p{3.1cm}}{- Human torso/male torso \newline - Female torso \newline - Portraits \newline - Hidden human faces \newline - Nature/landscape \newline - Human body \newline - Everyday objects \newline - Still life \newline - Animals} \\ \hline

{\emph {My Mother’s Clothes}} & \multicolumn{1}{p{4.4cm}|}{- Color photography \newline - High Saturation \newline - Medium saturation\newline - Low saturation\newline - Cool colors\newline - Warm colors \newline - Neutral colors\newline - Dark Value\newline - Light value \newline - Midtones  \newline - High Contrast \newline - Low contrast\newline - Medium contrast\newline - Chromatic} & \multicolumn{1}{p{3cm}|}{- Organic \newline - Geometric \newline - Textured \newline - Plain \newline - Open form \newline - Linear \newline - Closed form \newline - Decorative \newline - Pattern \newline - Floral \newline - Text } & \multicolumn{1}{p{4.5cm}|}{- Closed composition \newline - Tendency toward symmetry \newline - Asymmetrical \newline - Centered alignment of the subject matter \newline - Square frames \newline - Emphasis on a single subject matter \newline - Busy and crowded compositions \newline - Empty and quiet composition} & \multicolumn{1}{p{3.1cm}}{- Female clothes \newline - Still life \newline - Everyday objects \newline - Domestic space} \\ \hline

{\emph {Native}} & \multicolumn{1}{p{4.4cm}|}{- Color photography  \newline - High Saturation \newline - Medium saturation \newline - Cool colors \newline - Warm colors \newline - Dark Value \newline - Light value  \newline - Mid-tones  \newline - High Contrast \newline - Medium contrast \newline - Chromatic \newline - Colorful} & \multicolumn{1}{p{3cm}|}{- Figurative \newline - Organic \newline - Open form \newline - Closed form \newline - Architectural \newline - Decorative  \newline - Pattern \newline - Floral \newline - Textured \newline - Formless  \newline - Abstract \newline - Pattern}  & \multicolumn{1}{p{4.5cm}|}{- Busy and crowded \newline - Empty and quiet \newline - Asymmetrical \newline - Square frames \newline - Open composition \newline - Closed composition \newline - Tendency toward symmetry \newline - Centered alignment of the subject matter \newline - Emphasis on a single subject} &  \multicolumn{1}{p{3.1cm}}{- Nature/landscape \newline - Human body  \newline - Female body \newline - Male body \newline - Human torso  \newline - Male torso \newline - Female torso \newline - Portraits \newline - Hidden human faces \newline - Interior space \newline - Domestic space \newline - Shadows, reflections \newline - Human in nature}  \\ \hline

{\emph {New York, \newline Paris, and Rome}} & \multicolumn{1}{p{4.4cm}|}{- Black and white photography \newline - Medium to high contrast \newline - Monochromatic \newline - Neutral colors \newline - Dark value \newline - Light value \newline - Mid-tones} & \multicolumn{1}{p{3cm}|}{- Figurative \newline - Urban \newline - Architectural \newline - Organic \newline - Natural and organic \newline - Open forms \newline - Close forms \newline - Dynamic \newline - Formless \newline - Industrial text} & \multicolumn{1}{p{4.5cm}|}{- Open composition \newline - Closed composition \newline - Tendency toward \newline symmetry \newline - Asymmetrical \newline - Centered alignment of the subject matter \newline - Horizontal frame \newline - Vertical frame \newline - Square frame \newline - Busy and crowded compositions} & \multicolumn{1}{p{3.1cm}}{- Human body \newline - Female body \newline - Male body \newline - Human torso \newline - Portraits \newline - Hidden human faces \newline - Animals \newline - Nature/landscape \newline - Architecture \newline - Industrial space \newline - Urban spaces \newline - Shadows, reflections \newline - Cars} \\ \hline

{\emph {Painted Nudes}} & \multicolumn{1}{p{4.4cm}|}{- Mixed Media (combination of photography and painting) \newline - High Saturation \newline - Medium saturation \newline - Low saturation \newline - Cool colors \newline - Warm colors \newline - Neutral colors \newline - Dark Value \newline - Light value  \newline - Mid-tones \newline - High contrast \newline - Medium contrast \newline - Chromatic \newline - Color full} & \multicolumn{1}{p{3cm}|}{- Figurative \newline - Textured \newline - Dynamic brush strokes \newline - Open form  \newline - Painterly \newline - Brush strokes \newline - Abstract \newline - Formless \newline - Chaotic}  & \multicolumn{1}{p{4.5cm}|}{- Open composition \newline - Closed composition \newline - Asymmetrical \newline - Horizontal frame \newline - Vertical frame \newline - Busy and crowded compositions \newline - Centered alignment of the subject matter} &  \multicolumn{1}{p{3.1cm}}{- Female body \newline - Female torso}  \\ \hline

{\emph {Paradise Lost}} & \multicolumn{1}{p{4.4cm}|}{- Color photography \newline - High Saturation  \newline - Medium saturation \newline - Cool colors \newline - Warm colors \newline - Dark Value \newline - Light value  \newline - Mid-tones  \newline - High Contrast \newline - Medium contrast \newline - Chromatic \newline - Colorful} & \multicolumn{1}{p{3cm}|}{- Figurative \newline - Organic \newline - Geometric \newline - Open form \newline - Closed form \newline - Architectural \newline - Decorative \newline - Pattern \newline - Floral}  & \multicolumn{1}{p{4.5cm}|}{- Busy and crowded \newline - Asymmetrical \newline - Square frames \newline - Open composition \newline - Closed composition \newline - Tendency toward symmetry \newline - Centered alignment of the subject matter \newline - Emphasis on a single subject} &  \multicolumn{1}{p{3.1cm}}{- Human body \newline - Female body \newline - Male body \newline - Human torso \newline - Male torso \newline - Female torso \newline - Portraits \newline - Hidden human faces \newline - Nature/landscape \newline - Architecture \newline - Interior space \newline - Domestic space \newline - Shadows, reflections} \\ \hline

{\emph {Persephone}} & \multicolumn{1}{p{4.4cm}|}{- Mixed media (photography + paint) \newline - High Saturation \newline - Medium saturation \newline - *Ed colors \newline - Cool colors \newline - Warm colors \newline - Neutral color \newline - Dark Value \newline - Light value \newline - Mid-tones \newline - Medium contrast \newline - Chromatic \newline - Colorful} & \multicolumn{1}{p{3cm}|}{- Organic \newline - Open form \newline - Textured \newline - Painterly \newline - Formless \newline - Chaotic \newline - Brush strokes \newline - Abstract \newline - Floral \newline - Abstract}  & \multicolumn{1}{p{4.5cm}|}{- Open composition \newline - Closed composition \newline - Asymmetrical \newline - Horizontal frame \newline - Vertical frame \newline - Busy and crowded compositions} &  \multicolumn{1}{p{3.1cm}}{- Landscape/nature}  \\ \hline

{\emph {Private}} & \multicolumn{1}{p{4.4cm}|}{- Color photography \newline - Brownish dominant color \newline - Medium saturation \newline - Warm colors \newline - Neutral colors \newline - Dark Value \newline - Light value \newline - Mid-tones \newline - Medium contrast \newline - Monochromatic \newline - Chromatic} & \multicolumn{1}{p{3cm}|}{- Figurative \newline - Organic \newline - Textured \newline - Plain \newline - Open form \newline - Closed form \newline - Formless \newline - Abstract \newline - Pattern \newline - Text}  & \multicolumn{1}{p{4.5cm}|}{- Square frames \newline - Open composition \newline - Closed composition \newline - Symmetrical  \newline - Tendency toward symmetry \newline - Asymmetrical \newline - Centered alignment of the subject matter \newline - Emphasis on a single subject matter \newline - Busy and crowded \newline - Compositions \newline - Empty and quiet composition} &  \multicolumn{1}{p{3.1cm}}{- Human body \newline - Female body \newline - Male body \newline - Human torso \newline - Male torso \newline - Female torso \newline - Hidden human faces \newline - Nature/landscape \newline - Interior space \newline - Domestic space \newline - Shadows, reflections}  \\ \hline

{\emph {Scene}} & \multicolumn{1}{p{4.4cm}|}{- Black and white photography\newline - Monochromatic \newline - Dark Value \newline - Light value \newline - Midtones \newline - High Contrast \newline - Medium contrast} & \multicolumn{1}{p{3cm}|}{- Figurative \newline - Textured \newline - Plain \newline - Open form \newline - Closed-form\newline - Pattern} & \multicolumn{1}{p{4.5cm}|}{- Closed composition \newline - Tendency toward symmetry \newline - Asymmetrical \newline - Centered alignment of the subject matter \newline - Square frame \newline - Emphasis on a single subject matter} & \multicolumn{1}{p{3.1cm}}{- Human body  \newline - Female body \newline - Male body \newline - Human torso \newline - Male torso \newline - Female torso \newline - Interior space \newline - Shadows, reflections \newline - Artists} \\ \hline

{\emph {Sweet 16}} & \multicolumn{1}{p{4.4cm}|}{- Color photography \newline - High Saturation \newline - Medium saturation \newline - Cool colors \newline - Warm colors \newline - Neutral colors \newline - Chromatic \newline - Colorful \newline - Dark Value \newline - Light value  \newline - Mid-tones \newline - High Contrast} & \multicolumn{1}{p{3cm}|}{- Figurative \newline - Texture/pattern\newline on the background \newline - Floral \newline - Decorative \newline - Organic}  & \multicolumn{1}{p{4.5cm}|}{- Centered figures \newline - Vertical \newline - Tendency toward symmetry \newline - Closed compositions \newline - Emphasis on a single subject matter} & \multicolumn{1}{p{3.1cm}}{- Female torso \newline - Portraits \newline - Interior space}  \\ \hline

{\emph {The Fall \newline of Spring \newline Hill}} & \multicolumn{1}{p{4.4cm}|}{- Color photography \newline - High Saturation \newline - Medium saturation \newline - Cool colors \newline - Warm colors \newline - Neutral colors \newline - Dark Value \newline - Light value  \newline - Mid-tones \newline - High Contrast \newline - Medium contrast \newline - Chromatic} & \multicolumn{1}{p{3cm}|}{- Figurative \newline - Organic \newline - Figurative \newline - Organic \newline - Textured \newline - Plain \newline - Dynamic \newline - Closed form \newline - Architectural}  & \multicolumn{1}{p{4.5cm}|}{- Open composition \newline - Closed composition \newline - Asymmetrical \newline - Horizontal frame \newline - Emphasis on a single subject matter  \newline - Busy and crowded compositions} &  \multicolumn{1}{p{3.1cm}}{- Human body \newline - Female body \newline - Children \newline - Human torso \newline - Nature/landscape \newline - Interior spaces \newline - Domestic spaces \newline - Shadow/reflections  \newline - Everyday objects \newline - Still-life}  \\ \hline

{\emph {The Fallen Fawn}} & \multicolumn{1}{p{4.4cm}|}{- Color photography \newline - Medium saturated colors \newline - High saturated colors \newline - Medium contrast} & \multicolumn{1}{p{3cm}|}{- Figurative \newline - Everyday objects \newline - Interior spaces \newline - Organic \newline - Natural \newline - Open form \newline - Painterly \newline - Closed form \newline - Chaotic \newline - Decorative \newline - Pattern \newline - Floral}  & \multicolumn{1}{p{4.5cm}|}{- Horizontal \newline - Asymmetrical \newline - Busy and crowded \newline - Open composition \newline - Emphasis on a single subject matter} &  \multicolumn{1}{p{3.1cm}}{- Human body \newline - Female body \newline - Male body \newline - Human torso \newline - Male torso \newline - Female torso \newline - Portraits \newline - Still life \newline - Animals \newline - Nature/landscape \newline - Interior space \newline - Domestic space \newline - Shadows, reflections \newline - Everyday objects \newline - Dolls/toys \newline - Cars}  \\ \hline

{\emph {The \newline Garden}} & \multicolumn{1}{p{4.4cm}|}{- Color photography \newline - Diverse colors \newline - Cool colors \newline - Warm colors \newline - Neutral colors \newline - Dark Value \newline - Light value \newline - Mid-tones \newline - High Contrast \newline - Medium contrast \newline - Chromatic \newline - Colorful} & \multicolumn{1}{p{3cm}|}{- Figurative organic \newline - Textured \newline - Plain \newline - Dynamic \newline - Open form \newline - Painterly \newline - Closed form \newline - Formless \newline - Chaotic \newline - Abstract \newline - Decorative \newline - Pattern \newline - Floral}  & \multicolumn{1}{p{4.5cm}|}{- Closed composition \newline - Symmetrical \newline - Tendency toward symmetry \newline - Asymmetrical \newline - Centered alignment of the subject matter \newline - Horizontal frame \newline - Vertical frame \newline - Emphasis on a single subject matter \newline - Busy and crowded compositions \newline - Empty and quiet composition} &  \multicolumn{1}{p{3.1cm}}{- Human body \newline - Female body \newline - Female torso \newline - Portraits \newline - Hidden human faces \newline - Nature/landscape \newline - Shadows, reflections}  \\ \hline

{\emph {The \newline Unknown}} & \multicolumn{1}{p{4.4cm}|}{- Mixed media photography \newline - Diverse colors \newline - Cool colors \newline - Warm colors \newline - Neutral colors  \newline - Dark Value \newline - Light value  \newline - Mid-tones  \newline - High Contrast \newline - Medium contrast \newline - Chromatic \newline - Colorful} & \multicolumn{1}{p{3cm}|}{- Figurative \newline - Organic \newline - Textured \newline - Plain \newline - Open form \newline - Painterly \newline - Closed form \newline - Brush stroke}  & \multicolumn{1}{p{4.5cm}|}{- Closed composition \newline - Symmetrical \newline - Tendency toward symmetry \newline - Asymmetrical \newline - Centered alignment of the subject matter \newline - Horizontal frame \newline - Vertical frame \newline - Emphasis on a single subject matter \newline - Busy and crowded \newline compositions \newline - Empty and quiet composition} &  \multicolumn{1}{p{3.1cm}}{- Human body \newline - Female body \newline - Female torso \newline - Portraits \newline - Hidden human \newline faces}  \\ \hline

{\emph {Trigger}} & \multicolumn{1}{p{4.4cm}|}{- Color photography\newline - Medium saturation \newline - Low saturation \newline - Neutral colors \newline - Cool colors \newline - Dark value \newline - Light value \newline - Mid-tones  \newline - High Contrast \newline - Low contrast \newline - Medium contrast \newline - Chromatic} & \multicolumn{1}{p{3cm}|}{- Plain \newline - Organic\newline - Geometric \newline - Closed form}  & \multicolumn{1}{p{4.5cm}|}{- Closed composition \newline - Tendency toward symmetry \newline - Asymmetrical \newline - Centered alignment of  the subject matter \newline - Horizontal frame \newline - Vertical frame \newline - Emphasis on a single subject matter \newline - Empty and quiet composition} &  \multicolumn{1}{p{3.1cm}}{- Interior space \newline - Domestic space \newline - Still life  \newline - Animals}  \\ \hline
\end{longtable}

\small
\setlength{\tabcolsep}{5pt}
\renewcommand{\arraystretch}{1.7}
\begin{longtable}{p{0.11\linewidth}|llll}
\captionsetup{margin=-2.25in}
\caption{\bf Galleries' NEXPs}
\\ 
\rowcolor{lightgray}\multirow{\bf\normalsize Gallery} & \multicolumn{3}{c}{\bf\normalsize NEXPs}  \\ 
\rowcolor{lightgray!50} 
\cellcolor{lightgray} & \multicolumn{1}{l|}{\bf\normalsize Context^3} & \multicolumn{1}{l|}{\bf\normalsize Intention^4} & \multicolumn{1}{l}{\bf\normalsize Meaning} \\ 

{\emph {30 Years of Women}} & \multicolumn{1}{p{4cm}|}{- Women art 20-21 centuries photographers \newline - Global/cross-cultural \newline - Contemporary \newline - Feminism} & \multicolumn{1}{p{6cm}|}{- Modern and contemporary women art \newline - Female gaze, female identity} & \multicolumn{1}{p{6.5cm}}{- Female body \newline - Female gaze \newline - Identity \newline - female perspective and gaze \newline - Female body (not all) \newline - Global feminism \newline - Sensuality \newline - Female fashion \newline - Ethnicity and cultural identity \newline - Realism and indexicality \newline - Intersection of feminism, gender, and sexuality} \\ \hline

{\emph {Boarding House}} & \multicolumn{1}{p{4cm}|}{- Contemporary, American and South African \newline - Jewish culture \newline - Contemporary surrealism} & \multicolumn{1}{p{6cm}|}{- Surrealism art, or art that relates to human psyche \newline - Indexicality} & \multicolumn{1}{p{6.5cm}}{- A space between reality and fantasy \cite{56} \newline - Human existence in relation to place and environment \cite{56}\newline - Outmoded objects, abandoned places, and things, elements that eco-surrealism in art and film}  \\ \hline

{\emph {Bonsai}} & \multicolumn{1}{p{4cm}|}{- Japanese modern and contemporary art \newline - Zen philosophy \cite{60}\newline - Romanticism \newline - Traditions of landscape art} & \multicolumn{1}{p{6cm}|}{- Contemporary Japan \newline - Japanese philosophy of Zen \newline - Painterly quality in photography, meaning a photography that is poetic and subjective vs. Realistic \newline - Being one with nature, monism of human and nature, mind, and body} & \multicolumn{1}{p{6.5cm}}{- Zen philosophy \newline - The relationship between human and nature, human as a part of the nature or in a monistic relation with nature \newline - Female body as a metaphor of a landscape \newline - Being one with nature, the monism of human and nature, mind and body \newline - A dream-like quality \newline - *Strong emotions and subjective expression}  \\ \hline

{\emph {Bullets}} & \multicolumn{1}{p{4cm}|}{- Orientalism \newline - Globalization \newline - Contemporary photography \newline - Middle eastern art \newline - Imperialism/ colonialism} & \multicolumn{1}{p{6.cm}|}{- Middle eastern identity in relation to western orientalism and imperialism \cite{52} \newline - Identity politics in visual art and culture} & \multicolumn{1}{p{6.5cm}}{- Orientalist representation of the artist’s cultural identity \newline - Identity politics}  \\ \hline

{\emph {Close}} & \multicolumn{1}{p{4cm}|}{- German-Brazilian contemporary photographer \newline - Traditional nude imagery rooted in Western classical and academic art} & \multicolumn{1}{p{6cm}|}{- Aesthetics and beauty of the human body, inspired by the Western classical art historical discourse of representation of a nude figure \newline - Intimate connection, intimate space with human figures} & \multicolumn{1}{p{6.5cm}}{- Intimate depiction of the models \cite{55}\newline - A natural state of the human body, and reflecting being comfortable in one own’s skin \cite{55}\newline - Sensual quality \newline - Formal beauty, often stereotypical, traditional and classical idea of beauty \newline - Human body integrated or in a monistic relationship with nature}  \\ \hline

{\emph {Converging \newline Territories}} & \multicolumn{1}{p{4cm}|}{- Arab cultures \newline - Islamic culture \newline - Arab-American  \newline - Modern and contemporary\newline - Orientalism \newline - Imperialism  \newline - Globalization} & \multicolumn{1}{p{6cm}|}{- Contemporary Arab American art \cite{52} \newline - Art that challenges Western Orientalism \cite{52}\newline - Immigrant’s identity in American \cite{52}} & \multicolumn{1}{p{6.5cm}}{- Challenging the stereotypes about Arab and Muslim women \cite{52} \newline - Challenging social, and cultural hierarchies \cite{52} \newline - Arab women Identity \cite{52} \newline - Islamic culture \cite{52}}  \\ \hline

{\emph {Eat \newline Flowers}} & \multicolumn{1}{p{4cm}|}{- Contemporary British Art \newline - Contemporary American\newline art \newline - Conceptual art} & \multicolumn{1}{p{6cm}|}{- Contemporary European and American art\newline - Representation of nature\newline and the human relationship with nature \newline - “I want my photographs to be a jolt, they explore the magic in the world while having one foot very much placed in reality.”\cite{61}\newline - Conceptual art} & \multicolumn{1}{p{6.5cm}}{- “Natural environment and familial relationships”  \cite{61}\newline - Influence by the natural environment of her hometown \cite{61}\newline - Poetic \newline - Dreamlike \newline - Memory \newline - Time}  \\ \hline

{\emph {Epilogue}} & \multicolumn{1}{p{4cm}|}{- Modern and contemporary photography \newline - Cross-cultural \newline - Realism \newline - Surrealism \newline - Pop art and culture} & \multicolumn{1}{p{6cm}|}{- Contemporary \newline - Realism \newline - Surrealism art \newline - Pop art and culture} & \multicolumn{1}{p{6.5cm}}{- Documenting aftermath of an event \newline - Realism and indexicality \newline - Epilogue (the title of the show) meaning “the moment after.” According to the gallery, the moment after refers to documenting a moment as an event. Recording the event is in fact a record of the aftermath of the event \cite{57}. \newline - Outmoded objects \newline - A world between fantasy and reality \newline - Memory and time}  \\ \hline

{\emph {Evidence}} & \multicolumn{1}{p{4cm}|}{- German-Brazilian contemporary photographer \newline - Traditional nude imagery rooted in Western classical and academic art} & \multicolumn{1}{p{6cm}|}{- Aesthetics and beauty of the human body, inspired by the Western classical art historical discourse of representation of nude figure \newline - Intimate connection \newline - Intimate space with human figures} & \multicolumn{1}{p{6.5cm}}{- Intimacy and intimate depiction of the models \cite{55}\newline - A natural state of the human body, and reflecting being comfortable in one own’s skin \cite{55}\newline - Sensual quality \newline - Formal Beauty, often stereotypical, traditional and classical idea of beauty \newline - Human body integrated or in a monistic relationship with nature}  \\ \hline

{\emph {Familiar Landscapes}} & \multicolumn{1}{p{4cm}|}{- Contemporary American\newline photography \newline - Landscape art \newline - Realism/romanticism, contemporary} & \multicolumn{1}{p{6cm}|}{- Landscape art \newline - Her hometown and its nature} & \multicolumn{1}{p{6.5cm}}{- Her hometown and the chronology of its inhabitants \cite{50} \newline - The nature and place of community \cite{50}\newline - Passing of time \cite{50}\newline - A sense of romanticism through dramatic light in nature}  \\ \hline

{\emph {Heat +\newline  High \newline Fashion}} & \multicolumn{1}{p{4cm}|}{- Modern contemporary \newline - New York, “bohemian.”} & \multicolumn{1}{p{6cm}|}{- Engaging female fashion through photography \newline - Realism yet a sense of ambiguity through blurred images and painterly qualities of the medium of photography (in that sense it can place itself against modernist photography and its notion of medium specificity)} & \multicolumn{1}{p{6.5cm}}{- Photographs of  dancer Isadora Duncan \newline - Fashion photography that reflects the time and historical context \newline - Female identity through fashion and clothing \newline - Realism yet a sense of ambiguity through blurred images and painterly qualities of the medium of photography}  \\ \hline

{\emph {Hivernacle}} & \multicolumn{1}{p{4cm}|}{- Contemporary Spanish art \newline - Fine arts photography inspired by fashion photography \newline - Female identity in the contemporary art \newline - Surrealism} & \multicolumn{1}{p{6cm}|}{- Contemporary \newline - Issue of identity, and identity \newline - Fine art photography inspired by fashion \newline - Dream like portraits \newline - Ambiguous portraits \newline - Artistic tradition of portraiture} & \multicolumn{1}{p{6.5cm}}{- Ambiguous identities (From the title) \newline - Human psyche \newline - Influenced by surrealism and psychoanalysis \newline - Relation between dream and reality \newline - Female identity \newline - Ambiguity  \newline - Portraiture}  \\ \hline

{\emph {Kawa = \newline  Flow}} & \multicolumn{1}{p{4cm}|}{- Japanese modern and contemporary art \newline - Zen philosophy \newline - Romanticism \newline - Traditions of landscape art} & \multicolumn{1}{p{6cm}|}{- Contemporary Japan \cite{60}\newline - Japanese philosophy of Zen\cite{60}\newline - Painterly quality in photography, meaning photography that is poetic and subjective vs. realistic \newline - Being one with nature, the monism of human and nature, mind and body} & \multicolumn{1}{p{6.5cm}}{- Zen philosophy \newline - The relationship between human and nature, human as a part of the nature or in a monistic relation with nature \newline - Female body as a metaphor of a landscape \newline - Being one with nature, the monism of human and nature, mind and body \newline - A dream-like quality \newline - Strong emotions and subjective expression}  \\ \hline

{\emph {Little Deaths}} & \multicolumn{1}{p{4cm}|}{- Contemporary American\newline photography \newline - Landscape art \newline - Realism/romanticism, contemporary} & \multicolumn{1}{p{6cm}|}{- Landscape art \newline - Her hometown and its nature} & \multicolumn{1}{p{6.5cm}}{- Her hometown and the chronology of its inhabitants \cite{50}\newline - The nature and place of community \cite{50}\newline - Passing of time \cite{50}\newline - A sense of romanticism through dramatic light in nature}  \\ \hline

{\emph {Mukono}}  & \multicolumn{1}{p{4cm}|}{- Contemporary photography \newline - Cultural studies}  & \multicolumn{1}{p{6cm}|}{- Realism and documentary photography} & \multicolumn{1}{p{6.5cm}}{- Documenting people around the world \newline - Race, ethnicity, culture-racial and cultural identity}  \\ \hline

{\emph {My Mother’s Clothes}}  & \multicolumn{1}{p{4cm}|}{- Contemporary photography and conceptual art (using ready-made objects as works of art)} & \multicolumn{1}{p{6cm}|}{- Conceptual photography inspired by conceptual art and the use of readymade/ordinary objects and blurring the boundary between art and life \newline -Photographing her mothers’ clothes and personal idem as a form of portraits or chronology of her mother’s life \newline - Using art and photography to cope with the loss of her mother and her mother’s suffering from Alzheimer} & \multicolumn{1}{p{6.5cm}}{- Her mother’s clothes and personal items as her mother’s portrait/body= clothes as a metonymy of the person \newline - Remembering the past, memories of her mother, perhaps a sense of nostalgia \newline - Coping with Truma of loss of her mother and her suffering from Alzheimer \newline - Gender expression/identity \newline - Social class in America} \\  \hline

{\emph {Native}} & \multicolumn{1}{p{4cm}|}{- German-Brazilian contemporary photographer \newline - Photography\newline influenced by the tradition of academic and representational painting} & \multicolumn{1}{p{6cm}|}{- “a personal journey. Metaphorically, I was thinking of a bird that flies back into the forest, searching for its childhood." \cite{59}\newline - Aesthetics and beauty of the human body, inspired by the Western classical art historical discourse of representation of nude figure \cite{59}\newline - Intimate connection, intimate space with human figures\cite{59}} & \multicolumn{1}{p{6.5cm}}{- “a personal journey. Metaphorically, I was thinking of a bird that flies back into the forest, search for its childhood.” \newline - “Abstract wishes and dreams” \newline - “Her native country” to re-interpret her past \newline - Intimacy and se intimate depiction of the models \newline - A natural state of the human body, and reflecting being comfortable in one own’s skin  \newline - Sensual quality \newline - Formal beauty, often stereotypical, traditional and classical idea of beauty \newline - Human body integrated or in a monistic relationship with nature}  \\ \hline

{\emph {New York, \newline Paris, and Rome}} & \multicolumn{1}{p{4cm}|}{- Modern and contemporary art \newline - American Art \newline - Realism \newline - Documentary photography} & \multicolumn{1}{p{6cm}|}{- Realism \newline - Documentary and indexicality in photography \newline - Europe and North American art in the mid-late 20th century \newline - Capturing a moment in time through photography} & \multicolumn{1}{p{6.5cm}}{- Documentary photography \newline - Urban everyday life \newline - Realism \newline - Indexicality \newline - Human relationships with their environments \newline - Artistic and cultural scenes in New York, Paris, and Rome}  \\ \hline

{\emph {Painted Nudes}} & \multicolumn{1}{p{4cm}|}{- Contemporary art \newline - American art} & \multicolumn{1}{p{6cm}|}{- Contemporary American mixed media art \newline - Experimental approach to the medium of photography \newline - Formal beauty} & \multicolumn{1}{p{6.5cm}}{- Searching for beauty in simple and everyday things \cite{53} \newline - Eroticism} \\ \hline

{\emph {Paradise Lost}} & \multicolumn{1}{p{4cm}|}{- German-Brazilian contemporary photographer} & \multicolumn{1}{p{6cm}|}{- Aesthetics and beauty of the human body, inspired by the Western classical art historical discourse of representation of nude figure \newline - Intimate connection, intimate space with human figures} & \multicolumn{1}{p{6.5cm}}{- Intimate depiction of the models \cite{55}\newline - A natural state of the human body, and reflecting being comfortable in one own’s skin \cite{55}\newline - Sensual quality \newline - Formal beauty, often stereotypical, traditional and classical idea of beauty \newline - Human body integrated or in a monistic relationship with nature}  \\ \hline

{\emph {Persephone}} & \multicolumn{1}{p{4cm}|}{- Contemporary American\newline nature} & \multicolumn{1}{p{6cm}|}{- Contemporary \newline - Contemporary American towns and communities \newline - Contemporary landscape mixed media art \newline - Representation of her hometown, a sense of community and the story of people who have lived there} & \multicolumn{1}{p{6.5cm}}{- “her hometown and the chronology of its inhabitant” \cite{50} \newline - An imaginative depiction of landscape vs a simple realism \newline - The brush strokes can show the passage of time or artist’s subjective feeling to her environment}  \\ \hline

{\emph {Private}} & \multicolumn{1}{p{4cm}|}{- German-Brazilian contemporary photographer} & \multicolumn{1}{p{6cm}|}{- Aesthetics and beauty of the human body, inspired by the Western classical art historical discourse of representation of nude figure \newline - Intimate connection, intimate space with human figures} & \multicolumn{1}{p{6.5cm}}{- Intimate depiction of the models \cite{55}\newline - A natural state of the human body, and reflecting being comfortable in one own’s skin \cite{55}\newline - Sensual quality \newline - Formal beauty, often stereotypical, traditional and classical idea of beauty \newline - Human body integrated or in a monistic relationship with nature}  \\ \hline

{\emph {Scene}} & \multicolumn{1}{p{4cm}|}{- 1960s underground/avant-garde artists’ scenes in the United States} & \multicolumn{1}{p{6cm}|}{- Realistic photographs of Avant-garde artists in New York during the 1960s \newline - Documentary photography} & \multicolumn{1}{p{6.5cm}}{- Photography and realism \newline - Indexicality \newline - Representation of avant-garde artists in NYC during the c1960s \newline - Human emotion and psychological expression}  \\ \hline

{\emph {Sweet 16}} & \multicolumn{1}{p{4cm}|}{- Contemporary American photography \newline - Tradition of portrait art \newline - Realism} & \multicolumn{1}{p{6cm}|}{- Contemporary, American \newline - Realism/representational art \newline - Tradition of portraiture painting, photography \newline - Realism} & \multicolumn{1}{p{6.5cm}}{- Engaging the historical tradition of portraiture \newline - The identity of the model the way they want to be perceived and the way they are perceived by the camera \cite{58}\newline - Her hometown and the chronology of its inhabitants. \cite{58}\newline - The nature and place of community\cite{58} \newline - Passing of time}  \\ \hline

{\emph {The Fall \newline of Spring \newline Hill}} & \multicolumn{1}{p{4cm}|}{- Contemporary American photography \newline - Life and culture in the suburbs/small town in America \newline - Representational art} & \multicolumn{1}{p{6cm}|}{- Contemporary photography \newline - Representation of life in suburbs/small towns in America \newline - “Cinematic Scenarios an incident from a summer church camp in which a child injures himself by falling from a dilapidated wooden play structure and the mothers’ fierce reaction to deconstruct it in retribution.”\cite{54}} & \multicolumn{1}{p{6.5cm}}{- Creating cinematic scenarios \newline - Representation of memory through photography \newline - Life and culture in the suburbs/small town in America \newline - Domestic life \newline - Performativity in photography}  \\ \hline

{\emph {The Fallen Fawn}} & \multicolumn{1}{p{4cm}|}{- Contemporary American\newline photography \newline - Life and culture in the suburbs/small town in America \newline - Representational art \newline - Contemporary film and cinema \newline - Narrative art} & \multicolumn{1}{p{6cm}|}{- Contemporary photography \cite{54}\newline - Representation of life in suburbs/small towns in America \cite{54}\newline - “Cinematic Scenarios an incident from a summer church camp in which a child injures himself by falling from a dilapidated wooden play structure and the mothers’ fierce reaction to deconstruct it in retribution.”\cite{54}} & \multicolumn{1}{p{6.5cm}}{- Creating cinematic scenarios \cite{54}\newline - Representation of memory through photography \cite{54}\newline - Life and culture in the suburbs/small town in America \cite{54}\newline - Domestic life \cite{54}\newline - Performativity in photography\cite{54}}  \\ \hline

{\emph {The \newline Garden}} & \multicolumn{1}{p{4cm}|}{- Contemporary Spanish art and photography \newline - Subjective images through photography vs. Realism} & \multicolumn{1}{p{6cm}|}{- Subjective expression and painterly qualities through photography rather using photography as a medium of realism} & \multicolumn{1}{p{6.5cm}}{- Nature and female body \newline - Mysterious and ambiguous identities and environments \newline - Psychological expression through photography vs. Realism \newline - Metaphor of female body as landscape \newline - A sense of timelessness}  \\ \hline

{\emph {The \newline Unknown}} & \multicolumn{1}{p{4cm}|}{- Contemporary Spanish art \newline - Fashion photography} & \multicolumn{1}{p{6cm}|}{ - Photography and art inspired by fashion photography Contemporary \newline - Issue of identity, ambiguity} & \multicolumn{1}{p{6.5cm}}{- Ambiguous identities (From the title) \newline - Human psyche \newline - Influenced by surrealism and psychoanalysis \newline - Relation between dream and reality \newline - Female identity \newline - Ambiguity \newline - Portraiture}  \\ \hline

{\emph {Trigger}} & \multicolumn{1}{p{4cm}|}{- Contemporary photography \newline - Conceptual photography} & \multicolumn{1}{p{6cm}|}{- Conceptual\newline photography \newline - Photographing everyday \newline objects and domestic space (her hometown) \newline - Capturing time passing\newline through photography} & \multicolumn{1}{p{6.5cm}}{- Artist’ hometown and the lives of people who has lived there -CITE-“Passing of time” \newline - Her personal experiences \newline - Collision of past and present \newline - Domestic space and\newline meaningful—perhaps personal\newline —everyday lives objects  \newline - Time, temporality}  \\ \hline
\end{longtable}
\begin{adjustwidth}{-2.25in}{0in}
\noindent\footnotesize{\(^3\)}Historical, social, political, cultural conditions in which the work is created \newline
\footnotesize{\(^4\)}Art-Making Intention Regarded to historical discourse of art \newline
\end{adjustwidth}

\normalsize
\setlength{\LTleft}{-2.25in}
\setlength{\LTright}{0in}
\small
\setlength{\tabcolsep}{5pt}
\renewcommand{\arraystretch}{1.7}
\begin{longtable}{p{0.1\linewidth}|p{6.8cm}p{0.5cm}p{0.11\linewidth}|p{6.8cm}}
\captionsetup{margin=-2.25in}
\caption{\bf Galleries' Outliers} \\
{\rowcolor{lightgray}\bf\normalsize Gallery} & {\bf\normalsize Outlier \(^5\)} & \cellcolor{white} &{\bf\normalsize Gallery} & {\bf\normalsize Outlier} \\ \cline{1-2} \cline{4-5}

{\emph {Bonsai}} & Although all these images have common EXPs that unite them, this image is slightly different due to a few factors. One is the subject matter (it includes both trees and owls). Second, the composition is different as it depicts the tree and the owl from a far distance. Third, the leaves in this image are more geometrical, while the plants in other pictures are more organic. 
    \begin{center}
        \includegraphics[width=0.4\linewidth,valign=t]{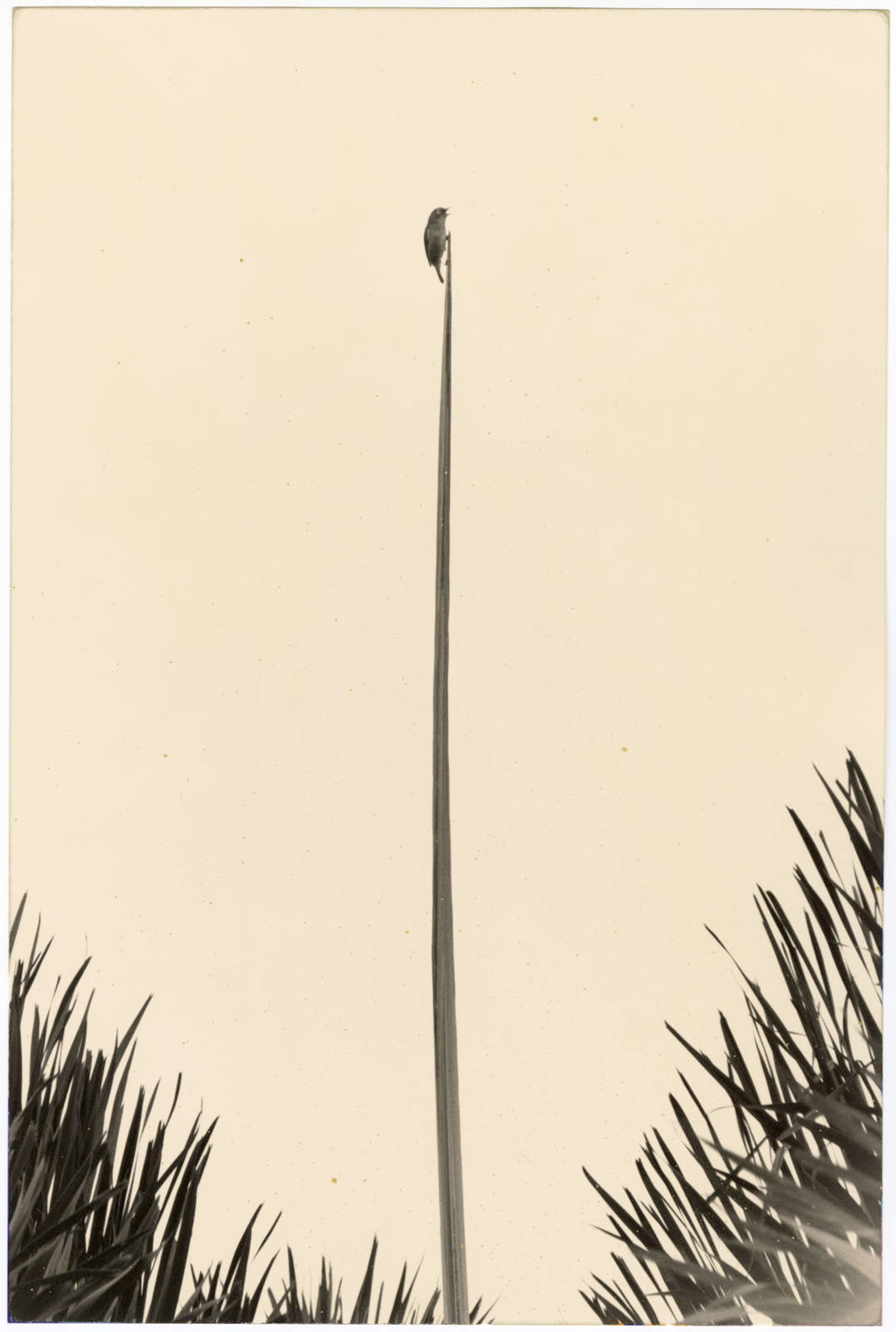}\vspace{1.4mm}
    \end{center} & &
{\emph {Epilogue}} & Not much different, seems more distinctive because of 4 parts composition(multi-frame).
    \begin{center}
        \includegraphics[width=1\linewidth,valign=t]{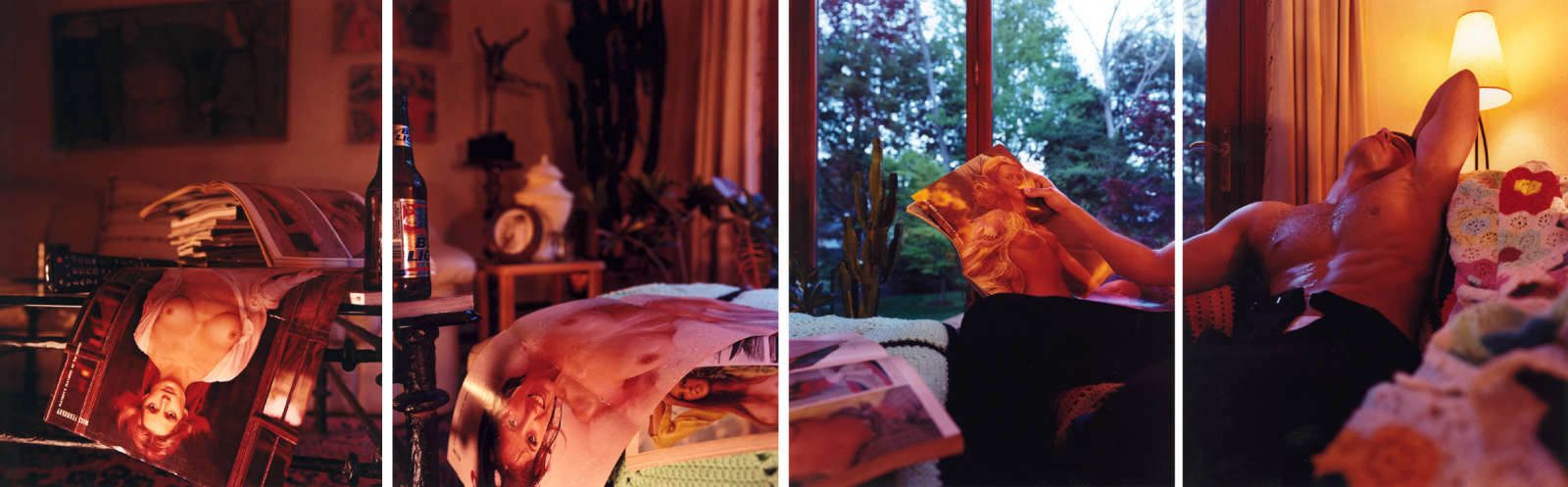}\vspace{1.4mm}
    \end{center}
\\ \cline{1-2} \cline{4-5}

{\emph {Familiar Landscapes}} & Because of having a figure. It barely is considered as an outlier. 
    \begin{center}
        \includegraphics[width=0.5\linewidth,valign=t]{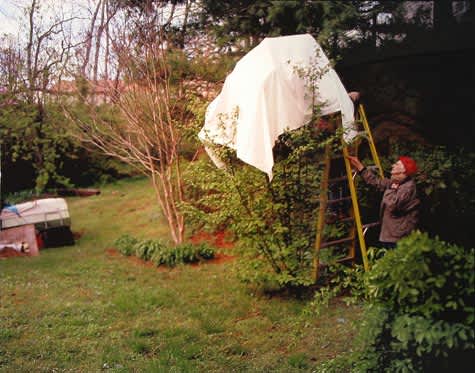}\vspace{1.4mm}  
    \end{center} & &
{\emph {Hivernacle}} & No gold paint, horizontal composition
    \begin{center}
        \includegraphics[width=0.5\linewidth,valign=t]{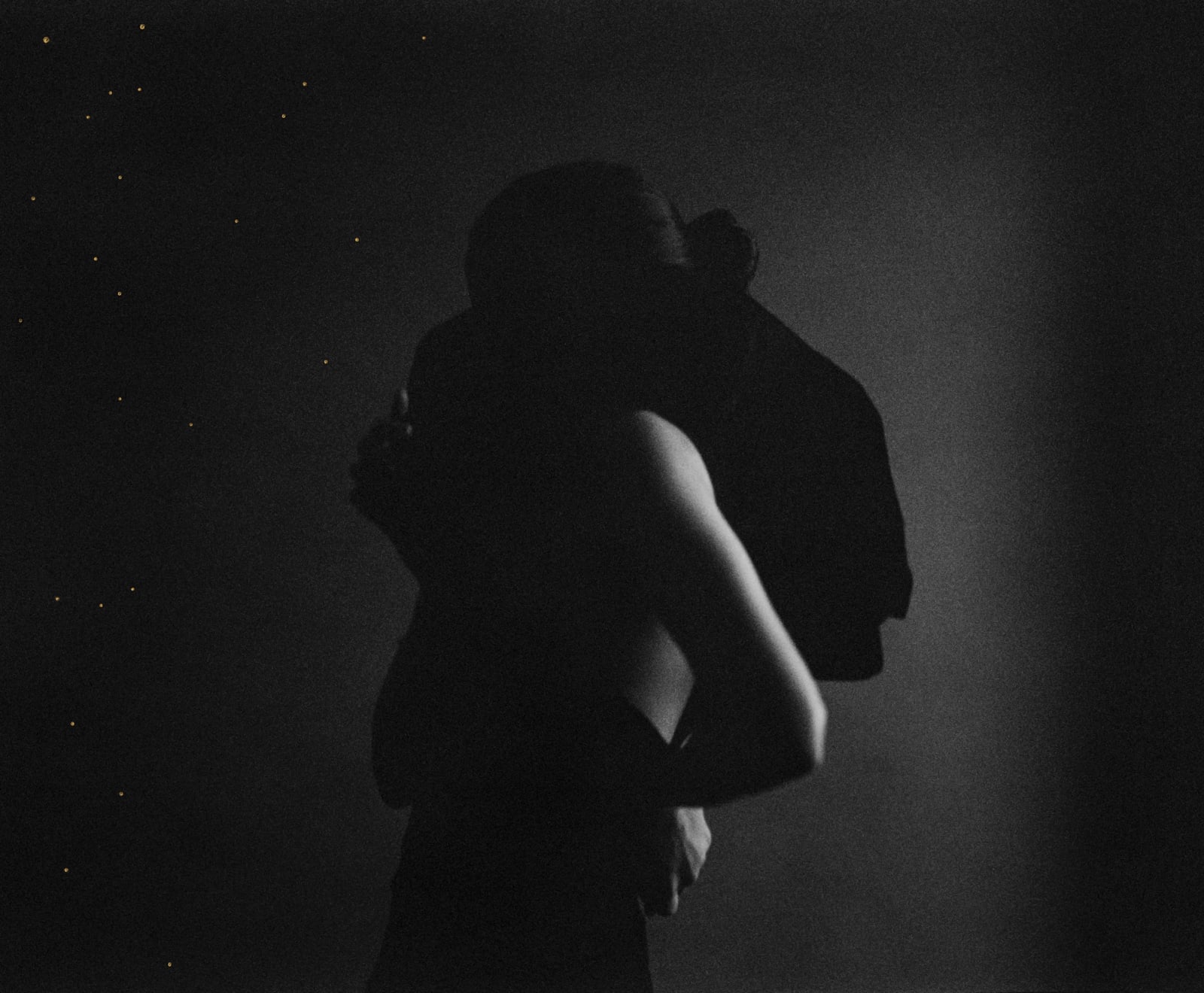}\vspace{1.4mm}
    \end{center}
\\ \cline{1-2} \cline{4-5}

{\emph {Little Deaths}} & Hardly qualify as outlier. The top one because of the close-up frame and the blurred landscape, which makes it a little different. The bottom one, because it's an interior space while the rest are landscapes. 
    \begin{center}
        \includegraphics[width=0.8\linewidth,valign=t]{Little Deaths.png}\vspace{1.4mm}
    \end{center} & &
{\emph {Mukono}}& Images in this gallery have a figure centered in the composition with a sharp tonal contrast with the background. These four images are exceptions.
    \begin{center}
        {\includegraphics[width=0.8\linewidth,valign=t]{Mukono_1.png}}
    \end{center} 
\\ \cline{1-2} \cline{4-5}
  
{\emph {Native}} & Left: Multiple numbers/layers of distinction from the rest: no nature, no human figure, different color range (e.g., less green and blue). Center \& Right: No nature, no human figure.
    \begin{center}
        \includegraphics[width=0.9\linewidth,valign=t]{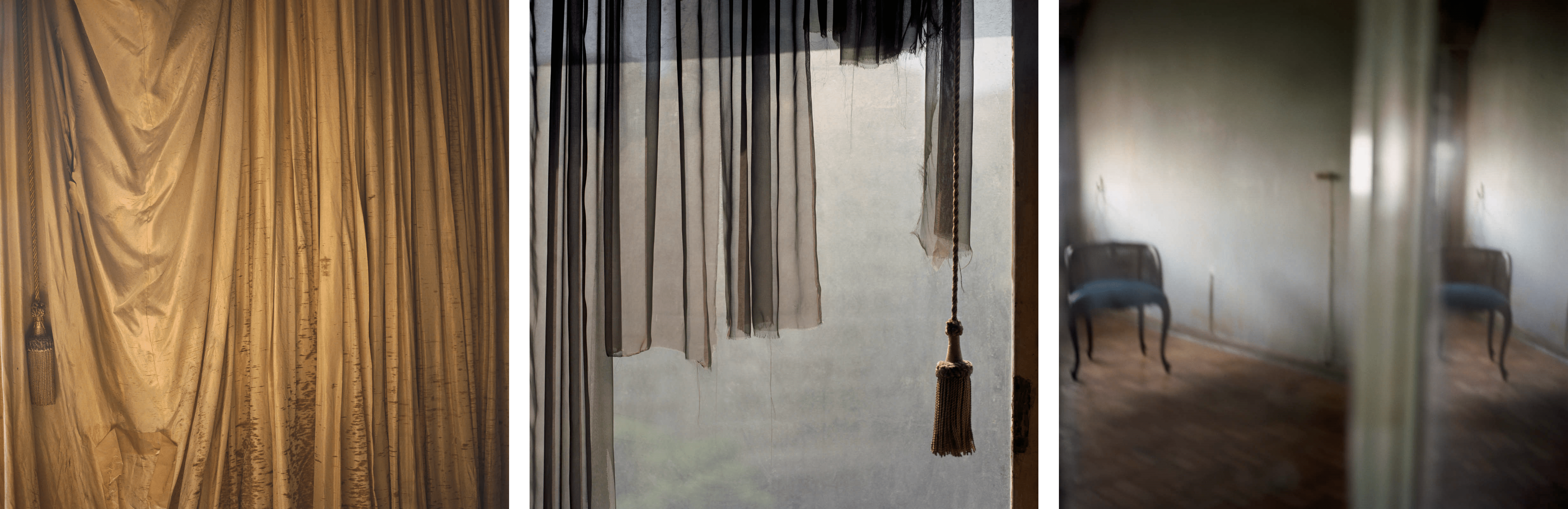}\vspace{1.4mm}  
    \end{center} & &

{\emph {Paradise Lost}} & These two were a little different in the sense that the human figure is less visible in the composition. 
    \begin{center}
        \includegraphics[width=0.7\linewidth,valign=t]{Paradise Lost.png}\vspace{1.4mm}  
    \end{center}
  \\ \cline{1-2}  \cline{4-5}

{\emph {Private}} &  The only images without human figure. 
    \begin{center}
        \includegraphics[width=1\linewidth,valign=t]{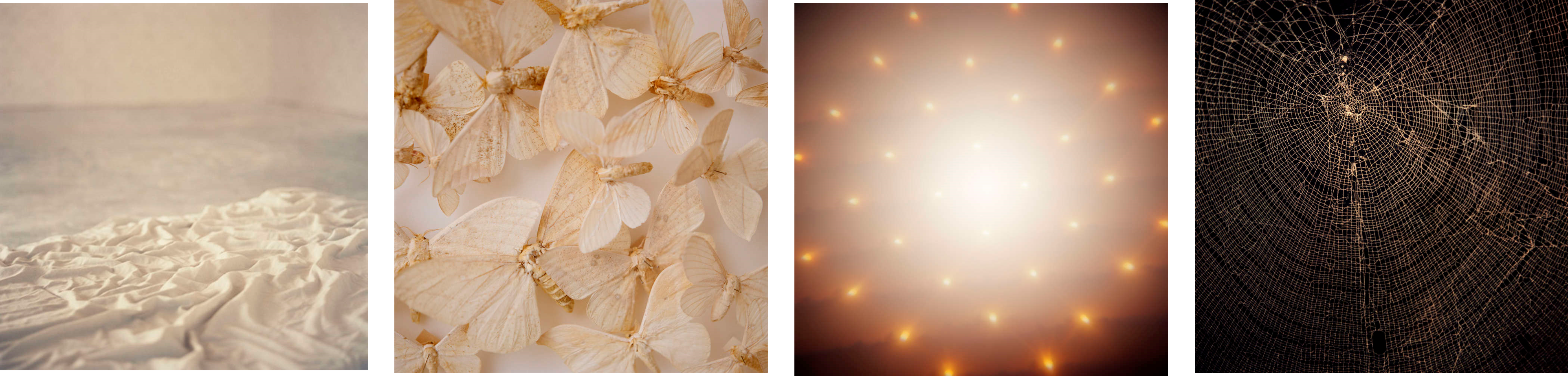}\vspace{1.4mm} 
    \end{center}  & &

{\emph {The Fall\newline of Spring Hill}} & This is the only image without figure. 
    \begin{center}
        \includegraphics[width=0.5\linewidth,valign=t]{The Fall of Spring Hill.jpg}\vspace{1.4mm}  
    \end{center}
\\ \cline{1-2} \cline{4-5}
    
{\emph {The Fallen Fawn}} & These two images have a couple of differences from the rest. They are among the images without human figures, they are also different from the rest in terms of their composition. The composition of most of the images are open composition but these two are closed (long-shot). 
    \begin{center}
        \includegraphics[width=1\linewidth,valign=t]{The Fallen Fawn.png}\vspace{1.4mm}  
    \end{center}  & &
    
{\emph {Trigger}}& The has multiple differences is not centered and is different from the rest in terms of its subject matter. 
    \begin{center}
        {\includegraphics[width=0.4\linewidth]{Trigger_1.jpg}}
    \end{center}
\\ \cline{1-2} \cline{4-5}

\end{longtable}
\end{singlespace}
\begin{adjustwidth}{-2.25in}{0in}
\noindent\footnotesize{\(^5\)}An image/images that is different from the rest due to multiple differences or one major distinction
\end{adjustwidth}

\normalsize
\subsubsection*{Dataset S1}
Galleries “Scene”, “Mukono”, and “Heat+ High Fashion” are grayscale photography containing human figures. Color and shape, as EXPs, increase the similarity between galleries. However, these galleries are conceptually very different, i.e., they have distinct NEXPs. Gallery “Heat and Hight Fashion” refers to women’s fashion and female identity during the 1960s and 1970s, gallery “Mukono” reflects contemporary (e.g., 2017) African ethnic identity, and gallery “Scene” represents New York City’s underground artistic culture during the 1980s \cite{47}. Galleries “Trigger” and “My Mother's Clothes” describe another situation as they have similar shapes/compositions, textures, and objects. Both consist of color photographs and contain everyday objects, like letter mails and casual clothing placed in the center of the frame. In addition, they are conceptual art pieces, therefore, they include strong conceptual meaning (NEXPs): gallery “Trigger” portrays the ideas on the passage of time and the artist’s self-reflection, as indicated by its description, “Her hometown and the chronology of its inhabitants, her personal experiences and connections to everyday objects, and the collision of past and present” \cite{48}. In contrast, the gallery “My Mother's Clothes” displays the artist's mother’s clothes and personal objects and was created as an attempt to cope with a loved one’s struggle with dementia and her subsequent death \cite{49}. Tables 1 to 3 summarize the EXP, NEXP, and outliers in the galleries used in Dataste S1.

\subsubsection*{Dataset S2}
From the perspective of EXPs, although galleries “Persephone” and “Painted Nudes” are similar in texture, i.e. their medium is photography with touches of paint, they have differences in color tones, shape, and objects, as one contains landscapes with natural forms and a variety of sharp tones, and the other has human figures with limited and dull tones. Finally, there is only one grayscale gallery, “Heat + High Fashion”, in this dataset. From the perspective of NEXPs, three galleries, “Bullets”, “Painted Nudes” and “Heat + High Fashion”, focus on the same concept, femininity, while the gallery “The Fall of Spring Hill” also considers femininity besides other ideas.

\subsubsection*{Dataset S3}
From the perspective of EXP dissimilarity, the gallery “Boarding House” is the only grayscale gallery, and “Painted Nudes” is the only mixed medium gallery. From the perspective of EXP similarity, galleries “Private” and “Painted Nudes” contain nude female figures, and galleries “Trigger” and “My Mother’s Clothes” have similar centerpiece compositions. From the perspective of NEXP, galleries “My Mother’s Clothes”, “Private” and “Painted Nudes” emphasize femininity.

\subsubsection*{Dataset G1}
For instance, the works in the gallery “30 Years of Women” are all by female artists, and therefore it is assumed that these works share a female perspective. We expected the dataset to be difficult to classify if two group shows were included. This is because according to Hypothesis II EXPs diversity within a gallery makes classification difficult. Also, we expected that DCNN model has its best performance for the gallery “The Unknown”, and its worst performance for the group shows “30 Years of Women” and “The Epilogue”. This is because the gallery “The Unknown” has distinct EXPs as compared to the other galleries and less diversity within itself. For example, all images in the gallery “The Unknown” have a female figure centered in the composition with a touch of a paintbrush on her face area.  

\subsubsection*{Dataset Info}
\begin{singlespace}
\setlength{\tabcolsep}{8pt}
\renewcommand{\arraystretch}{1.8}
\small 
\begin{longtable}{p{1.2cm}|p{3.6cm}|p{1.1cm}|p{0.8cm}p{0.7pt}p{1.2cm}|p{3.8cm}|p{1.1cm}|p{0.8cm}}
\captionsetup{margin=-2.25in}
\caption{\bf Dataset info: Galleries, Images (number of images per gallery), Total (total number of images\newline per dataset).}\\
\rowcolor{lightgray!50}
\bf\normalsize Dataset & \bf\normalsize Gallery & \bf\normalsize Images & \bf\normalsize  Total & \cellcolor{white} & \bf\normalsize Dataset & \bf\normalsize Gallery & \bf\normalsize Images & \bf\normalsize  Total \\

S1 & {\emph {Heat + High Fashion \newline  Mukono \newline  My Mother’s Clothes \newline Scene \newline Trigger}} & 6 \newline 16 \newline 22 \newline 13 \newline 7 & 64 &  & SF1 & {\emph {Converging Territories \newline Heat + High Fashion \newline  Hivernacle \newline Persephone \newline Trigger}} & 7 \newline 6 \newline 5 \newline 7 \newline 7 & 32 \\ \cline{1-4} \cline{6-9}

S2 & {\emph {Bullets \newline  Heat + High Fashion \newline  Painted Nudes \newline Persephone \newline The Fall of Spring Hill}} & 7 \newline 6 \newline 14 \newline 7 \newline 13 & 47 & & SF2 & {\emph {Familiar Landscapes \newline Little Deaths \newline Persephone \newline Sweet 16 \newline Trigger}} & 10 \newline 9 \newline 7 \newline 11 \newline 7 & 44 \\ \cline{1-4} \cline{6-9}

S3 & {\emph {Boarding House \newline My Mother’s Clothes \newline Painted Nudes \newline Private \newline Trigger}} & 13 \newline 22 \newline 14 \newline 16 \newline 7 & 72 & & SF3 & {\emph {Close \newline Evidence \newline  Kawa = Flow \newline  Native \newline The Fall of Spring Hill \newline The Fallen Fawn}} & 11 \newline 17 \newline 17 \newline 15 \newline 13 \newline 14 & 87 \\ \cline{1-4} \cline{6-9}

S4 & {\emph {Heat + High Fashion \newline  Mukono \newline  My Mother’s Clothes \newline Non-Art \newline Scene \newline Trigger}} & 6 \newline 16 \newline 22 \newline 34/18 \newline 13 \newline 7 & 98 & & SF4 & {\emph {Bonsai \newline Eat Flowers \newline  My Mother’s Clothes \newline New york, Paris, and Rome \newline The Garden \newline The Unknown}} & 27 \newline 25 \newline 22 \newline 32 \newline 20 \newline 18 & 144 \\ \cline{1-4} 

G1 & {\emph {30 Years of Women \newline Bullets \newline  Epilogue \newline  Paradise Lost \newline The Unknown \newline Trigger }} & 51 \newline 7 \newline 14 \newline 22 \newline 18 \newline 7 & 119 \\
\end{longtable}
\end{singlespace}

\subsection*{S3 Appendix: Follow-Up Experiments: The Effect of Dataset Size on Classification}

The few inconclusive situations observed in Experiment I were addressed by follow-up experiments to study the effect of dataset size on classification. If size, which is the number of images in each dataset, would have an impact on the DCNN model performance, then datasets of bigger sizes should produce better classification results than those of smaller sizes. However, experimental results showed otherwise (Fig 1), as the results for balanced and sample-size-conscious settings were like experiments in which size was not a factor. 

{\bf Dataset SF1 Description}. This dataset is the smallest dataset among all datasets in this work. It contains only 32 images in total, and 5 to 7 images per gallery. For this dataset, we expected the gallery “Trigger” to be the hardest gallery of the other four galleries, which also includes the gallery “Heat+ High Fashion”. The latter gallery unexpectedly was the hardest to classify in the experiment for dataset S1.

\begin{figure}[h!]
    \centering
    \includegraphics[scale=0.18]{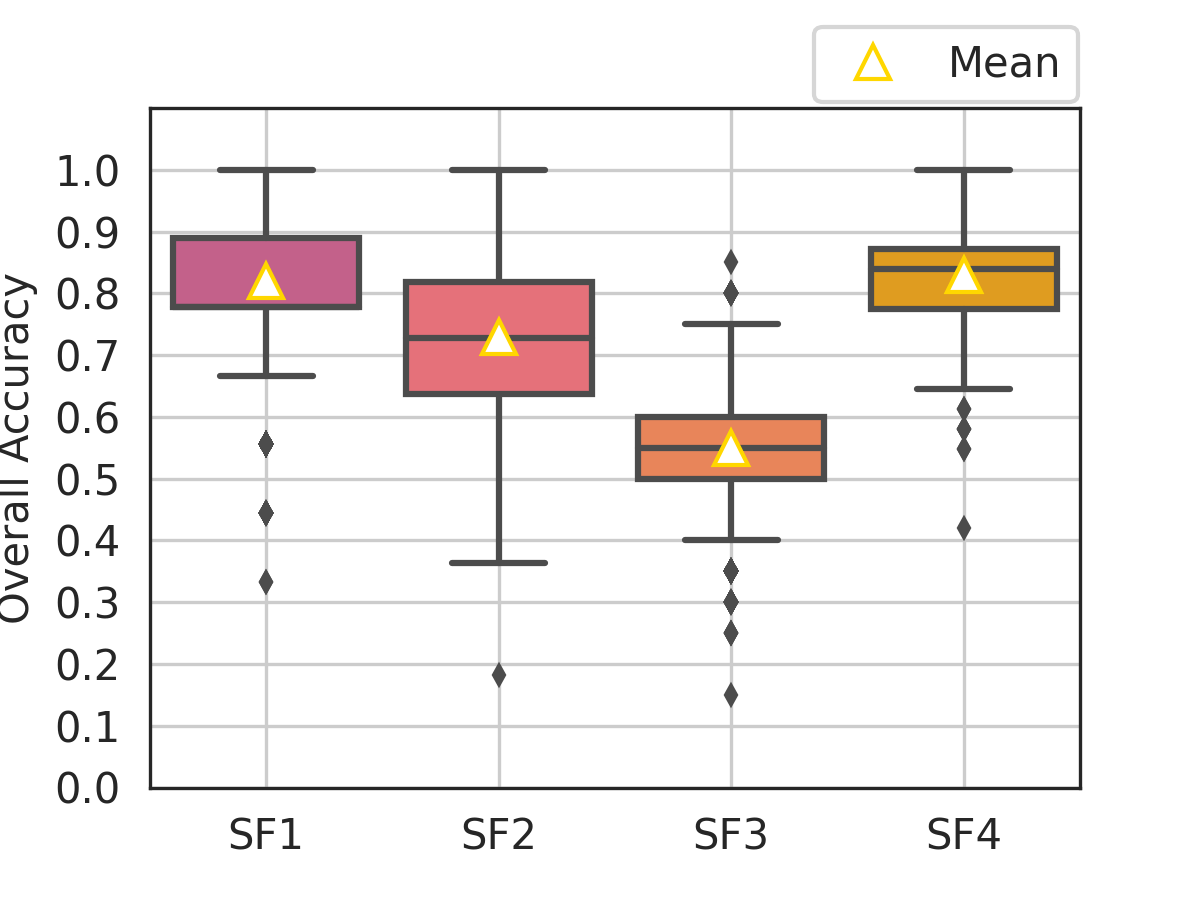}
    \caption{{\bf Box and whisker plot of overall accuracies for Datasets SF1 to SF4.}}
\end{figure}

{\bf Results and Observations}. If the small size increases the classification difficulty level of a gallery, then set SF1 should be difficult. However, the DCNN model had a high performance, MCC was 0.80 and ACC was 0.82, suggesting that the overall difficulty level of set SF1 was easy. Hence, one can induct that the small size of a gallery alone is not a determining factor in the difficulty level of a dataset. Class-wise metrics (Fig 2) show that gallery “Trigger” is indeed harder than the gallery “Heat+ High Fashion” if within-gallery EXP similarities are not present.

{\bf Dataset SF2 Description}. This dataset is the second smallest dataset among all datasets with 7 to 11 images per gallery, which are all works of a single artist. This dataset explored the DCNN model understanding of an artist’s signature. An artist’s signature is all characteristics that distinguish the artist’s work. An artist may or may not be aware of them, whereas an artist’s style is those characteristics that the artist intentionally adopts in his/her work. 
Results and Observations. If DCNN model would consider and understand NEXPs, it will be confused by the elements of an artist’s signature, thus it will perform poorly in this case. But if hypotheses I and II are correct, the model performs well since it ignores any common elements of the galleries of to the same artist. Experimental results support hypotheses I and II. The model performed well as MCC was 0.69 and ACC was 0.73. The hardest gallery was “Little Deaths” (Fig 2), which shares EXPs with galleries “Familiar Landscapes” and “Sweet 16”.

\begin{figure}[h!]
    \centering
        \includegraphics[width=\textwidth]{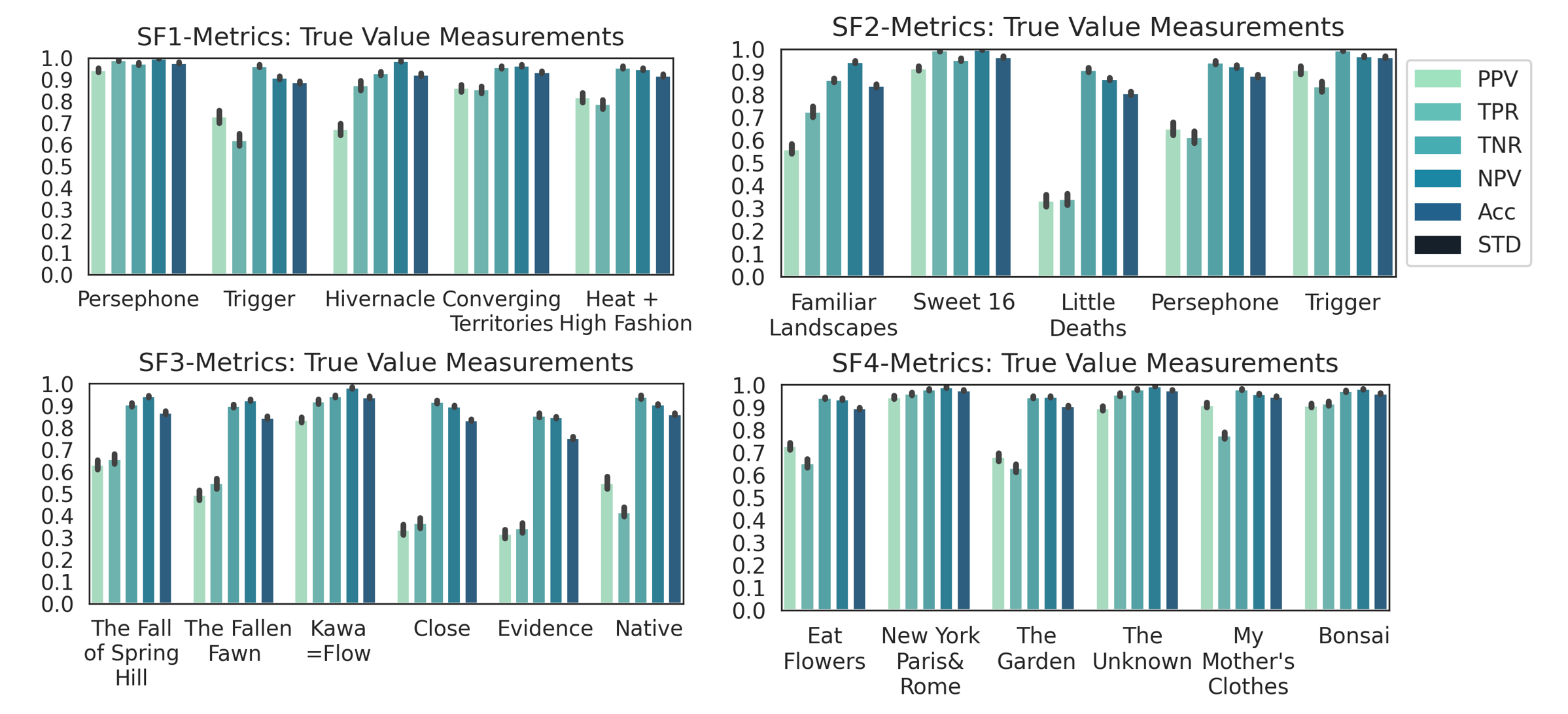}
    \caption{{\bf Class-wise metrics results for Datasets SF1 to SF4.}}
\end{figure}  
 
{\bf Dataset SF3 Description}. This dataset re-examined the setting with very similar EXPs and dissimilar NEXPs in a more balanced fashion, where the number of images in the galleries ranged from 11 to 17 images. Gallery pairs “The Fall of Spring Hill” and “The Fallen Fawn”, and “Close”, “Evidence”, and “Native” share EXPs but not NEXPs, and thus, we expected them to be confusing for the DNN model. Gallery “Kawa = Flow” is very distinct in its EXP from the other galleries in set SF3, and although it contains conceptual art pieces, e.g., NEXPs, hypotheses I and II would suggest obtaining a good classification performance for this gallery.
Results and Observations. MCC value was 0.48 and ACC value was 0.55. They confirmed the earlier findings for set S1. Hence, the balanced and size-conscious setting suggests that size is not a decisive factor in the level of difficulty. The class-wise metrics support our expectation that paired galleries were indeed causing difficulties for the DNN model and that gallery “Kawa = Flow” is the easiest in the SF3 set.
Dataset SF4 Description. This dataset is the largest dataset among the four datasets. It contains 144 images in total and 18 to 32 images per gallery. If size would influence the difficulty level, it should be the easiest classify. 
Results and Observations. MCC value was 0.80 and ACC value was 0.83. They suggest that the level of difficulty of this dataset was almost the same as that of the smallest dataset (SF1). These results also indicate that sample size on its own does not determine the level of difficulty for a dataset. Class-wise results denote galleries “The Garden” and “Eat Flowers” are the most difficult, which is consistent with the previous analysis as they contain NEXPs.

\newpage
\subsection*{S4 Appendix: Histogram Plots}
\begin{figure}[h]
    \includegraphics[width=\textwidth]{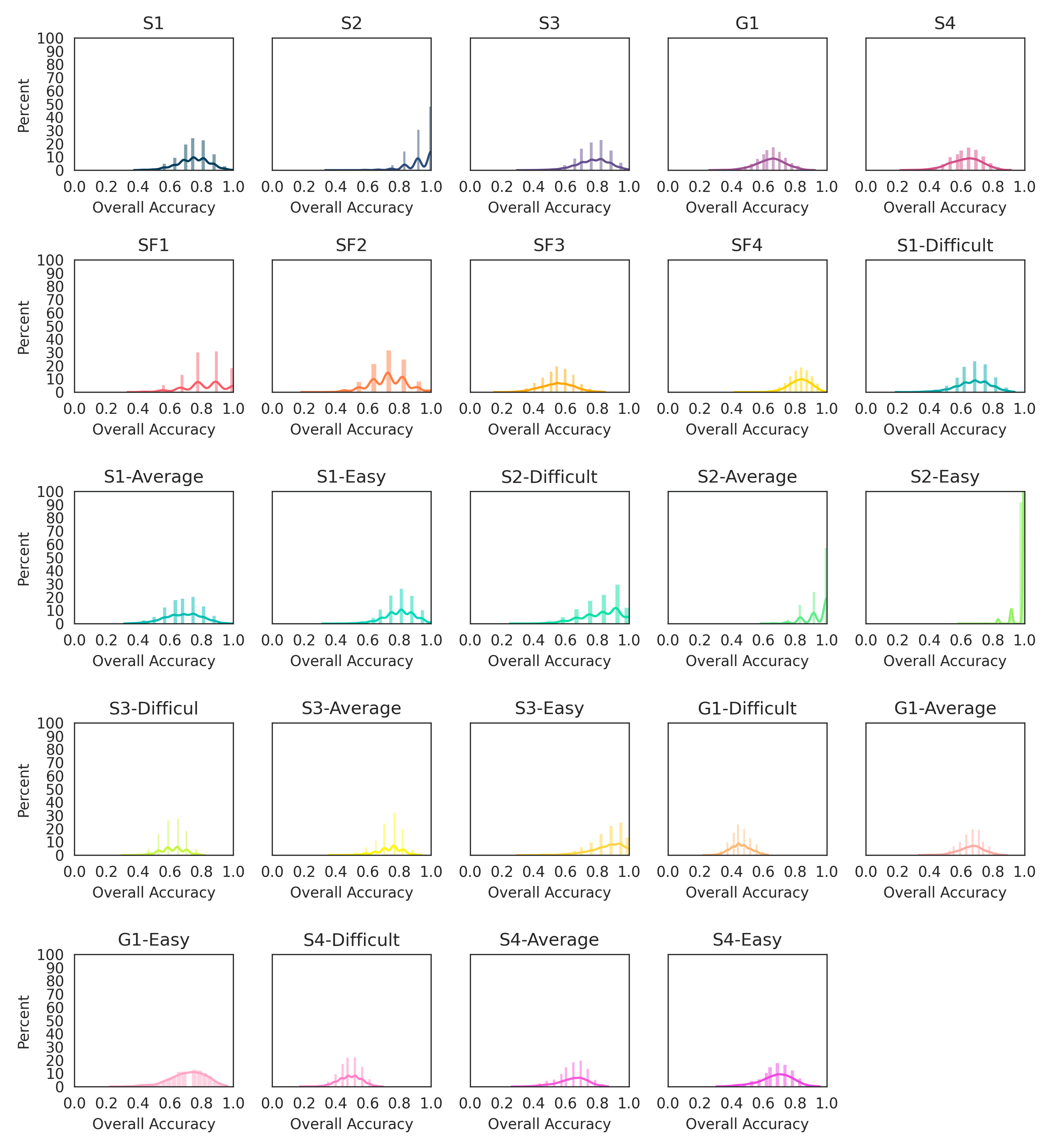}
    \caption{{\bf Accuracy histogram plots for all sets.} Histogram plots verify the distribution of the measured random variable.}
\end{figure} 

\section*{S5 Appendix: Summary Box Plots}

Fig 3 summarizes the experimental results showing that were employed to verify Hypotheses I and Hypothesis II. Experiments used datasets designed by an art expert, where datasets were of various difficulty levels. The obtained results and their agreement with PCA representations proved that the main parameter in the determination of the difficulty level was EXPs similarities among and within different galleries. That is when there were more similarities between EXPs of galleries, or there were more dissimilarities within a gallery the art curation classification was more difficult for the DCNN model. The diversity within EXPs of a gallery has much more presence in the art curation because curation often employs multiple artists, styles, media, etc. to create profound themes and meanings. In fact, the empirical finding shows that the mentioned causes more difficulties; as shown in Fig 3, the overall metrics of sets G1 and S4 are lower than S1, S2, and S3 sets, both in random (Fig 3(a)) and leveled sets (Fig 3(b)). The size remained to be irrelevant as a sole stressor given that the mentioned sets -G1 and S4- are among the largest sets.

\begin{figure}[h]
    \centering
       \includegraphics[width=\textwidth]{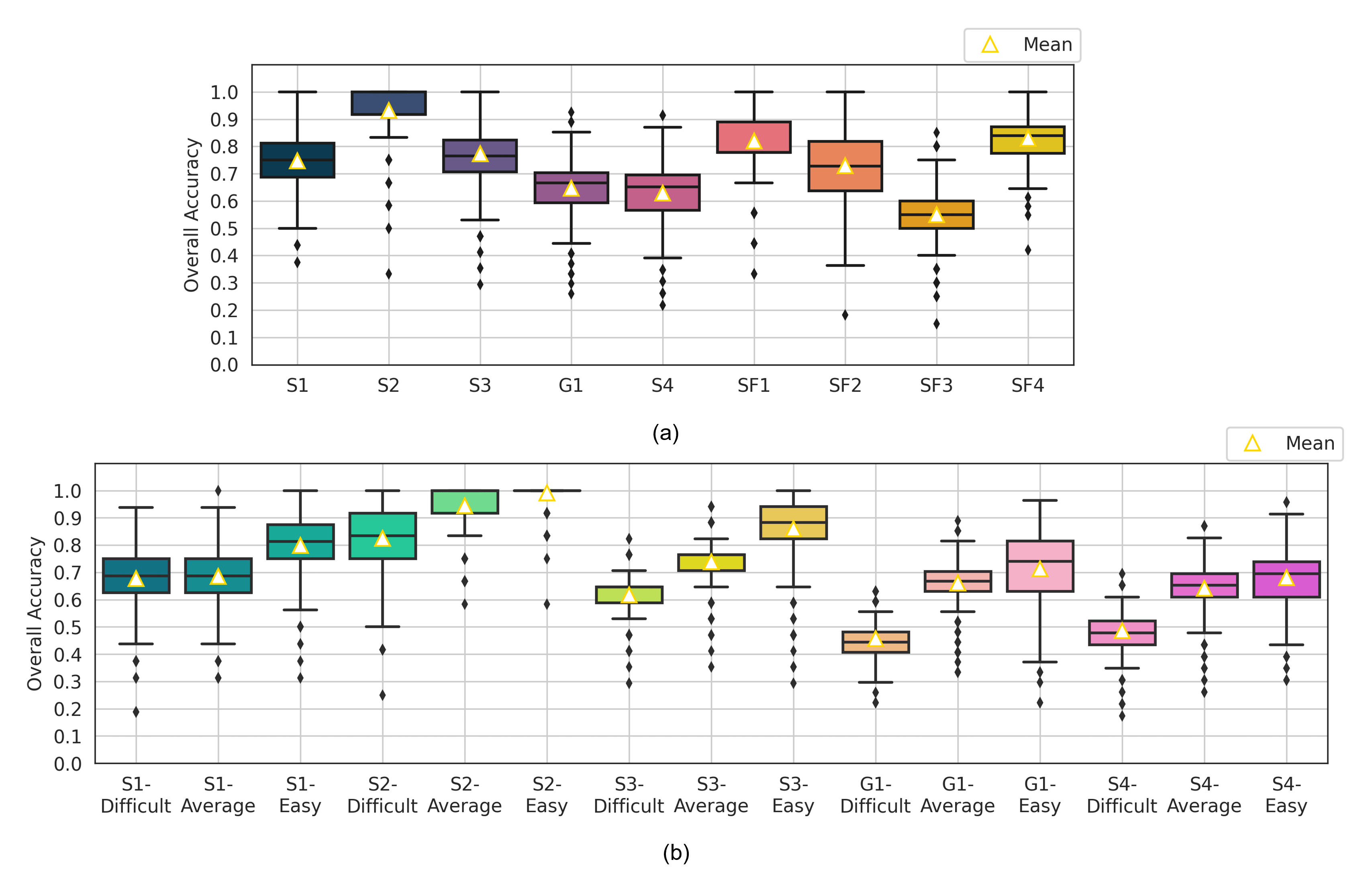}
    \caption{{\bf Box Plot Summaries.}(a) Box and whisker plot of overall accuracies for all datasets random sets (b) Box and whisker plot of overall accuracies for all datasets handpicked sets (difficult, average, and easy).}
\end{figure}  

\newpage
\subsection*{S6 Appendix: Normal Q-Q Plots}
\begin{figure}[h]
    \includegraphics[width=\textwidth]{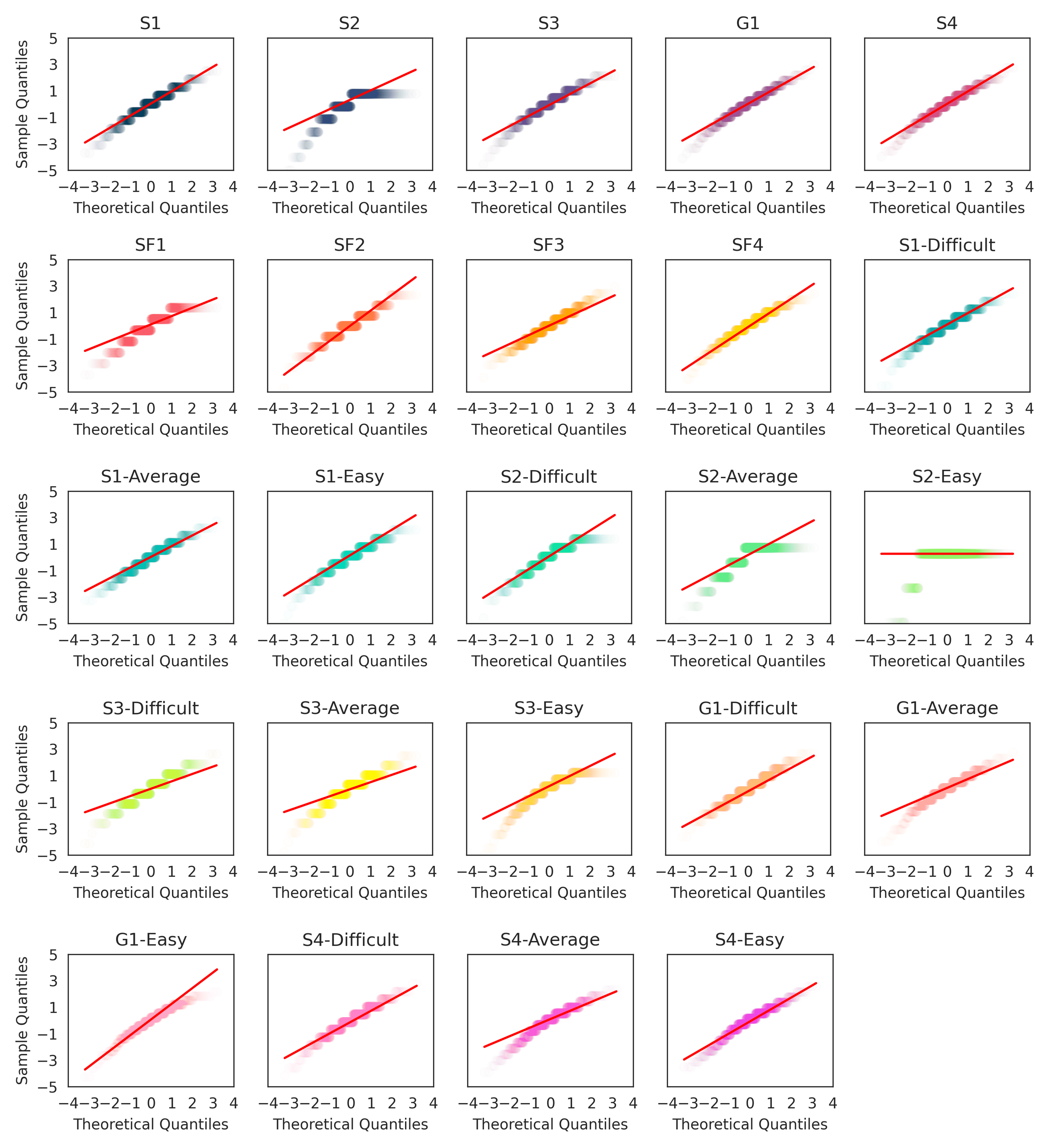}
    \caption{{\bf Normal Q-Q Plots for all sets.} Normal Q-Q plots are another method for verify the distribution of the measured random variable.}
\end{figure} 

\newpage
\subsection*{S7 Appendix: Class-wise Metrics (FPR, FNR, and FDR)}
\begin{figure}[h!]
    \includegraphics[width=\textwidth]{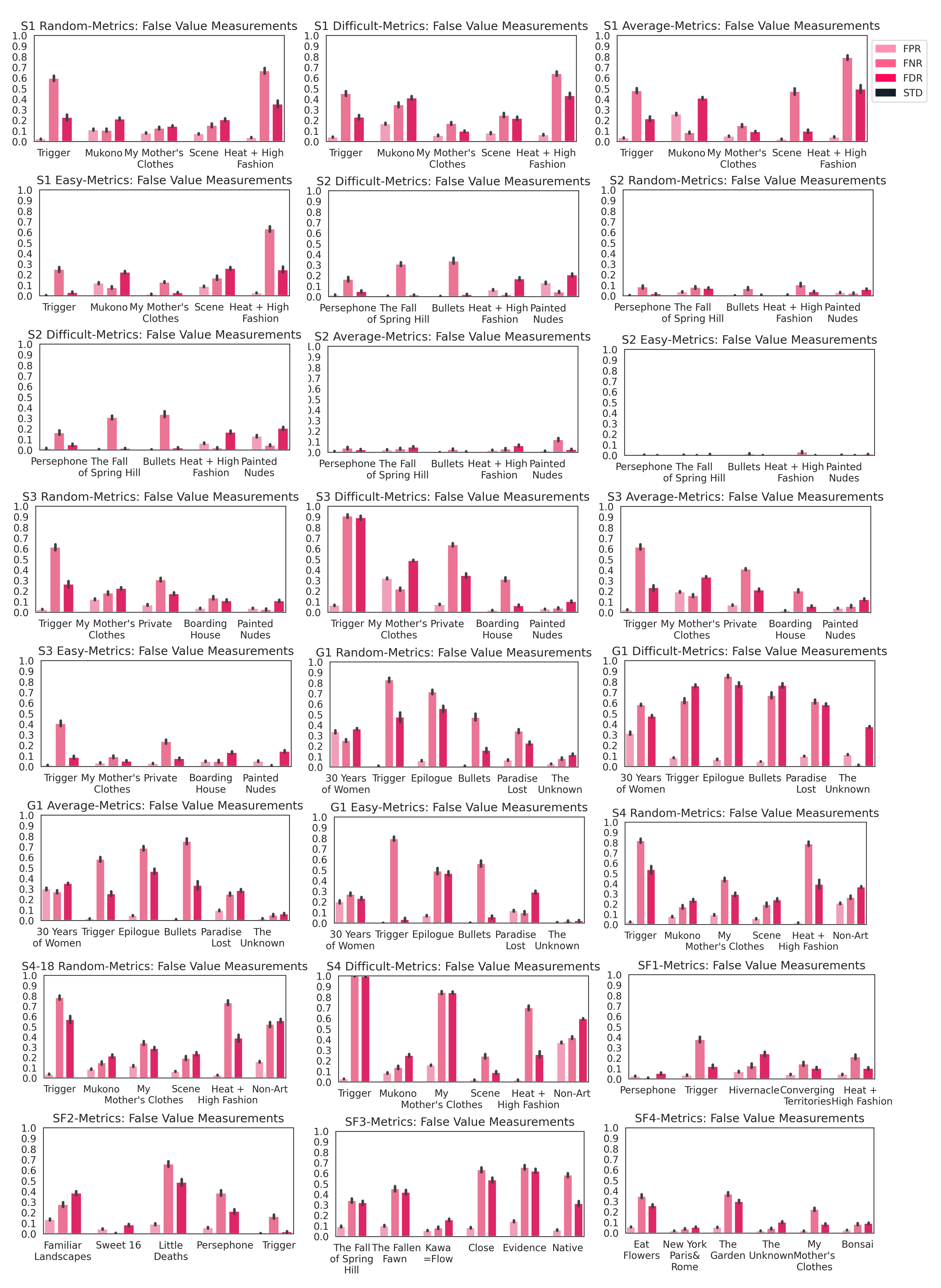}
    \caption{{\bf Remaining class-wise metrics (FPR, FNR, and FDR) } }
\end{figure} 

\newpage\subsection*{S8 Appendix: Statistical Details}
\begin{table}[H]
\begin{adjustwidth}{-2.25in}{0in} 
\caption{\bf Numerical values of statistical tests and measures.}
\setlength{\tabcolsep}{8pt}
\renewcommand{\arraystretch}{1.2}
\begin{tabular}{c|c|c|c|c|c|c|c|c|c|c}
\rowcolor{lightgray!50}
\bf Dataset &	\bf Max & \bf	Min &	\bf STD &	\bf Mean &	\bf Median &  \bf SW_p &	\bf Leven’s &	\bf DT3 &	\bf G-H &	\bf Tukey’s \\ 
\bf S1 &	1.000	& 0.375 &	0.101 &	0.747 &	0.750 &	\(< 0.001\) &	\(< 0.001\) &	\(< 0.001\) &	\(< 0.001\) &	\(< 0.001\) \\ \hline
\bf Difficult &	0.938	& 0.188 &	0.108 &	0.677 &	0.688 & \(< 0.001\) &	\(< 0.001\) &	0.374 &	0.283 &	0.236 \\ \hline
\bf Average &	1.000 &	0.313 &	0.115 &	0.684 &	0.688 &	\(< 0.001\) &	\(< 0.001\) &	0.374 &	0.283 &	0.236 \\ \hline
\bf Easy &	1.000 &	0.313 &	0.097 &	0.798 &	0.813 &	\(< 0.001\) &	\(< 0.001\) &	\(< 0.001\)&	\(< 0.001\) &	\(< 0.001\) \\ \hline \hline
\bf S2 &	1.000 &	0.333 &	0.087 &	0.931 &	0.917 &	\(< 0.001\) &	\(< 0.001\) &	\(< 0.001\) &	\(< 0.001\) &	\(< 0.001\) \\ \hline
\bf Difficult &	1.000 &	0.250 &	0.126 &	0.825 &	0.833 &	\(< 0.001\) &	\(< 0.001\) &	\(< 0.001\) &	\(< 0.001\) &	\(< 0.001\) \\ \hline
\bf Average &	1.000 &	0.364 &	0.069 &	0.952 &	1.000 &	\(< 0.001\) &	\(< 0.001\) &	\(< 0.001\) &	\(< 0.001\) &	\(< 0.001\) \\ \hline
\bf Easy &	1.000 &	0.583 &	0.032 &	0.991 &	1.000 &	\(< 0.001\) &	\(< 0.001\) &	\(< 0.001\) &	\(< 0.001\) &	\(< 0.001\) \\ \hline \hline
\bf S3 &	1.000 &	0.294 &	0.106 &	0.774 &	0.765 &	\(< 0.001\) &	\(< 0.001\) &	\(< 0.001\) &	\(< 0.001\) &	\(< 0.001\) \\ \hline
\bf Difficult &	0.824 &	0.294 &	0.079 &	0.617 &	0.647 &	\(< 0.001\) &	\(< 0.001\) &	\(< 0.001\) &	\(< 0.001\) &	\(< 0.001\) \\ \hline 
\bf Average &	0.941 &	0.353 &	0.081 &	0.738 &	0.765 &	\(< 0.001\) &	\(< 0.001\) &	\(< 0.001\) &	\(< 0.001\) &	\(< 0.001\) \\ \hline
\bf Easy &	1.000 &	0.294 &	0.113 &	0.860 &	0.882 &	\(< 0.001\) &	\(< 0.001\) &	\(< 0.001\) &	\(< 0.001\) &	\(< 0.001\) \\ \hline \hline
\bf G1 &	0.926 &	0.259 &	0.094 &	0.646 &	0.667 &	\(< 0.001\) &	\(< 0.001\) &	\(< 0.001\) &	\(< 0.001\) &	\(< 0.001\) \\ \hline
\bf Difficult &	0.630 &	0.222 &	0.065 &	0.456 &	0.444 &	\(< 0.001\) &	\(< 0.001\) &	\(< 0.001\) &	\(< 0.001\)&	\(< 0.001\) \\ \hline 
\bf Average &	0.889 &	0.333 &	0.082 &	0.660 &	0.667 &	\(< 0.001\) &	\(< 0.001\) &	\(< 0.001\) &	\(< 0.001\)&	\(< 0.001\) \\ \hline
\bf Easy &	0.963 &	0.222 &	0.116 &	0.713 &	0.741 &	\(< 0.001\) &	\(< 0.001\) &	\(< 0.001\) &	\(< 0.001\) &	\(< 0.001\) \\ \hline \hline
\bf S4 &	0.913 &	0.217 &	0.104 &	0.628 &	0.652 &	\(< 0.001\) &	\(< 0.001\) &	0.002 &	0.002 &	0.001\\ \hline
\bf Difficult &	0.696 &	0.174 &	0.075 &	0.486 &	0.478 &	\(< 0.001\) &	\(< 0.001\) &	\(< 0.001\) &	\(< 0.001\) &	\(< 0.001\) \\ \hline 
\bf Average &	0.870 &	0.261 &	0.098 &	0.642 &	0.652 &	\(< 0.001\) &	\(< 0.001\) &	0.002 &	0.002 &	0.001\\ \hline
\bf Easy &	0.957 &	0.304 &	0.107 &	0.681 &	0.696 &	\(< 0.001\) &	\(< 0.001\) &	\(< 0.001\) &	\(< 0.001\)&	\(< 0.001\) \\ \hline \hline
\bf SF1 & 1.000 &	0.333 &	0.132 &	0.821 &	0.778 &	\(< 0.001\)	& \(< 0.001\) &	0.162 &	0.128 &	0.141\\ \hline
\bf SF2	& 1.000 &	0.182 &	0.117 &	0.729 &	0.727 &	\(< 0.001\) &	\(< 0.001\) &	\(< 0.001\) &	\(< 0.001\)&	\(< 0.001\) \\ \hline
\bf SF3 &	0.850 &	0.150 &	0.103 &	0.550 &	0.550 &	\(< 0.001\) &	\(< 0.001\) &	\(< 0.001\) &	\(< 0.001\)&	\(< 0.001\) \\ \hline
\bf SF4 &	1.000 &	0.419 &	0.070 &	0.829 &	0.839 &	\(< 0.001\) &	\(< 0.001\) &	0.162 &	0.128 &	0.141\\ 
\end{tabular}
\end{adjustwidth}
\end{table}

\bibliographystyle{Vancouver}
\bibliography{Bib.bib}